\relax
\documentclass[letterpaper]{article} % DO NOT CHANGE THIS
\usepackage{aaai20}  % DO NOT CHANGE THIS
\usepackage{todonotes}
\usepackage{times}  % DO NOT CHANGE THIS
\usepackage{helvet} % DO NOT CHANGE THIS
\usepackage{courier}  % DO NOT CHANGE THIS
\usepackage[hyphens]{url}  % DO NOT CHANGE THIS
\usepackage{graphicx} % DO NOT CHANGE THIS
\usepackage{ctable}
\usepackage{subcaption}
\usepackage{pdfpages}
\usepackage{graphicx}  % DO NOT CHANGE THIS
\author{Christopher M.~Bender,\footnote[2]{denotes equal contribution}\textsuperscript{\rm 1} Kevin O'Connor,\textsuperscript{\rm $\dagger$ 2} Yang Li,\textsuperscript{\rm 1} Juan Jose Garcia,\textsuperscript{\rm 1} Manzil Zaheer,\textsuperscript{\rm 3} Junier Oliva\textsuperscript{\rm 1}\\ 
\textsuperscript{\rm 1}Department of Computer Science, UNC Chapel Hill \\
\textsuperscript{\rm 2}Department of Statistics and Operations Research, UNC Chapel Hill \\
\textsuperscript{\rm 3}Google Research \\
\texttt{\{bender,yangli95,jjgarcia,joliva\}@cs.unc.edu}, \, \texttt{koconn@live.unc.edu},\,  \texttt{manzilz@google.com} \\
}

\usepackage[utf8]{inputenc} % allow utf-8 input
\usepackage[T1]{fontenc}    % use 8-bit T1 fonts
\usepackage{url}            % simple URL typesetting
\usepackage{booktabs}       % professional-quality tables
\usepackage{amsfonts}       % blackboard math symbols
\usepackage{nicefrac}       % compact symbols for 1/2, etc.
\usepackage{microtype}      % microtypography
\usepackage{amsmath}
\usepackage{graphicx}
\usepackage{subcaption}
\usepackage{xspace}
\usepackage{amsthm}
\usepackage{booktabs} % for professional tables
\usepackage{enumitem}
\usepackage[symbol]{footmisc}
\renewcommand{\thefootnote}{\fnsymbol{footnote}}
\newcommand\blfootnote[1]{%
  \begingroup
  \renewcommand\thefootnote{}\footnote{#1}%
  \addtocounter{footnote}{-1}%
  \endgroup
}

\usepackage{xcolor}

\newcommand{\Abs}[1]{\left\lvert #1 \right\rvert}

\newcommand{\pind}[2]{#1^{(#2)} }
\newcommand{\xind}[1]{x^{(#1)} }

\newcommand{\bx}{\mathbf{x}}

\newcommand{\R}{\mathbb{R}}

\newcommand{\calX}{\mathcal{X}}

\newcommand{\calD}{\mathcal{D}}

\newcommand{\calA}{\mathcal{A}}

\newcommand{\ud}{\mathrm{d}}

\newcommand{\tiid}{\emph{i.i.d.}\xspace}

\newcommand{\bruno}{BRUNO\xspace}
\newcommand{\vae}{NS\xspace}
\newcommand{\method}{FlowScan\xspace}

\newcommand{\reffig}[1]{Fig.~\ref{#1}}
\newcommand{\reftab}[1]{Tab.~\ref{#1}}

\newcommand{\refeq}[1]{Eq.~\ref{#1}}

\newtheorem{prop}{Proposition}

\title{Exchangeable Generative Models with Flow Scans}

\begin{document}

\maketitle

\begin{abstract}
In this work, we develop a new approach to generative density estimation for exchangeable, non-\tiid data.
The proposed framework, \method, combines invertible flow transformations with a sorted scan to flexibly model the data while preserving exchangeability.
Unlike most existing methods, \method exploits the intradependencies within sets to learn both global and local structure.
\method represents the first approach that is able to apply sequential methods to exchangeable density estimation without resorting to averaging over all possible permutations.
We achieve new state-of-the-art performance on point cloud and image set modeling.
\end{abstract}

\section{Introduction}
Modeling unordered,  non-\tiid data is an important problem in machine learning and data science. 
Collections of data objects with complicated intrinsic relationships are ubiquitous.\blfootnote{\textsuperscript{\rm $\dagger$}Equal contribution}\footnote{This paper is an updated version of preliminary work detailed in \cite{bender2019permutation}}
These collections include sets of 3d points sampled from the surface of complicated shapes like human organs, sets of images shared within the same web page, or point cloud LiDAR data observed by driverless cars.
%It is clear that 
In any of these cases, the collections of data objects do not possess any inherent ordering of their elements.
%For this reason,
Thus, any generative model which takes these data as input should not depend on the order in which the elements are presented \emph{and} must be flexible enough to capture the dependencies between co-occurring elements.

\begin{figure}[t]
    \centering
    \includegraphics[scale=0.4]{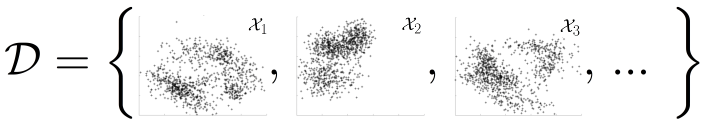}
    \caption{A training dataset of sets. Each instance $\calX_i$ is a set of points $\calX_i = \{x_{i,j}\in \mathbb{R}^d\}_{j=1}^{n_i}$ ($d=2$ shown). We estimate $p(\calX_i)$, 
    %learning a generative model 
    from which we can sample distinct sets.}
    \label{fig:setsofsets_graphic}
    %\vspace{-3mm}
\end{figure}

The unorderedness of these kinds of collections is captured probabilistically by the notion of \emph{exchangeability}.
Formally, a set of points $\{x_j\}_{j=1}^n \subset \mathbb{R}^d$ with cardinality $n$, dimension $d$, and probability density $p(\cdot)$ is called exchangeable if
\begin{align}\label{eq:exchangeable}
    p(x_1, ..., x_n) = p(x_{\pi_1}, ..., x_{\pi_n})
\end{align}
for every permutation $\pi$.
In practice $\{x_j\}_{j=1}^n$ often represent 2d or 3d spatial points (see \reffig{fig:setsofsets_graphic}) in which case we refer to the set as a point cloud.
In other settings, the points of interest may be more complex like images represented as very high-dimensional vectors. 

As a simple example, one may trivially generate a set of exchangeable points by drawing them \tiid from some distribution.
More commonly, elements within an exchangeable set share information with one another,
providing structure.
Despite the abundance of such data, the bulk of existing approaches either ignore the relation between points
(\tiid methods) or model dependencies in a manner that depends on inherent orderings (sequential methods) \cite{rezatofighi2017deepsetnet,you2018graphrnn}.
In order to accurately learn the structure of a set whilst preserving the exchangeability of its likelihood, one cannot rely solely on either approach.

In this work, we focus on the task of tractable, non-\tiid density estimation for exchangeable sets.
We explore both low cardinality sets of high dimension (10-20 points with many hundreds of dimensions each, e.g.~collections of images) and high cardinality sets of low dimension (hundreds of points with 2-7 dimensions each, e.g.~point clouds).
We develop a generative model suitable for exchangeable sets in either regime, called \method, which does not rely on \tiid assumptions and is provably exchangeable.
%Contrary to intuition, we show that one can scan along the data in a sorted manner not only preserving exchangeability,\todo{run on sentence} but also yielding a model which outperforms the current state of the art in both point cloud and image set density estimation tasks. 
Contrary to intuition, we show that one can preserve exchangeability while scanning over the data in a sorted manner.
\emph{\method is the first method to achieve a tractable, non-\tiid, exchangeable likelihood by leveraging traditional (e.g.~sequential), non-exchangeable density estimators.}
%\emph{\method is the first method to leverage traditional (e.g.~sequential) non-exchangeable density estimators for a tractable exchangeable likelihood model}.
%This technique, when combined with equivariant flow transformations and an autoregressive base likelihood, outperforms the current state of the art in both point cloud and image set density estimation tasks.

\textbf{Main Contributions.}
1) We show that transforming points with an equivariant change of variables allows for modeling sets in a different space.
2) We introduce a scanning-based technique for modeling exchangeable data, relating the underlying exchangeable likelihood to that of the sorted covariates.
%3) We demonstrate that this scanning-based approach unlocks the powerful tools of traditional density estimators, particularly sequential autoregressive likelihoods, for the task of exchangeable density modeling.
3) We demonstrate how traditional density estimators may be used for the task of principled and feasible exchangeable density estimation via a scanning-based approach.
% 3) We expand/extend/pioneer traditional density estimators for the task of principled and feasible exchangeable density estimation via a scanning-based approach.
%\todo{mention that it is first such approach}
4) We show empirically that \method achieves the state-of-the-art for density estimation tasks in both synthetic and real-world point cloud and image set datasets.

\section{Motivation and Challenges}
% I don't think we are "recasting" the problem
%Before describing the details of our methods, we recast the problem of exchangeable density estimation in terms of learning a generative model \emph{over generative models} and discuss the challenges therein.
%Before describing the details of our methods, we motivate exchangeable density estimation in terms of learning a generative model \emph{over generative models}.
%Furthermore, we discuss the challenges therein, and the limitation of an \tiid approach.
%To see how modeling set data fits into this task, consider a set-generative process, $p(\calX \mid \phi)$, given \emph{latent} ``parameters'' $\phi$. 
We motivate our problem with a simple, yet common, set generative process that requires a \emph{non}-\tiid, exchangeable density estimator.
Consider the following generative process for a set: 1) generate \emph{latent} ``parameters'' $\phi \sim p_\Phi(\cdot)$ and then 2) generate a set $\calX \sim p(\cdot \mid \phi)$.
Here $p(\cdot \mid \phi)$ may be as simple as a Gaussian model (where $\phi$ is the mean and covariance parameters) or as complex as a nonparametric model (where $\phi$ may be infinite-dimensional).
%As we explain below, this process can generate typical point cloud data.

% Consider a set-generative process, $p(\calX \mid \phi)$, given \emph{latent} ``parameters'' $\phi$. 
% Here $p(\cdot \mid \phi)$ may be as simple as a Gaussian model (where $\phi$ is the mean and covariance parameters) or as complex as a nonparametric model (where $\phi$ may be infinite-dimensional).
% Also consider a prior generative process over parameters, $p_\Phi(\phi)$. 
% The complete generative process for a set is to: 1) generate a parameter $\phi \sim p_\Phi(\cdot)$ and then 2) generate a set $\calX \sim p(\cdot \mid \phi)$.
% As we explain below, this process can generate typical point cloud data.
%(Thus, this akin to generating multiple \emph{generative models} $p(\cdot \mid \phi)$.)
%When observing a dataset \emph{of sets} from this process, we have
%$\calD = \{ \calX_i \sim p(\cdot\mid \phi_i) \}_{i=1}^N$ (for unobserved $\phi_i$'s). As each set is drawn from its own generative process, modeling $\calD$ is akin to learning to produce a generative model over generative models.

%Next, we highlight the short-comings of an \tiid approach, even for the case when the ground truth point generative process is \emph{conditionally} \tiid.
This simple set generative process requires a \emph{non}-\tiid approach, even for the case when the ground truth conditional set likelihood, $p(\calX \mid \phi)$, is \emph{conditionally} \tiid.
We show this by first noting that
with \emph{conditionally} \tiid ${p(\calX \mid \phi)} = {\prod_{j=1}^n p(x_j \mid \phi)}$, the complete set likelihood is:
\begin{align}
    p(\calX) = \int p_\Phi(\phi) \prod_{j=1}^n p(x_j \mid \phi)  \,\ud \phi. \label{eq:gen_proc_sets}
\end{align}
(Note, that \refeq{eq:gen_proc_sets} is in the same vein as De Finetti's theorem \cite{bernardo2009bayesian}.)
%However, the complete process above \eqref{eq:gen_proc_sets} does \emph{not} produce \emph{marginally} \tiid sets even in the case where $p(\calX \mid \phi)$ is conditionally \tiid.
%Even though $p(\calX \mid \phi)$ is \emph{conditionally} \tiid here, the complete process above \eqref{eq:gen_proc_sets} does \emph{not} produce \emph{marginally} \tiid sets.
One can show dependency (\emph{non}-\tiid) with the conditional likelihood of a single point $x_k$ given a disjoint subset $S \subset \calX \setminus \{x_k\}$: 
$
    p(x_k \mid S) 
    = {\int p_\Phi(\phi \mid S)\, p(x_k \mid S ,\phi) \, \ud \phi} 
    = {\int p_\Phi(\phi \mid S)\, p(x_k \mid \phi) \, \ud \phi} 
    \neq  {\int p_\Phi(\phi)\, p(x_k \mid \phi) \, \ud \phi}
    = p(x_k).
$
That is, the conditional likelihood $p(x_k \mid S)$ depends on other points in $\calX$ via the posterior $p_\Phi(\phi \mid S)$, which accounts for what
$\phi$ was
%set generative model, $p(\cdot \mid \phi)$, were 
likely to have generated $S$. 
As a consequence, the complete generative process \eqref{eq:gen_proc_sets} is \emph{not marginally} \tiid, notwithstanding the \emph{conditional} \tiid ${p(\calX \mid \phi)}$. Thus, any model built on an \tiid assumption may be severely biased.

The generative process in Eq. \eqref{eq:gen_proc_sets} is especially applicable for surface point cloud data.
For such sets, $\calX_i$, points are drawn \tiid from (conditioned on) the surface of a shape with (\emph{unknown}) parameters $\phi_i$ (e.g.~object class, length, orientation, noise, etc.), resulting in the dataset $\calD = \{ \calX_i \sim {p(\cdot\mid \phi_i)} \}_{i=1}^N$ of $N$ sets.
As shown above, modeling such point cloud set data requires a \emph{non}-\tiid approach even though points may be drawn independently given the surface parameters.
\method will not only yield an exchangeable, \emph{non}-\tiid generative model, but will also 
directly model elements in sets without latent parameters.
In effect, \method will automatically marginalize out dependence on latent parameters of a given set, and is thus capable of handling  complicated $p(\cdot\mid \phi)$.
%As our goal in exchangeable density estimation is to capture the probabilistic variation in $\mathcal{D}$, one can think of this task as trying to produce a generative model over generative models.

% One approach to estimating \eqref{eq:gen_proc_sets} without \tiid assumptions is to estimate the latent parameters $\phi$, using variational techniques \cite{iclr:edwards}.
% While this approach can deliver high quality samples in some settings, it can only yield a lower bound for likelihoods and introduces the added task of encoding sets, which is an active research area itself.
% Instead, we effectively marginalize out any dependence on latent set parameters to compute a likelihood over observed point sets $\bx$ directly.
% Empirical results (detailed in the Experiments section) also indicate that estimating the marginal set likelihood exactly yields better performance in terms of likelihoods and samples.

Broadly, the primary challenge in direct exchangeable density estimation is designing a flexible, invariant architecture which yields a valid likelihood.
As explained above, using an \tiid assumption to enforce this property will severely hamper the performance of a model.
To avoid this simplification, techniques often shoehorn invariances to observed orderings by feeding randomly permuted data into sequential models \cite{rezatofighi2017deepsetnet,you2018graphrnn}.
Such approaches attempt to average out the likelihood of the model over all permutations:
\begin{align} \label{eq:permavg}
p(\calX) = \frac{1}{n!} \sum_{\pi} p_s(x_{\pi_1}, \ldots, x_{\pi_n}),
\end{align}
where $p_s$ is some sequential model.
Of course, the observation of all potential orderings for even a modest collection of points is infeasible.
Furthermore, there are often no guarantees that the sequential model $p_\mathrm{seq}$ will learn to ignore orderings,
especially for unseen test data \cite{vinyals2015order}.

%Then, given that an \tiid assumption is not a viable option for inducing exchangeability, what operation should be used to ensure permutation invariance of the architecture?
Given that an \tiid assumption is not robust 
and averaging over all permutations is infeasible, what operation should be used to ensure permutation invariance of the architecture?
Instead of attempting to wash out the effect of order in an architecture as in \refeq{eq:permavg}, we propose to enforce invariance by adopting a prespecified ordering and scanning over elements in this order.
As will be discussed in the Methods section, the benefit of estimating a likelihood over sorted data is that it frees us from the restriction of exchangeability.
Given the sorted data, we can apply any number of traditional density estimators.
However, such an approach presents its own challenges:
%some challenges of its own:
\begin{itemize}[leftmargin=*]
    \item \textbf{Determining a suitable way to scan through an exchangeable sequence.} That is, one must map the set $\calX = \{x_j\}_{j=1}^n$ to a sequence $\calX \mapsto (x_{[1]}, \ldots, x_{[n]})$ where $x_{[j]}$ denotes the $j$'th point in the sorted order.
    \item \textbf{Relating the likelihood of the scanned sequence to likelihood of the exchangeable set.} Modeling the exchangeable likelihood through a scanned likelihood is not immediately obvious; \emph{a simple equality of the two does not hold}, $p(\calX) \neq p(x_{[1]}, \ldots, x_{[n]})$.
    \item \textbf{Scanning in a space that is beneficial for modeling.} The native input space may not be best suited for modeling or scanning, hence it would be constructive to transform the exchangeable input prior to the scan.
    \item \textbf{Developing an architecture that exploits the structure gained in the scan.} The scanning operation will introduce sequential correlations among elements which need to be modeled successfully.
\end{itemize}

Next, we develop the \method model while addressing each of these challenges.

\section{Methods}
\begin{figure*}[t]
    \centering
    \includegraphics[scale=0.6]{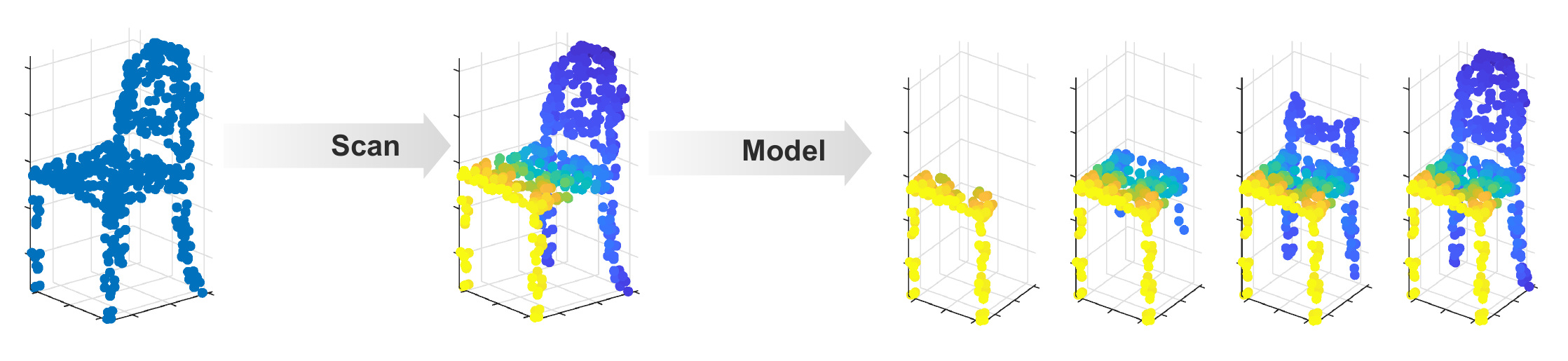}
    \caption{Illustration of our proposed method. First, input sets are scanned (in a possibly transformed space). After, the scanned covariates are modeled (possibly in a autoregressive fashion, as shown).}
    \label{fig:scan_model}
\end{figure*}

\method consists of three components: 1) a sequence of equivariant flow transformations ($\hat{q}_e$), \emph{to map the data to a space that is easier to model}; 2) a sort with correction factor \emph{to allow for the use of non-exchangeable density estimators}; 3) a density estimator ($\hat{p}_s$) (e.g.~an autoregressive model which may utilize sequential flow transformations, $\hat{q}_c$), \emph{to estimate the likelihood while accounting for correlations induced by sorting} (see \reffig{fig:scan_model}).
In this section, we motivate each piece of the architecture and detail how they
%can be combined 
combine
to yield a highly flexible, exchangeable density estimator.

\subsection{Equivariant Flow Transformations}
\method first utilizes a sequence of equivariant flow transformations.
So-called ``flow models'' rely on the change of variables formula to build highly effective models for traditional non-exchangeable generative tasks (like image modeling) \cite{kingma2018glow}.
Using the change of variables formula, flow models approximate the likelihood of a $d$-dimensional distribution over real-valued covariates $x = (x^{(1)}, ..., x^{(d)}) \in \R^d$, by applying an invertible (flow) transformation $\hat{q}(x)$ to an estimated base distribution $\hat{f}$:
\begin{align}
    \hat{p}(x^{(1)}, ..., x^{(d)}) = \bigg|\mbox{det}\frac{\ud \hat{q}}{\ud x}\bigg| \hat{f}(\hat{q}(x)), \label{eq:TAN}
\end{align}
where $|\mbox{det}\frac{\ud \hat{q}}{\ud x}|$ is the Jacobian of the transformation $\hat{q}$.
Often, the base distribution is a standard Gaussian.
However, \cite{oliva2018transformation} recently showed that performance may be improved with a more flexible base distribution on transformed covariates such as an autoreggressive density \cite{pmlr-v37-germain15,pmlr-v32-gregor14,pmlr-v15-larochelle11a,JMLR:uria,NIPS2013_uria}.

There are a myriad of possible invertible transformations, $\hat{q}$, that one may apply to inputs $\bx \in \R^{n \times d}$ in order to model elements in a more expressive space.
However, in our case one must take care to preserve exchangeability of the inputs when transforming the data.
%For instance, naively applying an autoregressive change of variables on points will be sensitive to the order in which the elements in $\bx$ were observed and will result in a space that is no longer exchangeable.
For example, a simple affine change of variables will be sensitive to the order in which the elements of $\bx$ were observed, resulting in a space which is no longer exchangeable.
One can circumvent this problem by requiring that any transformation, $\hat{q}$, used is \emph{equivariant}.
That is, for all permutation operators, $\Gamma$, we have that
$
    \hat{q}(\Gamma \bx) = \Gamma \hat{q}(\bx) % \label{eq:permeq}
$.
Proposition \ref{prop:eq_trans} states that equivariance of the transformations in conjunction with invariance of the base distribution is enough to ensure that exchangeability is preserved, \emph{allowing one to model set data in a transformed space}. 
The proof is straightforward and relegated to the Appendix.

\setcounter{section}{1}
\setcounter{theorem}{1}
\begin{prop}\label{prop:eq_trans}
Let $\hat{q}: \R^{n \times d} \mapsto \R^{n \times d}$ be a \emph{permutation equivariant}, invertible transformation and the base distribution, $\hat{f}$, be exchangeable.
Then the likelihood, $\hat{p}(\bx) = \big| \emph{det} \frac{\ud \hat{q}}{\ud \bx}\big| \hat{f}(\hat{q}(\bx))$, is exchangeable.
\end{prop}

Given an invertible transformation, $q: \mathbb{R}^d \rightarrow \mathbb{R}^d$, one may construct a simple permutation equivariant transformation by applying it to each point in a set independently:
$(x_1, ..., x_n) \mapsto (q(x_1), ..., q(x_n))$.
However, it is possible to engineer equivariant transformations which utilize information from other points in the set while still preserving equivariance.
Proposition \ref{prop:eq_trans} shows that \method is compatible with any combination of these transformations.

\begin{figure}[h!]
\centering
\includegraphics[width=0.98\linewidth]{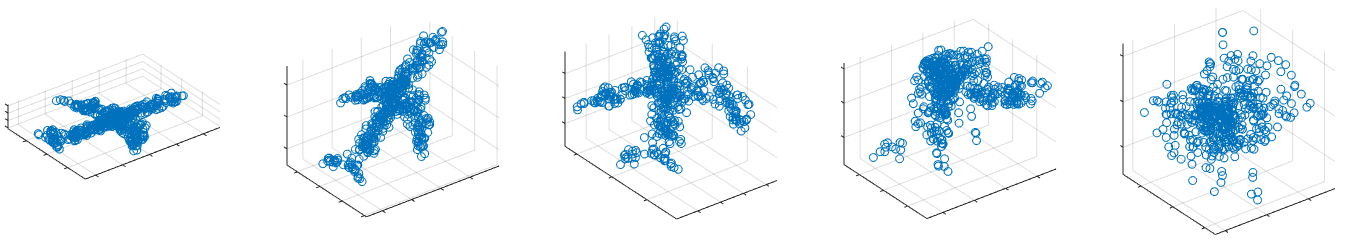}
\caption{An illustration of how set-coupling transformations act on a set. The first plot shows the input data to be transformed. In the subsequent plots, the set is transformed in an invertible, equivariant fashion by stacking set-coupling transformations. Iteratively transforming dimensions of a set in this way yields a set with simpler structure that may be modeled more easily, as shown in the last plot.}
\label{fig:setnvp_graphic}
\end{figure}

\paragraph{Set-Coupling} Among others, we propose a novel set-level scaling and shifting coupling transformation \cite{DBLP:journals/corr/DinhSB16}.
For $d$-dimensional points, the coupling transformation scales and shifts one subset, $S \subset \{1, \ldots, d\}$ of the $d$ covariates given then rest, $S^c$ as (letting superscripts index point dimensions):
\begin{align}
    \xind{S} &\mapsto \exp\left(f\left(\xind{S^c}\right)\right) \cdot \xind{S} + g\left(\xind{S^c}\right)  \notag \\
    \xind{S^c} &\mapsto \xind{S^c} \label{eq:nvp_coup},
\end{align}
for learned functions $f, g: \R^{|S^c|} \mapsto \R^{|S|}$. 
We propose a set-coupling transformation as follows:
% \begin{align}
%     \pind{x_i}{S} \mapsto & \  \exp {\scriptstyle\left(  f \left(\varphi(\pind{\bx}{S^c}), \pind{x_i}{ S^c}\right)\right)} \cdot \pind{x_i}{S} 
%     + {\scriptstyle g\left(\varphi(\pind{\bx}{S^c}), \pind{x_i}{S^c}\right)}  \notag \\
%     \pind{x_i}{S^c} \mapsto & \,\pind{x_i}{S^c},
% \end{align}
% \begin{align}
%     \pind{x_i}{S} \mapsto &  \exp \resizebox{0.3\columnwidth}!{\left(  f \left(\varphi(\pind{\bx}{S^c}), \pind{x_i}{ S^c}\right)\right)} \cdot \pind{x_i}{S} + \resizebox{0.3\columnwidth}!{ g\left(\varphi(\pind{\bx}{S^c}), \pind{x_i}{S^c}\right)}  \notag \\
%     \pind{x_i}{S^c} \mapsto & \,\pind{x_i}{S^c},
% \end{align}
\begin{align}
    &\resizebox{0.95\columnwidth}{!}{$\pind{x_i}{S} \mapsto   \exp \left(  f \left(\varphi(\pind{\bx}{S^c}), \pind{x_i}{ S^c}\right)\right) \cdot \pind{x_i}{S} +  g\left(\varphi(\pind{\bx}{S^c}), \pind{x_i}{S^c}\right)$}  \notag \\
    &\pind{x_i}{S^c} \mapsto  \,\pind{x_i}{S^c},
\end{align}
% \begin{align}
%     \pind{x_i}{S} \mapsto & \  \exp \left(  f \left(\varphi(\pind{\bx}{S^c}), \pind{x_i}{ S^c}\right)\right) \cdot \pind{x_i}{S} \\
%     &+  g\left(\varphi(\pind{\bx}{S^c}), \pind{x_i}{S^c}\right)  \notag \\
%     \pind{x_i}{S^c} \mapsto & \,\pind{x_i}{S^c},
% \end{align}
where $\pind{\bx}{S^c} \in \R^{n \times |S^c|}$ is the set of unchanged covariates, $\varphi(\pind{\bx}{S^c})\in \R^r$ are learnable permutation invariant features (using an architecture like \cite{zaheer2017deep}),
%which may be fixed (using statistics such as mean, variance, etc.) or learned (using an architecture like \cite{zaheer2017deep}), 
and $f, g: \mathbb{R}^{r+|S^c|} \rightarrow \mathbb{R}^{|S|}$ are learned functions.
The embedding $\varphi$ is responsible for capturing the set-level information from other covariates.
This is then combined with each point $\pind{x_i}{S^c}$ to yield shifts and scales with both point- and set-level dependence (see~\reffig{fig:setnvp_graphic}).
%In this work, we take $|S|=1$, transforming one dimension based on the rest, and composing in a round robin fashion. 
%Fig. \ref{fig:setnvp_graphic} gives a visual illustration of how a set-NVP transformation acts on a point cloud.
The log-determinant and inverse are detailed in the Appendix along with several other examples of flexible, equivariant transformations.

\subsection{Invariance Through Sorting}\label{sec:scan_invariance}
After applying a series of equivariant flow transformations, \method performs a sort operation and corrects the likelihood with a factor of $1/n!$.
Sorting in a prespecified fashion ensures that different permutations of the input map to the same output.
%We correct the likelihood with a factor of $1/n!$ in order to account for the $n!$ elements of the sample space which are assigned the same output.
In this section, we prove that this yields an analytically correct likelihood and comment on the advantages of such an approach.
Specifically, we show that the exchangeable (unordered) likelihood of a set of $n$ points $p_e(x_1, \ldots x_n)$ (where $x_j \in \R^d$) can be written in terms of the non-exchangeable (ordered) likelihood of the points in a sorted order $p_s(x_{[1]}, \ldots, x_{[n]})$ as stated in Prop.~\ref{prop:exch_lkhd_vs_sort_lkhd} below.
\begin{prop}\label{prop:exch_lkhd_vs_sort_lkhd}
Let $p_e$ be an exchangeable likelihood which is continuous and non-degenerate (e.g.~$\forall j\in\{1, \ldots, d\}$ $\Pr[x^{(j)}_1 \neq x^{(j)}_2 \neq \ldots \neq x^{(j)}_n] = 1$).
Then,
\begin{align}
p_e(x_1, \ldots x_n) = \frac{1}{n!} p_s(x_{[1]}, \ldots, x_{[n]}), \label{eq:sort_like}
\end{align}
where $x_{[j]}$ is the $j$th point in the sorted order.
\begin{proof}
We derive \refeq{eq:sort_like} from a variant of the change of variables formula \cite{casella2002statistical}.
It states that if we have a partition of our input space, $\{\calA_j\}_{j=1}^M$, such that a transformation of variables $q$ is invertible in each partition $\calA_j$ with inverse $q^{-1}_j$, then we may write the likelihood $f$ of $z = q(u)$ in terms of the likelihood $p$ of the input data $u$ as:
\begin{align}
    f(z) = \sum_{j=1}^M \Abs{\det \frac{\ud q^{-1}_j}{\ud z}} p(q^{-1}_j(z)) \label{eq:changevars}.
\end{align}

For the moment, suppose that the points $\{x_j\}_{j=1}^n$ are sorted according to the first dimension. 
That is, $x_{[1]}, \ldots, x_{[n]}$ in \refeq{eq:sort_like} are such that $x_{[1]}^{(1)} < \ldots < x_{[n]}^{(1)}$.
The act of sorting these points amounts to a transformation of variables $s: \R^{n \times d} \mapsto \R^{n \times d}$, $s(x_1, \ldots, x_n) = (x_{[1]}, \ldots, x_{[n]})$.
The transformation $s$ is one-to-one on the partitions of the input space $\R^{n \times d}$ defined by the relative order of points.
In other words, we may partition the input space according to the permutation that would sort the data: $\calA_{\pi} = \{\bx \in \R^{n \times d} \mid x_{\pi_1}^{(1)} < x_{\pi_2}^{(1)} < \ldots < x_{\pi_n}^{(1)} \}$.
We may invert $s$ in $\calA_{\pi}$ via the inverse permutation matrix of $\pi$, $\Gamma^{-1}_\pi$.
Letting $\Pi$ be the set of all permutations, \refeq{eq:changevars} yields:
\begin{align}
    p_s(s(\bx)) \overset{*}{=} \sum_{\pi \in \Pi} \Abs{\Gamma^{-1}_\pi} p_e(\Gamma^{-1}_\pi s(\bx)) \overset{**}{=}  n!\, p_e(\bx),
\end{align}
where (*) follows from \refeq{eq:changevars} and (**) follows from the exchangeability of $p_e$. 
Thus, we may compute the exchangeable likelihood $p_e(\bx)$ using the likelihood of the sorted points, as in \refeq{eq:sort_like}.
Trivially, similar arguments also hold when sorting according to a dimension other than the first.
Furthermore, it is possible to sort according any appropriately transformed space of $x_j$, rather than any native dimension itself (as this is equivalent to applying a transformation, sorting, and inverting said transformation).
\end{proof}
\end{prop}

Consequently, the exchangeable likelihood may be estimated via an approximation of the scanned covariates: 
$p_e(\bx) \approx  \frac{1}{n!}\, \hat{p}_s(s(\bx))$.
% \begin{align}
%     p_e(\bx) \approx  \frac{1}{n!}\, \hat{p}_s(s(\bx)) \label{eq:ordapprox1d}.
% \end{align}
Since the density of sorted scan is not exchangeable, we may estimate $\hat{p}_s$ using traditional density estimation techniques.
\emph{This gives a principled approach to reduce the problem of exchangeable likelihood estimation to a flat vector (or sequence) likelihood estimation task.}
%%Furthermore, it is possible to sort in any appropriately transformed space of $x_j$, rather than any native dimension itself.
% Since the scanned likelihood in Eq. \eqref{eq:sort_like} yields an exchangeable likelihood, one may use as the base likelihood following a permutation equivariant transformation as in Prop. \ref{prop:eq_tran}.
% This enables us to apply the sorting step after performing any number of equivariant transformations and improve the flexibility of the model as a result.
% As no generality is lost, we do choose to sort on the first dimension in our experiments detailed below.
% It is important to note that while we can propagate gradients through the sorting operation, upstream layers cannot adjust their behavior to better inform the sort.

\subsection{Autoregressive Scan Likelihood}
After performing equivariant flow transformations and sorting, \method applies a non-exchangeable density estimator to model the transformed and sorted data.
Let $z = s(\hat{q}(\bx)) \in \R^{n \times d}$ be the sorted covariates.
Since $z$ is not exchangeable, one can apply any traditional likelihood estimator on its covariates, e.g.~one may treat $z$ as a vector and model $\hat{p}_s(\mathrm{vec}(z))$ using a flat density estimator.
However, flattening in this way suffers from several disadvantages.
First, it is inflexible to varying cardinalities.
Furthermore, the total number of covariates, $nd$, may be large for sets with large cardinality or dimensionality.
Finally, a general flat model loses the context that covariates are from multiple points in some shared set.
To address these challenges, we use an autoregressive likelihood:
%$ \hat{p}(z_k) = \prod_{k=1}^n \hat{p}(z_k \mid h(z_1, \ldots, z_{k-1})) = \prod_{k=1}^n \hat{p}(z_k \mid h_{<k})$,
 \begin{align}
     \hat{p}(z_k) = \prod_{k=1}^n \hat{p}(z_k \mid h_{<k}), \label{eq:autoreg_fs}
 \end{align}
where $\hat{p}(z_k \mid h_{<k})$ is itself a $d$-dimensional density estimator (such as \refeq{eq:TAN}) conditioned on a recurrent state $h_{<k} = h(z_1, \ldots, z_{k-1})$. This proposed approach is capable of sharing parameters across the $n$ $d$-dimensional likelihoods %estimation tasks 
and is more amenable to large, possibly varying, cardinalities.

\paragraph{Correspondence Flow Transformations}
In much the same way that nearby pixels are correlated in image space, nearby points will be correlated in a scan space.
Thus, we also propose a coupling \cite{Dinh2014NICENI} invertible transformation to transform adjacent points, exploiting existing correlations among points as follows.
We note that it is straightforward to use a sequential coupling transformation to shift and scale points $z_i$ as in \refeq{eq:nvp_coup}, but based on inputting a recurrent output $h_{<i}$ to $f$ and $g$ functions.
In addition, it is also possible to split individual points for coupling as follows.
First, split the scanned points $z = s(\hat{q}(\bx)) = (z_1, \ldots, z_n)$ into two groups depending on the parity (even/odd) of their respective index. Second, transform each even point, with a scale and shift based on the corresponding odd point. That is for pairs of points $(z_{2j}, z_{2j+1})$ we perform the following transformation:
$(z_{2j}, z_{2j+1}) \mapsto \left(s(z_{2j+1}) z_{2j} + m(z_{2j+1}),\, z_{2j+1}\right)$,
% \begin{align}
%     (z_{2j}, z_{2j+1}) \mapsto \left(s(z_{2j+1}) z_{2j} + m(z_{2j+1}),\, z_{2j+1}\right), \label{eq:corr_flow}
% \end{align}
where $s: \R^d \mapsto \R^d, m: \R^d \mapsto \R^d$ are scale and shifting functions, respectively, parameterized by a learnable fully connected network.
This correspondence coupling transformation $z \mapsto z^\prime$ is easily invertible and has analytical Jacobian determinant $\big|\mbox{det}\frac{\ud z^\prime}{\ud z}\big| = \prod_{j=0}^{n/2-1} |s(z_{2j+1})|$.
Several of these transformations may be stacked before the autoregressive likelihood 
% TODO: \eqref{eq:autoreg_fs} 
by alternating between shifting and scaling even points based on odd and vice-versa odd points based on even.
%Let $\hat{q}_\mathrm{c} = \hat{q}_\mathrm{odd} \circ \hat{q}_\mathrm{even} \circ \ldots \circ \hat{q}_\mathrm{odd} \circ \hat{q}_\mathrm{even}$ be the composition of the correspondence coupling transformations, and let $z_c = \hat{q}_\mathrm{c}\left(s(\hat{q}_\mathrm{e}(\bx))\right)$ be the resulting covariates from corresponding coupling transforming the flow scanned covariates, $s(\hat{q}_\mathrm{e}(\bx))$.
We shall also make use of a similar splitting scheme to split sets of images into 3d tensors that are fed into 3d convolution networks for shifting and scaling.
% We note that, like the first set of equivariant transformations, \method is compatible with any general sequential flow transformations.
% While our experiments suggested that correspondence couplings yielded the best results, a deeper exploration of new sequential flow transformations for \method is an interesting direction for future research.

\subsection{Complete \method Architecture}
Since the scanned likelihood in \refeq{eq:sort_like} yields an exchangeable likelihood, one may use as the base likelihood following a permutation equivariant transformation as in Prop.~\ref{prop:eq_trans}.
This enables us to apply the sorting step after performing any number of equivariant transformations and improve the flexibility of the model as a result.
As no generality is lost, we choose to sort on the first dimension in our experiments detailed below.
% It is important to note that while we can propagate gradients through the sorting operation, upstream layers cannot adjust their behavior to better inform the sort.
Combining the three components detailed above, we arrive at the complete \method architecture: a sequence of equivariant flow transformations, a sort with correction factor, and an autoregressive scan likelihood.
The estimated exchangeable likelihood that results is:
\begin{align}
    \hat{p}_\mathrm{fs}(\bx) = \frac{1}{n!}\, \bigg|\mbox{det}\frac{\ud \hat{q}_\mathrm{e}}{\ud x}\bigg| \hat{p}_s(s(\hat{q}_\mathrm{e}(\bx))), \label{eq:flowscan}
\end{align}
where $\hat{q}_\mathrm{e}$ and $\hat{p}_s$ are the estimated (via maximum likelihood) equivariant flow transformation and sorted flow scan covariate likelihood, respectively.
When correspondence flow transformations are included after the sort operation, we obtain an estimated exchangeable likelihood:
%\begin{align}
%    \hat{p}_\mathrm{fs}(\bx) = \frac{1}{n!}\, \bigg|\mbox{det}\frac{\ud \hat{q}_\mathrm{e}}{\ud x}\bigg| \, \bigg|\mbox{det}\frac{\ud \hat{q}_\mathrm{c}}{\ud x}\bigg| \, \prod_{k=1}^n \hat{p}(z_{c\, k} \mid h(z_{c\, 1}, \ldots, z_{c\, k-1})). \label{eq:flowscan_ultimate}
%\end{align}
\begin{align}
    \hat{p}_\mathrm{fs}(\bx) = \frac{1}{n!} \bigg|\mbox{det}\frac{\ud \hat{q}_\mathrm{e}}{\ud x}\bigg|  \bigg|\mbox{det}\frac{\ud \hat{q}_\mathrm{c}}{\ud x}\bigg| \prod_{k=1}^n \hat{p}(\mathbf{z}_{k} \mid h(\mathbf{z}_{<k})), \label{eq:flowscan_ultimate}
\end{align}
%\begin{align}
%    \hat{p}_\mathrm{fs}(\bx) = \frac{1}{n!}\, \bigg|\mbox{det}\frac{\ud \hat{q}_\mathrm{e}}{\ud x}\bigg| \, \bigg|\mbox{det}\frac{\ud \hat{q}_\mathrm{c}}{\ud x}\bigg| \, \prod_{k=1}^n \hat{p}(z_{c\, k} \mid h(z_{c,\, \<k})), 
%    \label{eq:flowscan_ultimate}
%\end{align}
%where $z_{c\, 1:(k-1)} = (z_{c\, 1}, \ldots, z_{c\, k-1})$. 
where $\mathbf{z}$ is the resulting covariates from corresponding coupling transforming the flow scanned covariates.
In both cases, \method gives a valid, provably exchangeable density estimate relying neither on variational lower bounds of the likelihood nor averaging over all possible permutations of the inputs.
Furthermore, \method is easily adapted to input sets with varying cardinalities, as is commonly observed in practice.
In the Experiments section, we demonstrate empirically that \method is highly flexible and capable of modeling sets of both points clouds and images.

\section{Related Work}
Unlike the recent surge in flexible density estimation for flat vectors with deep architectures \cite{Dinh2014NICENI,DBLP:journals/corr/DinhSB16,kingma2018glow,pmlr-v15-larochelle11a,NIPS2013_uria,JMLR:uria,pmlr-v32-gregor14,pmlr-v37-germain15,oliva2018transformation}, exchangeable treatments of data in ML have been limited with some notable exceptions.
Some recent work \cite{lee2018set,qi2017pointnet,zaheer2017deep} has explored neural architectures for constructing a permutation invariant set embeddings.
They featurize input sets exchangeably in a way that is useful for (typically supervised) downstream tasks; \emph{but the embeddings themselves will not result in valid likelihoods}.
In other work, Generative Adversarial Networks (GAN) have been explored as a means of sampling point clouds \cite{zaheer2018pcgan}.
However, none of these methods provide a valid exchangeable likelihood estimate as is our focus.

A recently proposed model for exchangeable data, BRUNO \cite{korshunova2018bruno}, preserves exchangeability by performing independent point-wise changes of variables, a simple equivariant linear transformation, and an \tiid base exchangeable process in the latent space.
%BRUNO was shown to perform well in the modeling of exchangeable sequences of images and will be used as one of the baselines in our experiments.
The Neural Statistician (\vae) \cite{iclr:edwards} estimates a permutation invariant code produced by an exchangeable VAE.
That is, the Neural Statistician uses an encoder, called a statistics network, on the entire exchangeable set to get an approximate posterior on the latent code.
Given the success of a point cloud autoencoder with a DeepSet network as the statistics network in \cite{oliva2018transformation}, we consider this architecture for the variational Neural Statistician which is an especially strong baseline, representing the state-of-the-art likelihood method for point cloud data.

%% Experiments
\section{Experiments}\label{sec:experiments}
In this section, we compare the performance of \method to that of \bruno and \vae in a variety of exchangeable point cloud and image modeling tasks.
In each experiment, our goal is to estimate an exchangeable likelihood $p(\bx)$ for $\bx \in \R^{n \times d}$ which models the inputs well.
As is standard in density estimation tasks, we measure the success of the model via the estimated likelihood of a held out test set for each experiment.
For readability, we report the estimated per point log likelihoods (PPLL): $\frac{1}{n} \log\, \hat{p}(\bx)$.
As \vae does not yield a likelihood, we report its estimated variational lower bound on the PPLL. Results for each datasets can be found in \reftab{table:loglikes}.
%As \vae does not technically yield a likelihood, we report its estimated variational lower bound on the PPLL. Results across all models for each datasets can be found in \reftab{table:loglikes}.

As a qualitative assessment of each model's performance, we also include samples generated by each trained model.
Those which are not reported in the main text can be found in the Appendix.
Unless stated explicitly, the figures included are \emph{not reconstructions}, but completely synthetic point clouds or images generated by each model.
Further implementation details can be found in the Appendix and code will be made available at \texttt{https://github.com/lupalab/flowscan}.

\begin{table}[b!]
\centering
\footnotesize
 \begin{tabular}{c c c c c} 
 \specialrule{.1em}{.05em}{.05em}
 Dataset & \bruno & \vae & \method \\ 
 \specialrule{.1em}{.05em}{.05em}
    \texttt{Synthetic} & -2.28 & -1.07 & \textbf{0.14}\\
 \hline
  \texttt{Airplanes}   & 2.71 & 4.09 & \textbf{4.81} \\ 
  \texttt{Chairs}      & 0.75 & 2.02 & \textbf{2.58} \\
  \texttt{ModelNet10}  & 0.49 & 2.12 & \textbf{3.01} \\
  \texttt{ModelNet10a} & 1.20 & 2.82 & \textbf{3.58} \\
 \hline
 \texttt{Caudate} & 1.29 & 4.49 & \textbf{4.87} \\
 \texttt{Thalamus} & -0.815 & 2.69 & \textbf{3.12} \\
 \hline
 \texttt{SpatialMNIST} & -5.68 & -5.37 & \textbf{-5.26}\\
 \specialrule{.1em}{.05em}{.05em}
\end{tabular}
\caption{Per-point log-likelihood (PPLL) of the test set for all point cloud experiments. Higher PPLL indicates better modeling of the test set.}
\label{table:loglikes}
\end{table}

\subsection{Shuffled Synthetic Sequential Data}\label{sec:synth_data_section}
\begin{figure}[h]
    \centering
    \includegraphics[height=22mm]{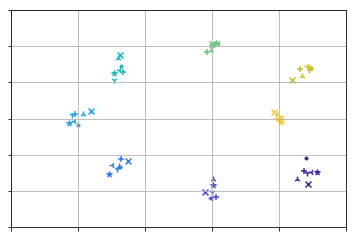}
    \includegraphics[height=22mm]{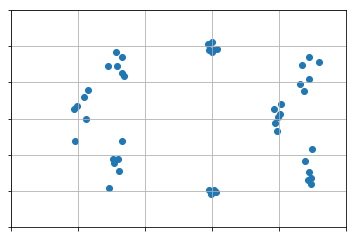}
    \caption{Left: true samples; markers and colors indicate instances and sequential order. Right: \method samples.}
    \label{fig:synthetic_samples}
\end{figure}

We begin with a synthetic point cloud experiment to test \method's ability to learn a known, ground truth likelihood.
To allow for complex interactions between points, we study a common scenario that leads to exchangeable data: sequential data with time marginalized out.
In other words, we suppose that all time-points $x_j \in \R^d$ of a sequence $(x_1, \ldots, x_n)$ are put into an unordered set $\{x_1, \ldots, x_n\}$.
Effectively, this yields observations of sequences in matrices that are randomly shuffled from the sequential order. 
Hence, exchangeable instances are $\bx = \Gamma_\pi \bx_s$, for permutations $\Gamma_\pi \in \R^{n \times n}$ (drawn uniformly at random) and sequential data $\bx_s = (x_1, \ldots, x_n) \in \R^{n \times d}$ (drawn via a sequential likelihood $p_\mathrm{seq}$).
Here we consider a synthetic ground truth sequential model $p_\mathrm{seq}$ where the likelihood of an instance is computed by marginalizing out the permutation: $p(\bx) = \sum_{\pi^\prime} \Pr(\pi = \pi^\prime)\, p_\mathrm{seq}(\Gamma^{-1}_{\pi^\prime} \bx)  = \frac{1}{n!} \sum_{\pi} p_\mathrm{seq}(\Gamma_{\pi}\bx)$.
To obtain interesting non-linear dependencies we consider a sinusoidal sequence (see Fig.~\ref{fig:synthetic_samples} and Appendix for details).
To allow for computing the ground truth likelihood in a timely manner, we consider $n=8$, leading to a large number, $8! = 40320$, of summands in the likelihood of the data.

%\begin{wraptable}{l}{0.45\linewidth}
%\centering
%\caption{PPLLs on Synthetic Data}
%\footnotesize
% \begin{tabular}{c c c}
% \hline
% \bruno & \vae & \method \\ [0.25ex] 
% \hline
%  -2.28 & -1.07 &  \textbf{0.14} \\
%  [0.25ex] 
% \hline
%\end{tabular}
%\label{table:syn_loglikes}
%\normalsize
%\end{wraptable}
Table \ref{table:loglikes} illustrates the per point log likelihood (PPLL) estimates across the synthetic sets using \bruno, the \vae, and \method.
The \method model outperforms the other methods, achieving nearly the same PPLL as the ground truth ($0.23$) despite not averaging over all $n!$ permutations.
For further comparison, we also trained a sequential model on the randomly permuted instances (and marginalizing out the permutation as in \refeq{eq:permavg}).
However, randomly permuting the input sequence proved to be ineffective and resulted in low test PPLLs (with severe overfitting).

%\TODO{This ran successfully, but each time the training likelihood was much different than the test likelihood (2 orders of magnitude). Instead of saying that the number of permutations was too large, can we say that attempting to learn permutation invariance by randomly permuting the input sequence proved to be ineffective and resulted in extremely low test PPLLs?}

\subsection{ModelNet}\label{sec:modelnet}

\begin{figure}[t]
    \centering
    \begin{subfigure}[t]{.33\linewidth}
        \begin{minipage}{\linewidth}
            \centering
            \includegraphics[height=16mm]{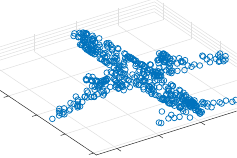}
            \includegraphics[height=16mm]{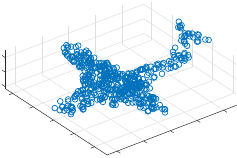} \\
            \includegraphics[height=16mm]{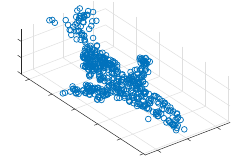}
            \includegraphics[height=16mm]{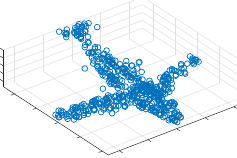}
        \end{minipage}
        \caption{\method}
    \end{subfigure}
    \begin{subfigure}[t]{.33\linewidth}
        \begin{minipage}{\linewidth}
            \centering
            \includegraphics[height=16mm]{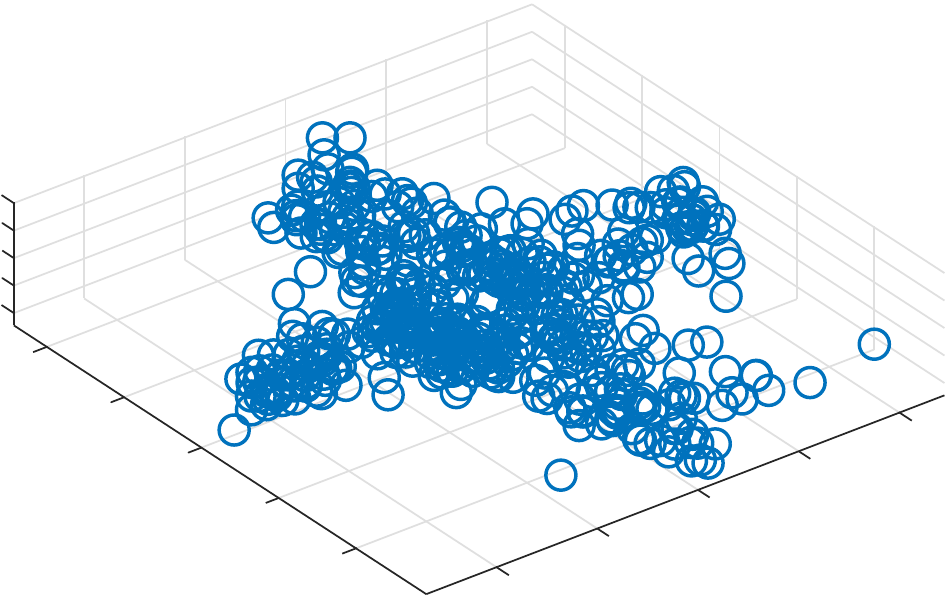}
            \includegraphics[height=16mm]{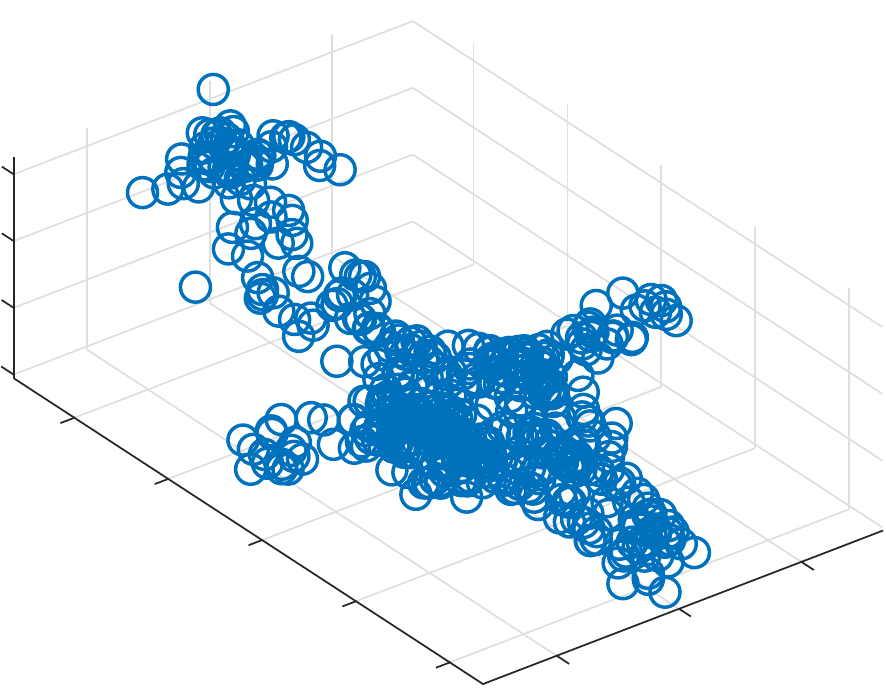} \\
            \includegraphics[height=16mm]{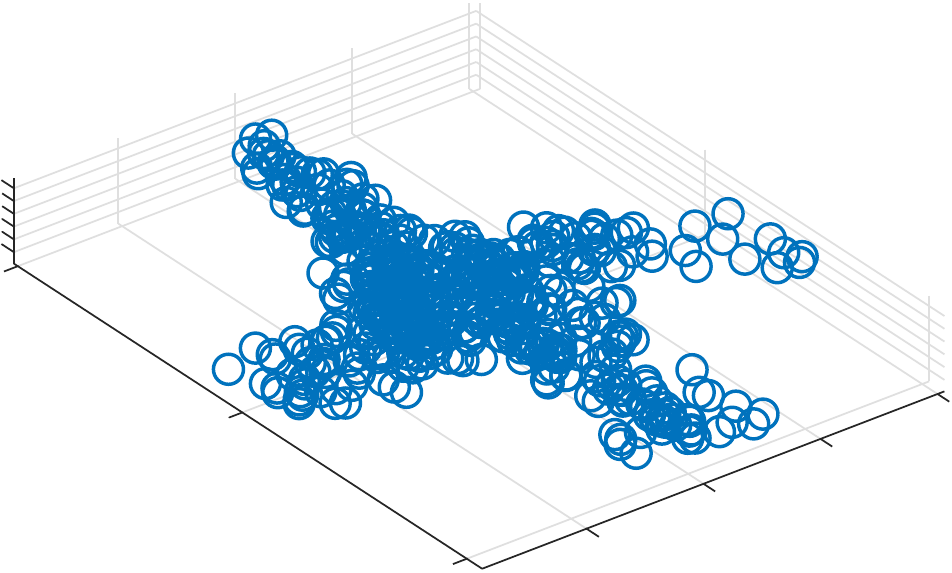}
            \includegraphics[height=16mm]{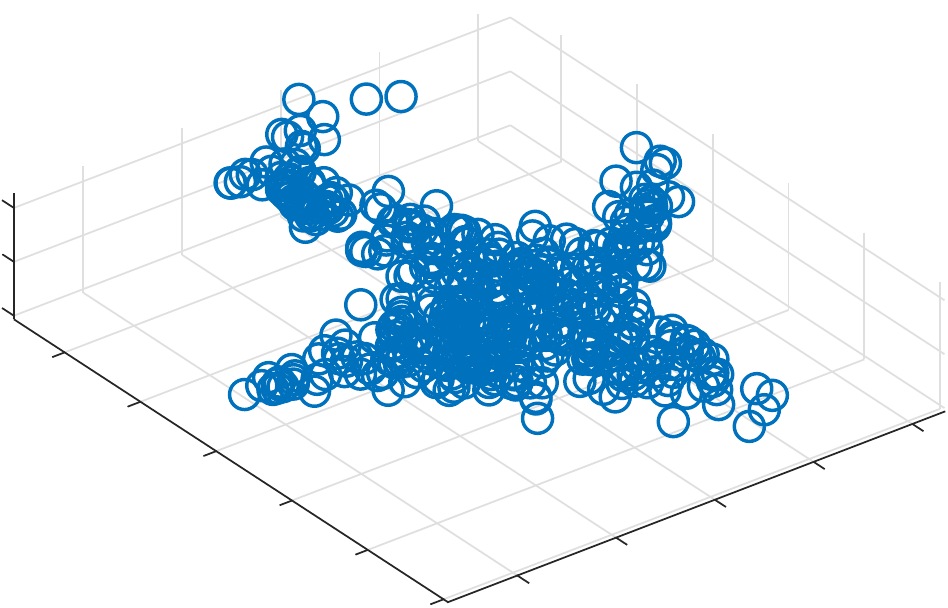}
        \end{minipage}
        \caption{\vae}
    \end{subfigure}
    \begin{subfigure}[t]{.32\linewidth}
        \begin{minipage}{\linewidth}
            \centering
            \includegraphics[height=16mm]{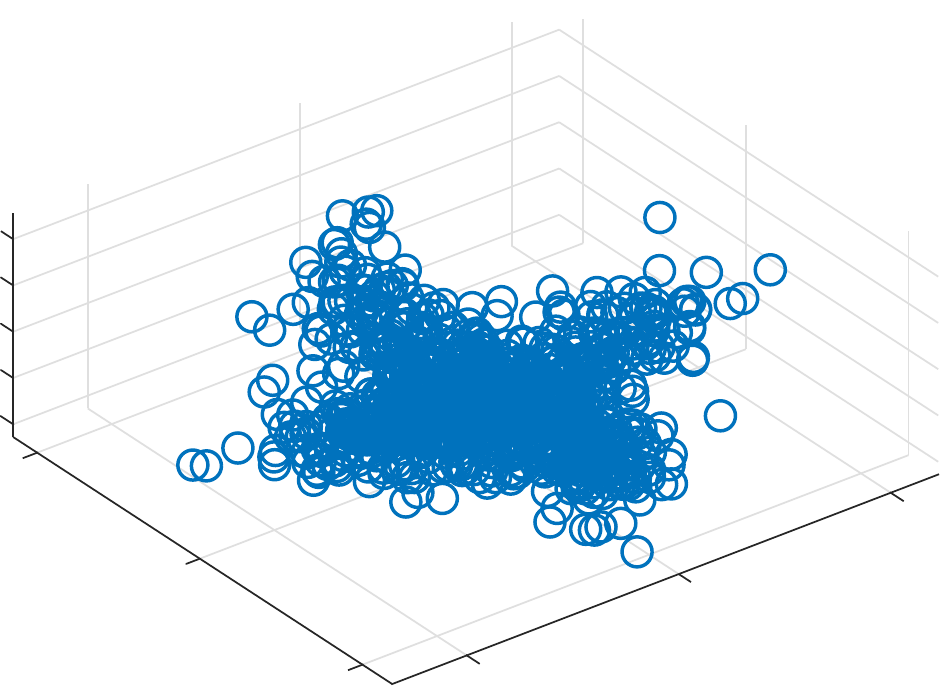}
            \includegraphics[height=16mm]{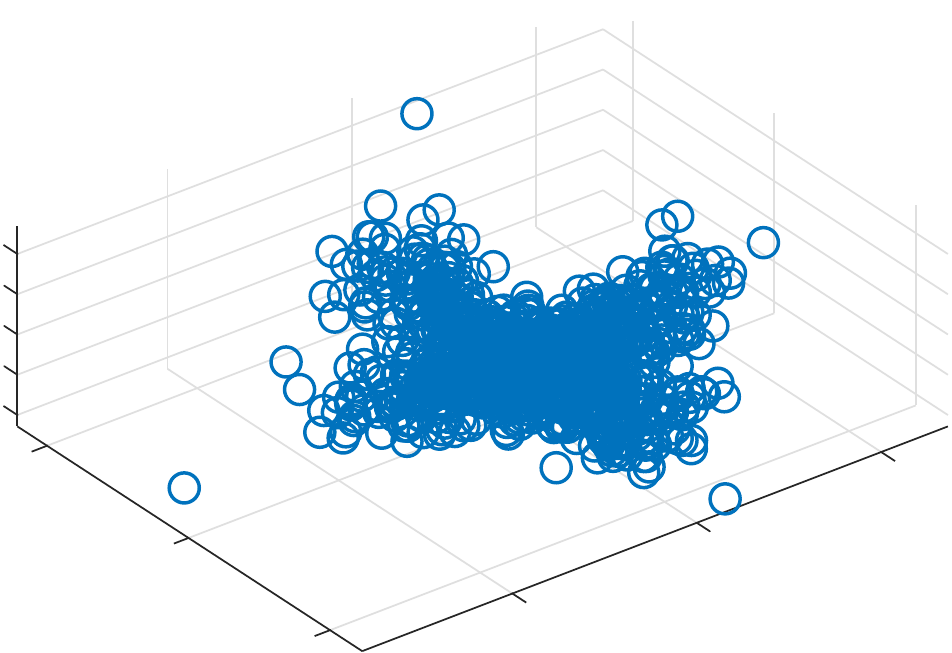} \\
            \includegraphics[height=16mm]{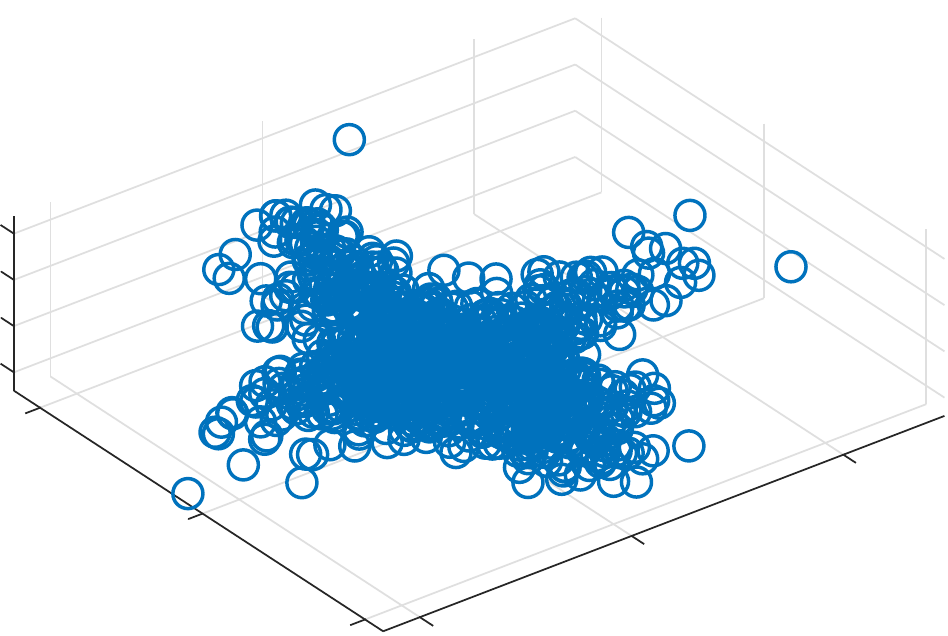}
            \includegraphics[height=16mm]{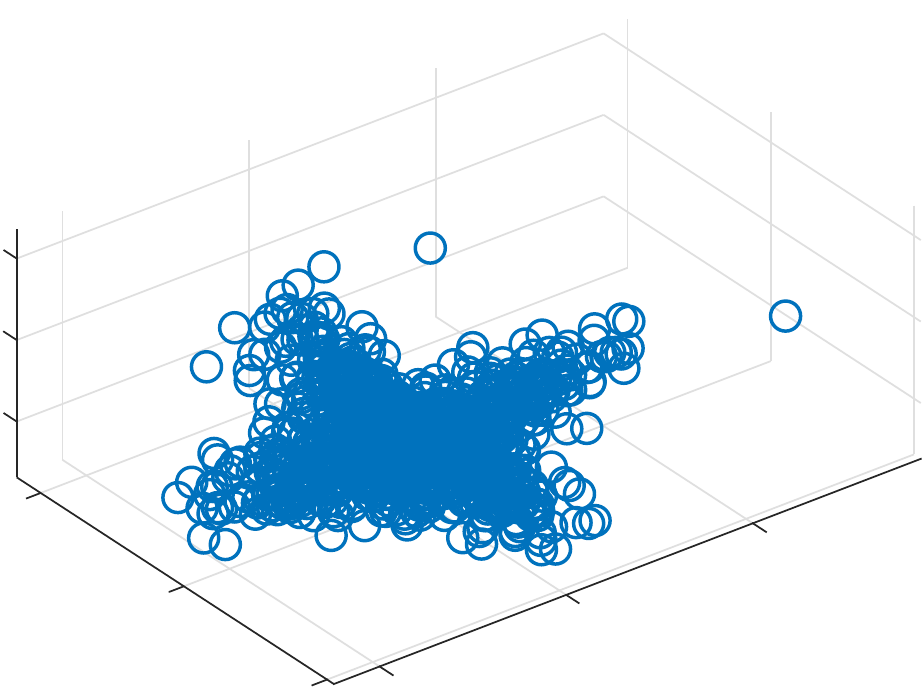}
        \end{minipage}
        \caption{BRUNO}
    \end{subfigure}
    \caption{Synthetic plane samples from trained models}
    \label{fig:plane}
\end{figure}

\begin{figure}[t]
    %\begin{minipage}{0.5\linewidth}
        \centering
        \includegraphics[height=18mm]{figures/samples/modelnet10/fs/modelnet10_1_crop}
        \includegraphics[height=16mm]{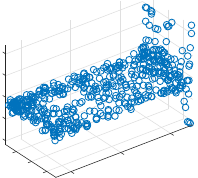}
        \includegraphics[height=18mm]{figures/samples/modelnet10/fs/modelnet10_3_crop}
        \includegraphics[height=18mm]{figures/samples/modelnet10/fs/modelnet10_4_crop}
    %\end{minipage}
    \caption{\method \texttt{ModelNet10} samples}
    \label{fig:modelnet10_samples}
\end{figure}
Next, we illustrate the efficacy of our model on real world point cloud data. 
We consider object classes from the ModelNet dataset \cite{wu20153d}, which contains CAD models of common real world objects. 
Point clouds were created by randomly sampling 512 points from the surface of each object. 
All point cloud sets are modeled in an unsupervised fashion. That is, we estimate $p(\bx)$, where $\bx \in \R^{512 \times 3}$.
Models are compared on the following datasets comprised of different subsets of point cloud classes: \texttt{airplanes}, \texttt{chairs}, \texttt{ModelNet10}, and \texttt{ModelNet10a}.
%Where \texttt{airplanes}, \texttt{chairs} are datasets containing only objects from class \emph{airplanes} and \emph{chairs}, respectively. 
\texttt{ModelNet10} is the standard subset \cite{wu20153d} consisting of \emph{bathtub},
\emph{bed},
\emph{chair},
\emph{desk},
\emph{dresser},
\emph{monitor},
\emph{night stand},
\emph{sofa},
\emph{table}, and
\emph{toilet} classes. 
Since \texttt{ModelNet10} is composed largely of furniture-like objects, we also select a more diverse, ten-class subset that we will refer to as \texttt{ModelNet10a}, containing \emph{airplane},
\emph{bed},
\emph{car},
\emph{chair},
\emph{guitar},
\emph{lamp},
\emph{laptop},
\emph{plant},
\emph{stairs}, and
\emph{table} classes.

Results can be found in Tab.~\ref{table:loglikes} and four samples from \method are included in Fig.~\ref{fig:modelnet10_samples}.
For each of the four datasets tested, we find that \method achieves the highest average test log-likelihood.
Qualitatively, we also observe superior samples from the \method model as can be seen in Fig.~\ref{fig:plane} and in the Appendix.
In addition to training on these ModelNet datasets, we also performed an ablation study (see the Appendix) where we see that our full architecture yields the best performance over alternatives.

\subsection{Brain Data}\label{sec:brain}
\begin{figure}[b]
    %\begin{minipage}{0.5\linewidth}
    \begin{subfigure}{0.48\linewidth}
        \centering
        \includegraphics[height=18mm]{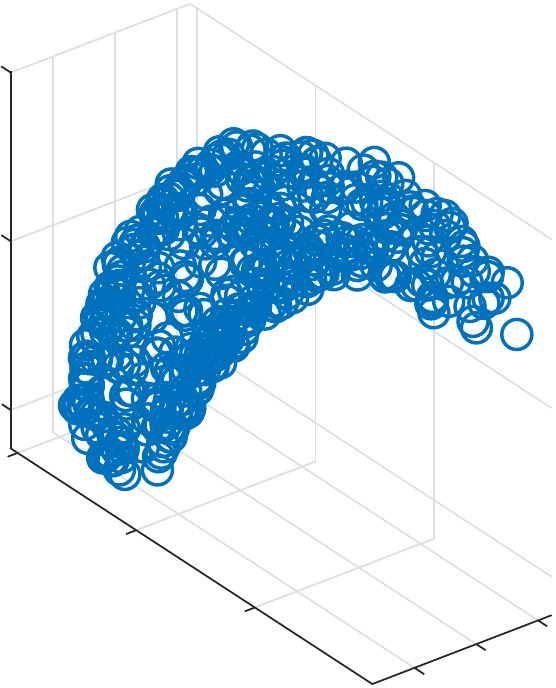}
        \includegraphics[height=18mm]{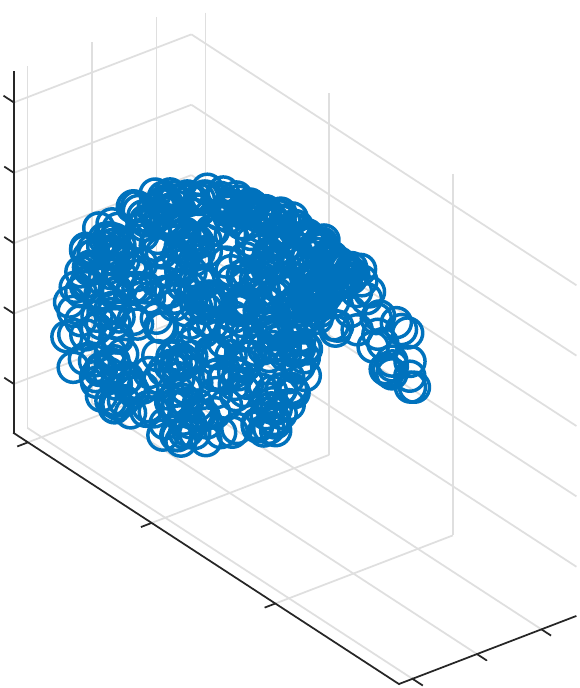}
        \caption{\texttt{Caudate}}
    \end{subfigure}
    \begin{subfigure}{0.48\linewidth}
        \centering
        \includegraphics[height=18mm]{figures/samples/thalamus/fs/sample_3}
        \includegraphics[height=18mm]{figures/samples/thalamus/fs/sample_4}
        \caption{\texttt{Thalamus}}
    \end{subfigure}
    %\end{minipage}
    \caption{\method \texttt{Caudate} and \texttt{Thalamus} samples}
    \label{fig:caud_thal_samples}
\end{figure}

We test \method's performance on a medical imaging task in a higher dimensional setting using samples of the Caudate and Thalamus \cite{nature2017Early}.
Each set contains 512 randomly sampled 7d points.
The first three dimensions contain the Cartesian coordinates of the surface boundary (as in ModelNet).
The next two dimensions represent the normal direction at the boundary in terms of angles.
The final two dimensions represent the local curvature (expressed as shape index and curvedness \cite{koenderink1990Solid}).
Table \ref{table:loglikes} enumerates the PPLL for both datasets across all three methods.
Comparing samples from \method (see \reffig{fig:caud_thal_samples}) to that of \vae and \bruno (included in the Appendix) we see that \method better captures the geometric features of the data than \vae.
Overall, superior PPLLs and samples suggest that \method seamlessly incorporates the additional geometric information to model point clouds more accurately than baseline methods.
    
%\begin{figure*}[t]
%    %\begin{minipage}{0.45\linewidth}
%    \begin{subfigure}{0.5\linewidth}
%        \centering
%        \includegraphics[height=25mm]{figures/samples/mnist/mnist_samples_test.png} \\
%        \includegraphics[height=25mm]{figures/samples/mnist/mnist_samples_test.png}
%        \caption{\texttt{MNIST}}
%        \label{fig:mnist_samples}
%    \end{subfigure}
%    %\end{minipage}
%    %\begin{minipage}{0.45\linewidth}
%    \begin{subfigure}{0.5\linewidth}
%        \centering
%        \includegraphics[height=25mm]{figures/samples/mnist/mnist_samples_test.png} \\
%        \includegraphics[height=25mm]{figures/samples/mnist/mnist_samples_test.png}
%        \caption{\texttt{Omniglot}}
%        \label{fig:omniglot_samples}
%    \end{subfigure}
%    %\end{minipage}
%    \caption{Samples from models trained on \texttt{MNIST} and \texttt{Omniglot}. Each row corresponds to a single set of 20 images generated by one model. The rows from top to bottom correspond to \bruno, \vae, and \method respectively.}
%    \label{fig:image_samples}
%\end{figure*}

\subsection{Spatial MNIST}
\begin{figure}
    %\begin{minipage}{0.5\linewidth}
    \centering
    \begin{subfigure}{\linewidth}
        \centering
        \includegraphics{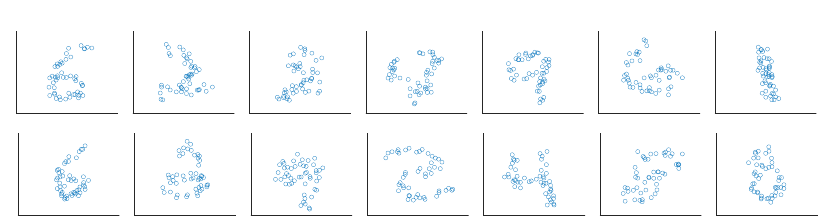}
        \caption{\method}
    \end{subfigure}
    \begin{subfigure}{\linewidth}
        \centering
        \includegraphics{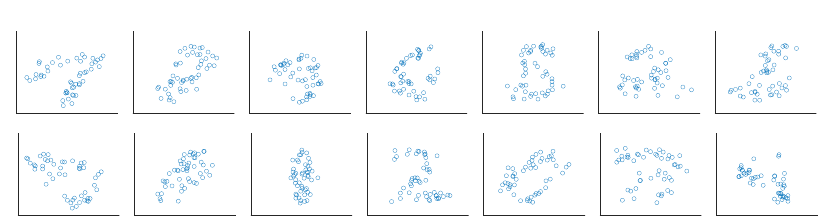}
        \caption{\vae}
    \end{subfigure}
    \begin{subfigure}{\linewidth}
        \centering
        \includegraphics{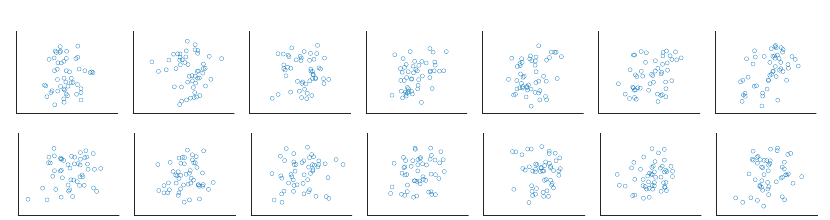}
        \caption{\bruno}
    \end{subfigure}
    %\end{minipage}
    \caption{\texttt{SpatialMNIST} samples from each model.}
    \label{fig:spatialmnist_samples}
\end{figure}

For a direct comparison to \vae, we also trained our model on the \texttt{SpatialMNIST} dataset, used by \cite{iclr:edwards}.
Each set consists of 50 points sampled uniformly at random from active pixels of a single \texttt{MNIST} \cite{lecun1998gradient} image with uniform noise added to ensure non-degeneracy.
The dataset that results consists of 2-dimensional point clouds each representing a digit from 0 to 9.
PPLLs for each model can be found in \reftab{table:loglikes} and a random selection of samples from each can be found in \reffig{fig:spatialmnist_samples}.
Both the (unconditioned) likelihoods and the samples indicate that \method gives superior performance in this task.

\subsection{MNIST}
\begin{figure}[t]
    \centering
    \begin{subfigure}{\linewidth}
        \centering  
        \includegraphics[height=4mm]{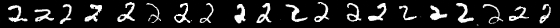} \\
        \includegraphics[height=4mm]{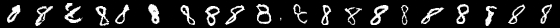} \\
        \includegraphics[height=4mm]{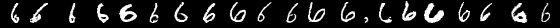} \\
        \includegraphics[height=4mm]{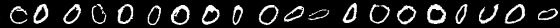} \\
        \includegraphics[height=4mm]{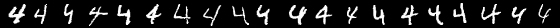} \\
        \includegraphics[height=4mm]{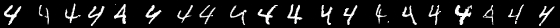} \\
        \caption{\method \label{fig:mnist_fs}}
    \end{subfigure}
    \begin{subfigure}{\linewidth}
    \centering
        \includegraphics[height=4mm]{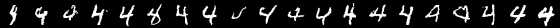}\\
        \includegraphics[height=4mm]{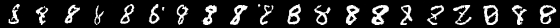}\\
        \includegraphics[height=4mm]{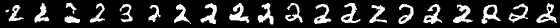}\\
        \includegraphics[height=4mm]{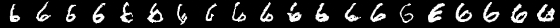}\\
        \includegraphics[height=4mm]{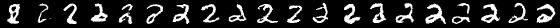}\\
        \includegraphics[height=4mm]{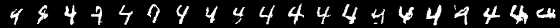}\\
        \caption{Set-Coupling \label{fig:mnist_sc}}
    \end{subfigure}
    \begin{subfigure}{\linewidth}
        \centering
        \includegraphics[height=4mm]{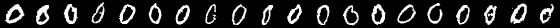} \\
        \includegraphics[height=4mm]{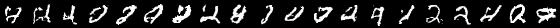} \\
        \includegraphics[height=4mm]{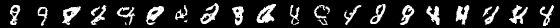} \\
        \includegraphics[height=4mm]{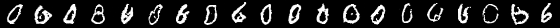} \\
        \includegraphics[height=4mm]{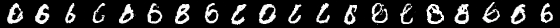} \\
        \includegraphics[height=4mm]{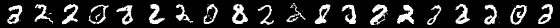} \\
        \caption{\bruno \label{fig:mnist_bruno}}
    \end{subfigure}
    \caption{Single digit set samples from \method, Set-Coupling, and \bruno trained on \texttt{MNIST}. Each row corresponds to a single set of 20 images generated by one model. %Sets generated from \bruno often contain unidentifiable or multiple digits whereas \method samples are relatively homogeneous and representative of the training set.
    }
    \label{fig:mnist_samples}
\end{figure}

Finally, we show that \method exhibits superior likelihoodsand samples in a high-dimensional, low-cardinality setting.
Following \cite{korshunova2018bruno}, sets are composed of 20 
%randomly-selected 
random
images corresponding to the \emph{same digit class} from the \texttt{MNIST} dataset.
%After training, PPLLs are evaluated over a similarly constructed, held out test set of image sets (test sets are constructed from unseen individual images).
After training, PPLLs are evaluated on held out test sets constructed from unseen images.
Our baseline is \bruno, which achieves a PPLL of $-643.6$. \bruno's unconditional samples (\reffig{fig:mnist_bruno}) often contain elements from different digits, indicating a lack of intra-set dependency in the resulting model.
We improve upon \bruno by first adding convolution-based Set-Coupling transformations (but keeping the \tiid base likelihood), which achieves a PPLL of $-634.8$.
Still, sample sets (\reffig{fig:mnist_sc}) show mixed digit classes.
Finally, we consider a full \method model that adds a sort, scan, and 3d convolution-based correspondence coupling transformations, which achieves the best PPLL of $-621.7$.
Furthermore, \method samples consistently contain the same digit class (\reffig{fig:mnist_fs}), showing that we are able to fully model the intra-set dependencies of elements.
% This indicates that \method is able to flexibly model sets not only in low-dimensional, high-cardinality settings,
% % (point clouds),
% but also in high-dimensional, low-cardinality tasks.

% We train models using \bruno, which achieves a PPLL of $-$, \bruno with convolution-based Set-Coupling ($-$), and \method, which achieves the best PPLL at $-$.
% Sample quality follows the same trend as the PPLLs: \bruno sample sets (fig) show mixed digit classes, Set-Coupling (fig) is improved but still shows some digit mixing, \method produces the best samples (fig) with little-to-no mixed classes, demonstrating that \method is able to capture the intra-set dependencies across elements.

% \setlength\intextsep{4pt}
% \begin{table}[h] %{r}{0.5\linewidth}
%     \centering
%     \begin{tabular}{c c c}
%         \specialrule{.1em}{.05em}{.05em}
%         \method & Set-Coupling & \bruno \\
%         \hline
%          $-621.7$ & $-634.8$ & $-643.6$ \\
%          \specialrule{.1em}{.05em}{.05em}
%     \end{tabular}
%     \caption{PPLL's for models trained on \texttt{MNIST}.}
%     \label{table:setmnist}
% \end{table}

% We observe that \bruno with set-coupling transformations improves upon \bruno's PPLL while still falling short of that of \method.
% Samples, depicted in \reffig{fig:mnist_samples}, also suggest that \method is able to model these sets with higher fidelity, generating sets consisting of very similar sets of digits.
% This indicates that \method is able to flexibly model sets not only in low dimensional settings such as point clouds, but also in high dimensional tasks.

%% Conclusion
\section{Conclusion}
In this work, we introduced \method for estimating exchangeable densities.
This is a difficult task, where models were previously limited to either exchangeable base likelihoods \cite{korshunova2018bruno}, or conditionally \tiid restrictions with variational approximations of the likelihood \cite{iclr:edwards}.
We explored how to map inputs to a space that is easier to model whilst preserving exchangeability via equivariant flow transformations.
Among others, we proposed the Set-Coupling transformation which extends existing pointwise coupling transformations \cite{Dinh2014NICENI} to sets.
Additionally, we demonstrated how to apply non-exchangeable density estimators to this task via sorting and scanning.
This is the first tractable approach to achieve this, avoiding averaging over any permutations of the data while unlocking a much larger class of base likelihoods for exchangeable density estimation.
Finally, we argued for the use of an autoregressive base likelihood with sequential transformations to exploit the sequential structure gained in the sort and scan.
Combining equivariant flow transformations, sorting and scanning, and an autoregressive likelihood, we arrived at \method. 
We showed empirically that \method's ability to model intradependencies within sets surpassed that of other state-of-the-art methods in both 
%synthetic and real-world experiments.
high-cardinality, low-dimensionality and low-cardinality, high-dimensionality settings.
Quantitatively \method's likelihoods were a substantial improvement (see \reftab{table:loglikes}).
Furthermore, there was a clear qualitative improvement in samples from \method.

\section*{Acknowledgements}
The authors would like to thank NIH grant HDO55741 for the use of the subcortical data set and Mahmoud Mostapha for preprocessing the data.
Kevin O'Connor would also like to acknowledge the support of NIH grant T32 LM12420.

%\clearpage
\bibliographystyle{aaai}
\bibliography{references}

\appendix
%\ \\ \ \\ \ \\ \ \\ \ \\ \ \\ \ \\ \ \\ \ \\ \ \\ \ \\ \ \\
%\ \\ \ \\ \ \\ \ \\ \ \\

\section{Proof of Prop. 1}
\begin{proof}
First, note that $|\mbox{det}\frac{\ud \hat{q}}{\ud \bx}|$ is invariant to $\Gamma$ since one may compute the Jacobian of $\hat{q}(\Gamma \mathbf{x})$ as the composition of $\hat{q}$ followed by a permutation and the determinant of the Jacobian of a permutation is one. Furthermore, 
$
   \hat{f}(\hat{q}(\Gamma \bx)) = \hat{f}(\Gamma \hat{q}(\bx)) = \hat{f}(\hat{q}(\bx)),
$
by the permutation equivariance of $\hat{q}$ and permutation invariance of $\hat{f}$. Hence, the total likelihood $\hat{p}(\bx) = \big|\mbox{det}\frac{\ud \hat{q}}{\ud \bx}\big| \hat{f}(\hat{q}(\bx))$ is permutation invariant.
\end{proof}

\section{Experiment Details}\label{sec:implementation}
Experiments were implemented in Tensorflow \cite{abadi2016tensorflow}.
We use multiple stacked Real NVP transformations with a Gaussian exchangeable process for BRUNO.
For both \bruno and \vae, we experimented with publicly available implementations in addition to our own.
We observed superior performance for each using our own implementations and thus report these results.
% JO: In my opinion this is a weird choice.. We should eventually validate # of Real NVP (I'm sure it won't make a difference, but it seems odd to choose like this.)
% The number of NVP transformations is adjusted based on the dimensionality of the dataset being modeled. We use four, six, and seven NVP layers for two, three, and seven-dimensional data, respectively.
All results reported for BRUNO are the best-of-six validated trained cases.
We validated eighteen different modifications of the Neural Statistician where we toggled if the variance of the code was fixed (learned or fixed at one), the hidden layer sizes (64, 128, or 256), and the code size (16, 64, or 256). See \cite{oliva2018transformation}, and  \cite{iclr:edwards} 
% JO: for some reason can't cite together??
for further details. (Largest models were typically not best.)
% JO: pretrain typically mean something different (e.g. pretrained on a different dataset)
Results are reported on the best of the eighteen models.
Additionally, each model was initialized six different times and the best model was then trained to convergence.
The best model was selected based 
%on the lowest negative log-likelihood from 
on a held-out validation set.
%The NBA shot data from Sec.~\ref{sec:nba_shots} was downloaded from \url{https://www.kaggle.com/boonpalipatana/nba-playoff-shots-2018}.
% JO: this is fine below.
For \method, we used equivariant pointwise transformations by stacking RNN-coupling and invertible leaky-ReLU transformations \cite{oliva2018transformation}; after the scan we implemented the autoregressive likelihood with a 2-layer GRU (256 units), which conditioned a TAN density \cite{oliva2018transformation} on points. Models were optimized for 40k iterations on TitanXP GPUs. Modelnet data was gathered as in \cite{zaheer2017deep}, brain data was gathered as in \cite{nature2017Early}. All datasets used a random 80/10/10 train/validation/test split.

\subsection{ModelNet10 Ablation Study}\label{sec:modelnet10_ablation}
We performed an ablation study using \texttt{ModelNet10}.
% Here we stripped various components of our model to validate our total method. 
We begin with a full flow scan model, which performs an equivariant flow transformation, scans, does corresponding coupling transformations, and uses an auto-regressive model.
Next, we omit the correspondence coupling transformation. 
After, we also remove the equivariant flow transformation. 
Finally, we considered a basic model without an autoregressive likelihood, that only scans and has a flat-vector density estimate on the vector of concatenated covariates. 
The models achieve per point log likelihoods of $3.01$, $2.67$, $2.34$, and $2.27$, respectively.
We see that each component of \method is improving the likelihood estimate.
It is also interesting that the basic scan model is still outperforming the \vae likelihood bound.

\section{Synthetic} \label{app:synthetic}
The synthetic data in Sec.~\ref{sec:synth_data_section} is generated as follows:
\begin{align*}
    x_1^{(1)} \sim \mathcal{N}(2, n^{-2}) \\
    x_1^{(2)} \sim \mathcal{N}(0, n^{-2}(1 + (\pi/3)^2)) \\
    x_k^{(1)} \sim \mathcal{N}(x_1^{(1)}\cos(\pi k/n), n^{-2}) \\
    x_k^{(2)} \sim \mathcal{N}(\cos(\pi k/n + x_1^{(2)}), n^{-2})
\end{align*}

\section{Permutation Equivariant Transformations}\label{app:transformations}
% simple ones already working well
% for sake of completeness 
For the sake of completeness, we develop several novel permutation equivariant transformations which do not transform each set element independently. 

Recall that a transformation $q: \mathbb{R}^{n\times d} \rightarrow \mathbb{R}^{n\times d}$ is permutation equivariant if for any permutation matrix $\Gamma$, $q(\Gamma \bx) = \Gamma q(\bx)$.
Furthermore, recall that one may construct a simple permutation equivariant transformation by transforming each element of a set identically and independently:
\begin{align}\label{eq:trans_map}
    (x_1, ..., x_n) \mapsto (q(x_1), ..., q(x_n)).
%    \mathbf{x} = (x_1, ..., x_n) \mapsto (q(x_1), ..., q(x_n)).
\end{align}
However, this transformation is unable to capture any dependencies between points, and operates in a \tiid fashion.
Instead, we propose equivariant transformations that transform each element of a set in a way that depends on other points in the set, yielding a richer family of models. In other words, transforming as
\begin{align}
%    (x_1, ..., x_n) \mapsto (q(x_1, \{x_i\}_{i=1}^n), ..., q(x_n, \{x_i\}_{i=1}^n)).
    (x_1, ..., x_n) \mapsto (q(x_1, \mathbf{x}), ..., q(x_n, \mathbf{x})).
\end{align}
Below, we propose several novel equivariant transformations with intra-set dependencies.

%% L-PEq
\subsection{Linear Permutation Equivariant (L-PEq)}
We start with a linear permutation equivariant transformation.
It can be shown \cite{zaheer2017deep} that any linear permutation equivariant map of one-dimensional points can be written in the form, $\mathbf{x} \mapsto (\lambda \mathbf{I} + \gamma \mathbf{1} \mathbf{1}^T) \mathbf{x}$ for some scalars $\lambda$ and $\gamma$, and $\mathbf{x} \in \R^{n \times 1}$.
Specifically, a linear permutation equivariant transformation is the result of a matrix multiplication with identical diagonal elements and off diagonal elements.

Such a transformation captures intradependencies by mapping the $j$th dimension of the $i$th point as
\begin{align}
    \pind{x_i}{j} \mapsto \pind{\lambda}{j} \pind{x_i}{j} + \frac{\pind{\gamma}{j}}{n} \sum_{k=1}^n \pind{x_k}{j}, \label{eq:linpermeq}
\end{align}
incorporating the mean of other points in the set.
We use the mean rather than the sum as in \cite{zaheer2017deep} because it allows for better symmetry with our proposed generalization in Sec.~\ref{sec:weight}.
It is trivial to go between the two formulations by scaling $\pind{\gamma}{j}$ by $n$.
The log-determinant of the transformation \eqref{eq:linpermeq} can be show to be $(n-1)\log |\lambda^{(j)}| + \log |\lambda^{(j)} + \gamma^{(j)}|$, and is invertible whenever $\lambda^{(j)} \neq 0$ and $\lambda^{(j)} + \gamma^{(j)} \neq 0$ with inverse:
\begin{align*}
    z^{(j)}_i \mapsto \frac{z^{(j)}_i}{\lambda^{(j)}} - \frac{\gamma^{(j)}}{n\lambda^{(j)} (\lambda^{(j)} + \gamma^{(j)})} \sum\limits_{k=1}^n z_k^{(j)}
\end{align*}

\subsection{Nonlinear Weighting (NW-PEq)}\label{sec:weight}
We propose a generalization of the linear permutation equivariant transformation \eqref{eq:linpermeq} here.
Instead of a direct mean, we propose to weight each element by some nonlinear function that depends on the element's value relative to a global operation over the set:
\begin{align}\label{eq:wpeq}
    \pind{x_i}{j} & \mapsto \pind{\lambda}{j} \pind{x_i}{j} + \pind{\gamma}{j} \tfrac{\sum_k \pind{x_k}{j} w\left(\pind{x_k}{j}\right)}{\sum_m w\left(\pind{x_m}{j}\right)} \\
    & \mapsto \pind{\lambda}{j} \pind{x_i}{j} + \pind{\gamma}{j} \pind{\eta}{j} \notag
\end{align}
where $w$ is the nonlinear weighting function and $\pind{\eta}{j}$ is the weighted mean.
The log-determinant of the Jacobian can be expressed as
\begin{align}\label{eq:wjac}
\begin{split}
    \log&|J| = (n-1)\log|\pind{\lambda}{j}| \\
    &+ \log\left|\pind{\lambda}{j} + \pind{\gamma}{j} \left(1 + 
    \tfrac{\sum_k \left(\pind{x_k}{j} - \pind{\eta}{j}  \right)w'\left(\pind{x_k}{j}\right)}
    {\sum_m w\left(\pind{x_m}{j}\right)} \right) \right|
\end{split}
\end{align}
where $w'$ is the first derivative of $w$. It is clear that \refeq{eq:wjac} simplifies to the linear determinant for constant $w$. Attempting to invert \refeq{eq:wpeq} results in an implicit function for $\pind{\eta}{j}$
\begin{align}\label{eq:qpeq_inv_gen}
    \pind{\eta}{j} = \left(1 + \pind{\gamma}{j}\right)^{-1}
    \tfrac{\sum_k \pind{x_k}{j} w\left(\pind{x_k}{j} - \pind{\gamma}{j} \pind{\eta}{j}\right)}
    {\sum_m w\left(\pind{x_m}{j} - \pind{\gamma}{j}\pind{\eta}{j}\right)}
\end{align}
(where $\pind{\lambda}{j}$ has been dropped for brevity) that could be solved numerically to perform the inverse for a general nonlinear weight. This formulation implies that the weighted permutation equivariant transform can be inverted even if $w$ is not an invertible function. Thus, allowing for a larger family of nonlinearities than is typically included in transformative likelihood estimators \cite{Dinh2014NICENI,DBLP:journals/corr/DinhSB16,kingma2018glow}.

The simplest method forward is to choose a weighting function such that a sum of function inputs decomposes into a product of outputs, e.g.~$w(a+b) = f(a)f(b)$ where $f$ is some nonlinear function. In this case, \refeq{eq:qpeq_inv_gen} simplifies to
\begin{align}\label{eq:qpeq_inv_simp}
    \pind{\eta}{j} = \left(1 + \pind{\gamma}{j}\right)^{-1}
    \tfrac{\sum_k \pind{x_k}{j} f\left(\pind{x_k}{j}\right)}
    {\sum_m f\left(\pind{x_m}{j}\right)}
\end{align}
and the inverse transform proceeds trivially.
Choosing $f$ to be the exponential function allows for the simplification, guarantees positive weights, and results in a softmax-weighted mean,
\begin{align}
  \pind{x_i}{j} \mapsto \pind{\lambda}{j} \pind{x_i}{j} + \pind{\gamma}{j} \tfrac{\sum_k \pind{x_k}{j} \exp\left(\pind{\beta}{j} \pind{x_k}{j}\right)}{\sum_m \exp\left(\pind{\beta}{j} \pind{x_m}{j}\right)}
\end{align}
% Note: Old version of the following paragraph containing justification of inverse temperature scaling has been moved to scraps.tex.  -KO
with inverse temperature scaling $\beta.$ It is apparent that this transformation reduces to the L-PEq transformation when $\beta=0$. Additionally, in the limit as $\beta \rightarrow \infty$ or $\beta \rightarrow -\infty$, the transformation tends to shift by the maximum or minimum of the set, respectively. The log-determinant of this transformation is identical to the linear case and the inverse comes directly from (\ref{eq:qpeq_inv_simp}):
\begin{align*}
    z^{(j)}_i \mapsto \frac{z^{(j)}_i}{\lambda^{(j)}} - \frac{\gamma^{(j)}}{\lambda^{(j)}\left(\lambda^{(j)} + \gamma^{(j)}\right)} \frac{\sum_k z^{(j)}_k \exp(\frac{\pind{\beta}{j} z^{(j)}_i}{\lambda^{(j)}})}{\sum_m \exp(\frac{\pind{\beta}{j}z^{(j)}_i}{\lambda^{(j)}})}
\end{align*}
where $\pind{\lambda}{j}$ has been reintroduced. Other choices for the nonlinear weight function $w$ are possible, however finding a good map that has both a closed-form log-determinant and inverse is non-trivial. Alternate weighting functions remain a direction for future research.

\section{Generated Samples for Point Cloud Experiments}\label{app:samples}

\begin{figure}[p]
    \centering
    \begin{subfigure}[t]{.32\linewidth}
        \begin{minipage}{\linewidth}
            \centering
            \includegraphics[height=22mm]{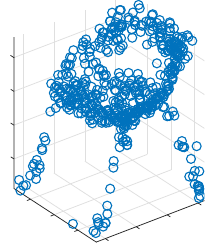}
            \includegraphics[height=22mm]{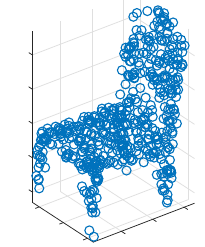} \\
            \includegraphics[height=22mm]{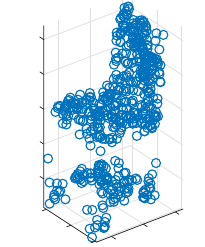}
            \includegraphics[height=22mm]{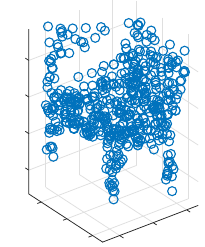}
        \end{minipage}
        \caption{\method}
    \end{subfigure}
    \begin{subfigure}[t]{.3\linewidth}
        \begin{minipage}{\linewidth}
            \centering
            \includegraphics[height=22mm]{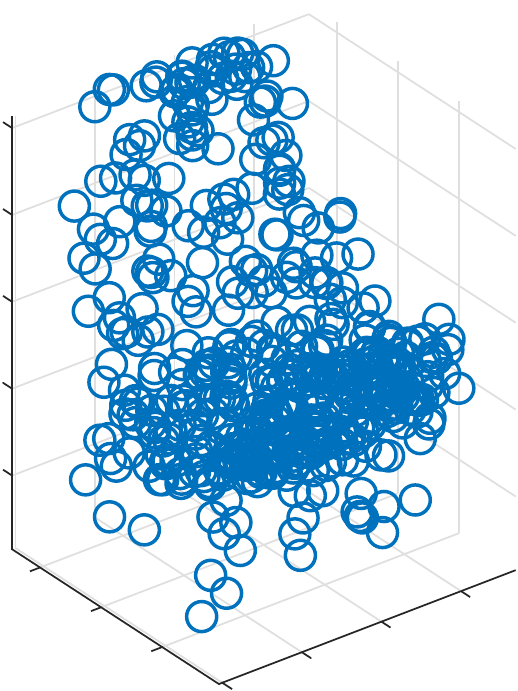}
            \includegraphics[height=22mm]{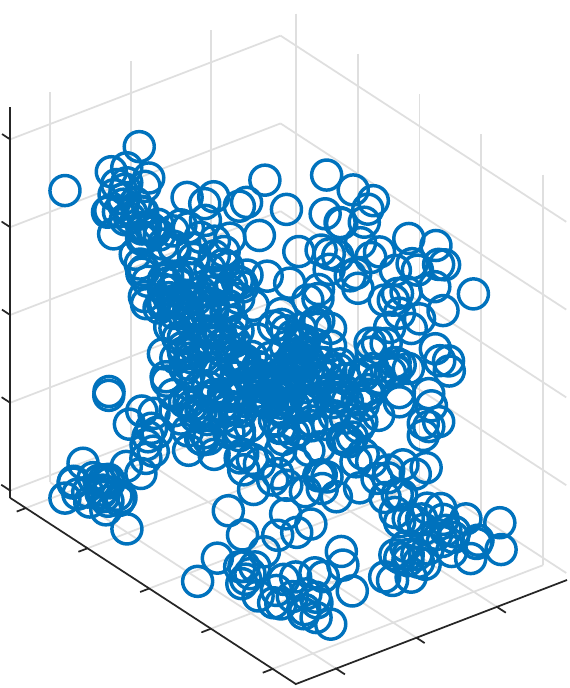} \\
            \includegraphics[height=22mm]{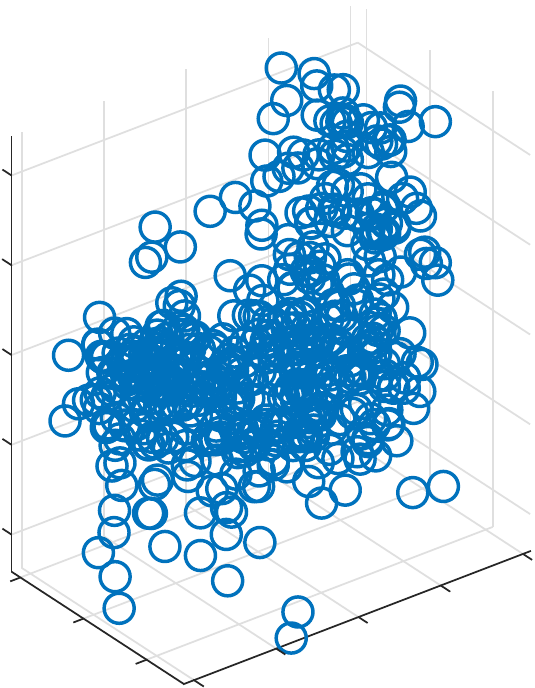}
            \includegraphics[height=22mm]{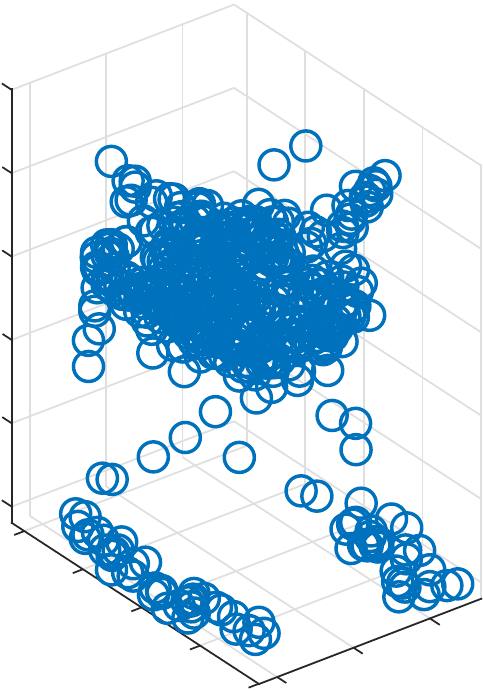}
        \end{minipage}
        \caption{\vae}
    \end{subfigure}
    \begin{subfigure}[t]{.3\linewidth}
        \begin{minipage}{\linewidth}
            \centering
            \includegraphics[height=22mm]{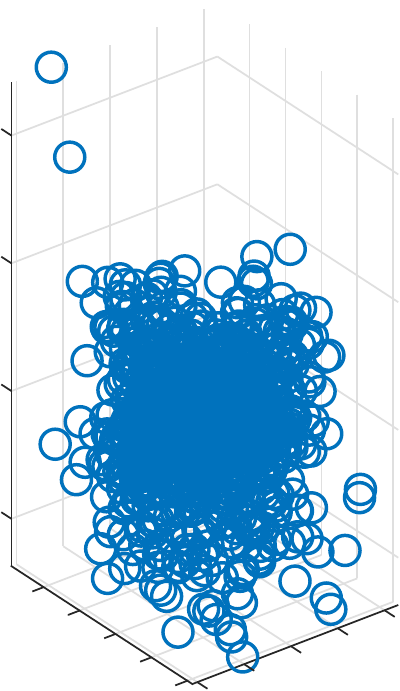}
            \includegraphics[height=22mm]{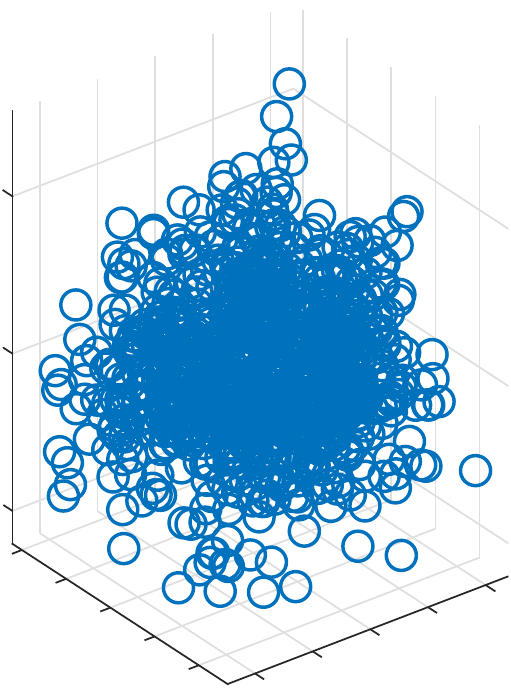} \\
            \includegraphics[height=22mm]{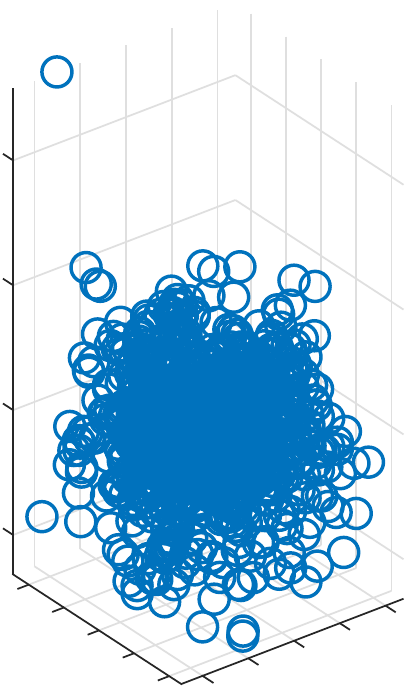}
            \includegraphics[height=22mm]{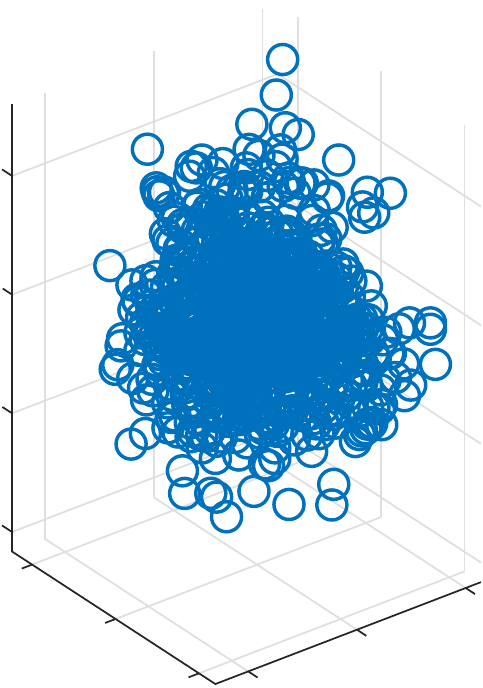}
        \end{minipage}
        \caption{BRUNO}
    \end{subfigure}
    \caption{Chair Samples }
    \label{fig:chair}
\end{figure}

\begin{figure}[p]
    \centering
    \begin{subfigure}[t]{.32\linewidth}
        \begin{minipage}{\linewidth}
            \centering
            \includegraphics[height=20mm]{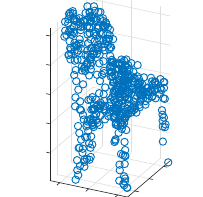}
            \includegraphics[height=20mm]{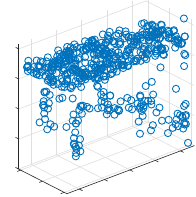} \\
            \includegraphics[height=20mm]{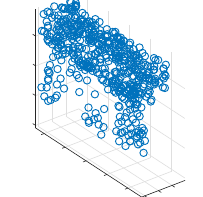}
            \includegraphics[height=20mm]{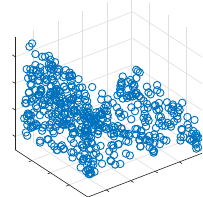}
        \end{minipage}
        \caption{\method}
    \end{subfigure}
    \begin{subfigure}[t]{.32\linewidth}
        \begin{minipage}{\linewidth}
            \centering
            \includegraphics[height=20mm]{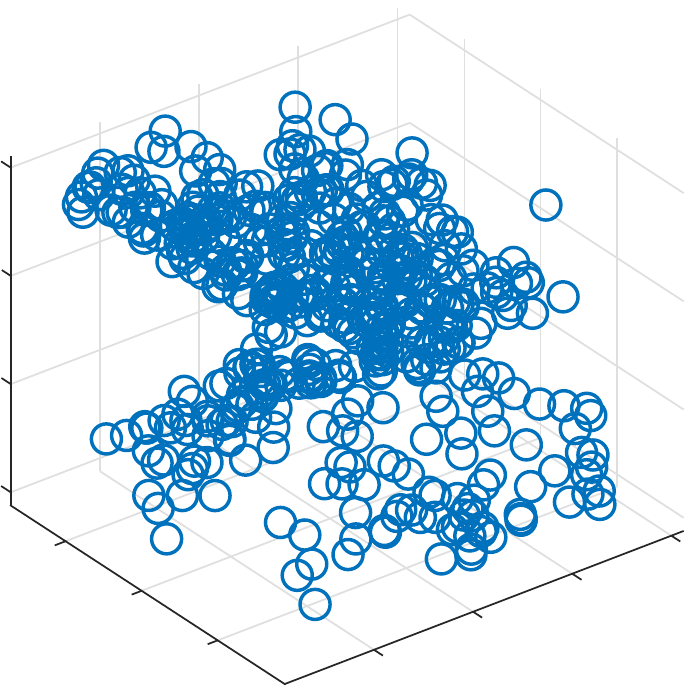}
            \includegraphics[height=20mm]{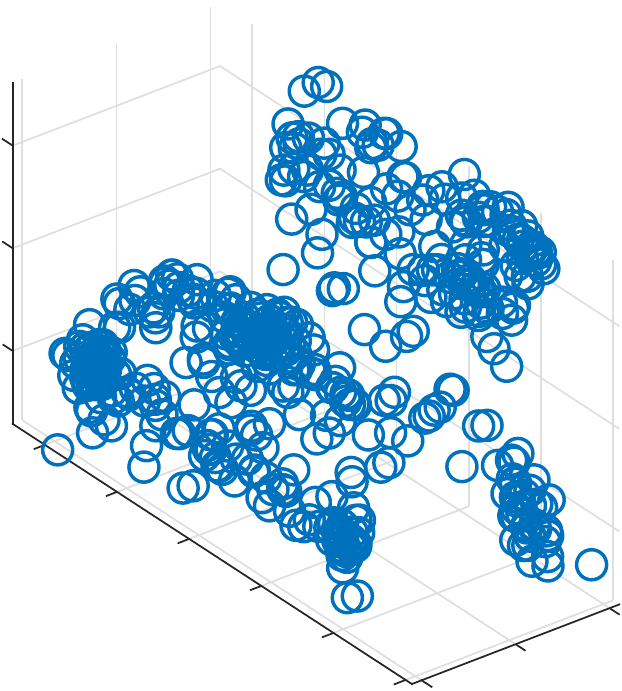} \\
            \includegraphics[height=20mm]{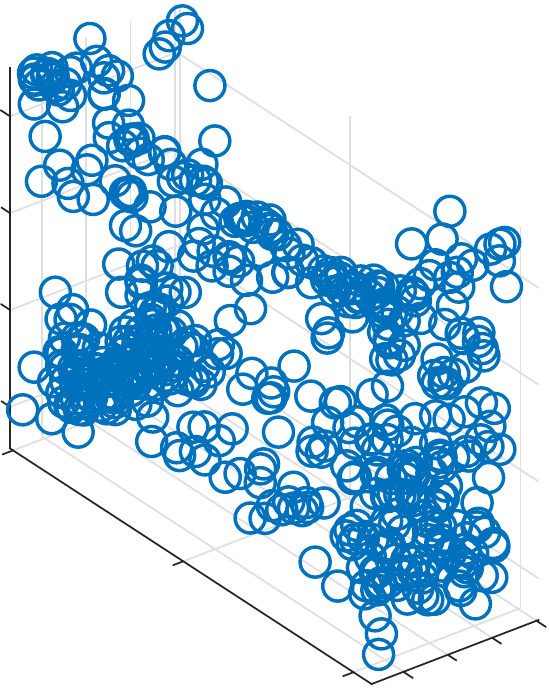}
            \includegraphics[height=20mm]{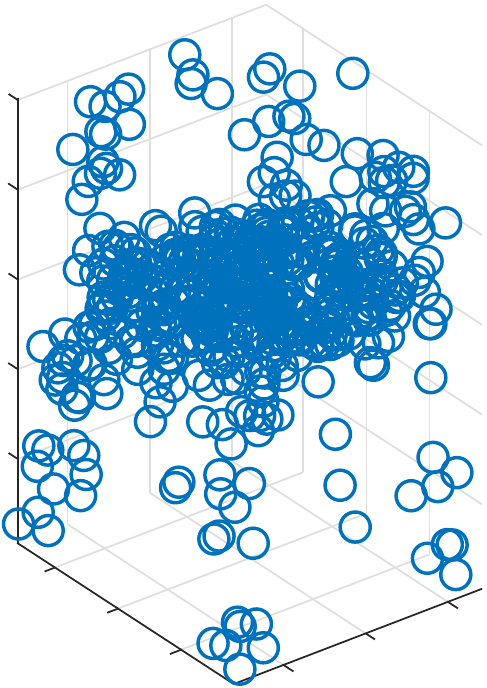}
        \end{minipage}
        \caption{\vae}
    \end{subfigure}
    \begin{subfigure}[t]{.3\linewidth}
        \begin{minipage}{\linewidth}
            \centering
            \includegraphics[height=20mm]{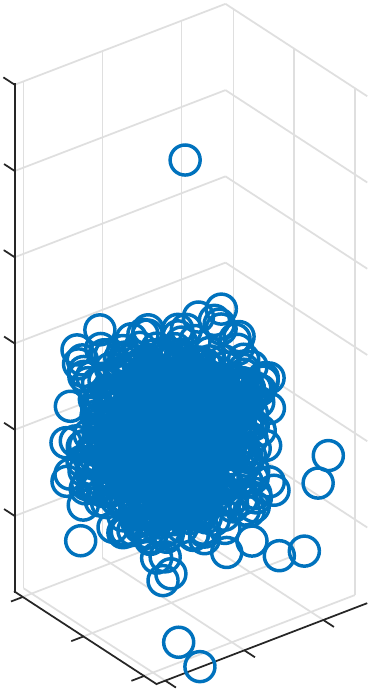}
            \includegraphics[height=20mm]{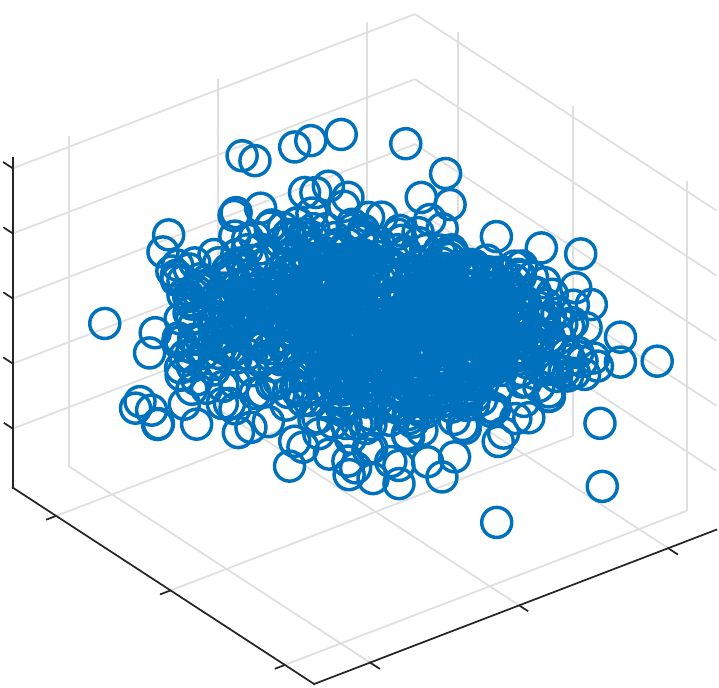} \\
            \includegraphics[height=20mm]{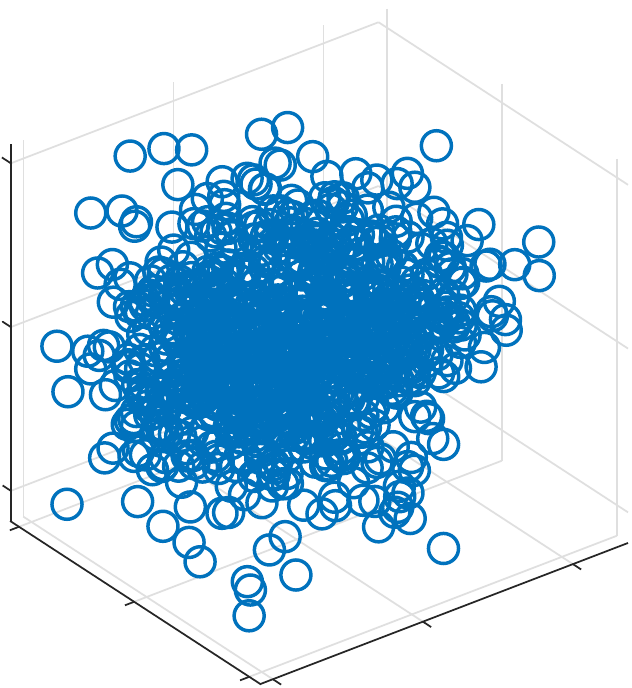}
            \includegraphics[height=20mm]{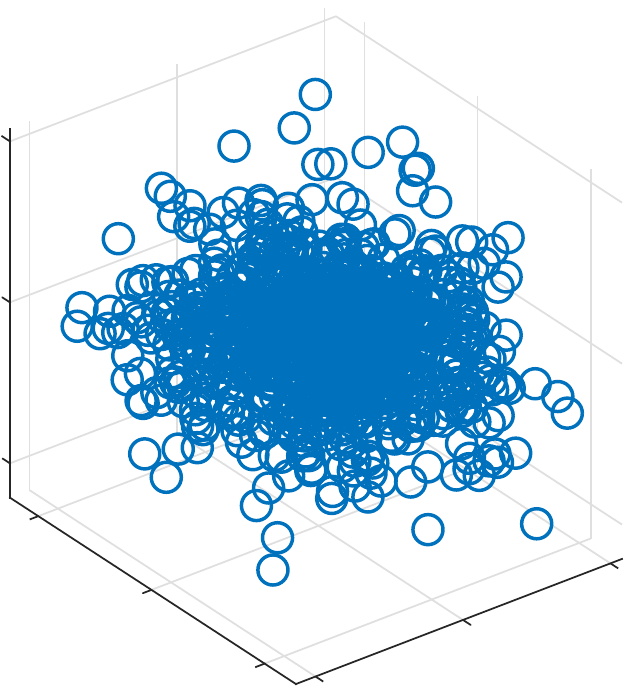}
        \end{minipage}
        \caption{BRUNO}
    \end{subfigure}
    \caption{ModelNet10 Samples }
    \label{fig:modelnet10}
\end{figure}

\begin{figure}[p]
    \centering
    \begin{subfigure}[t]{.32\linewidth}
        \begin{minipage}{\linewidth}
            \centering
            \includegraphics[height=16mm]{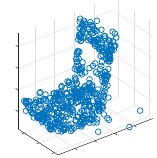}
            \includegraphics[height=10mm]{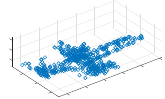} \\
            \includegraphics[height=10mm]{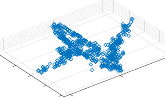}
            \includegraphics[height=14mm]{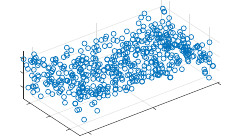}
        \end{minipage}
        \caption{\method}
    \end{subfigure}
    \begin{subfigure}[t]{.32\linewidth}
        \begin{minipage}{\linewidth}
            \centering
            \includegraphics[height=14mm]{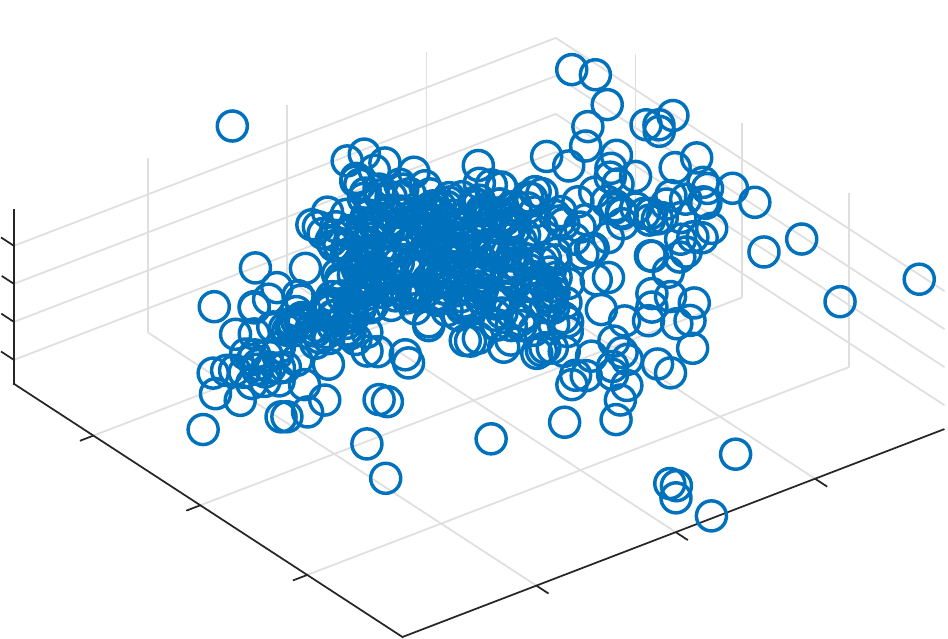}
            \includegraphics[height=12mm]{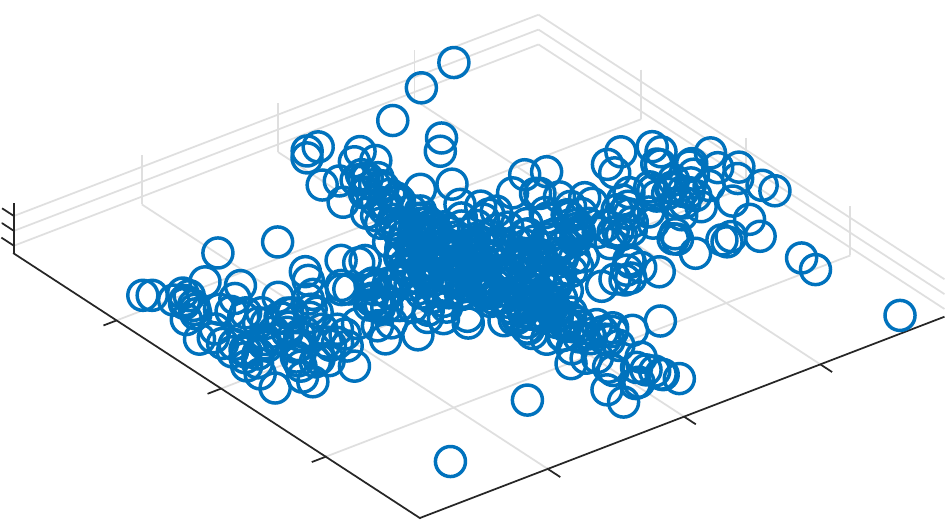} \\
            \includegraphics[height=16mm]{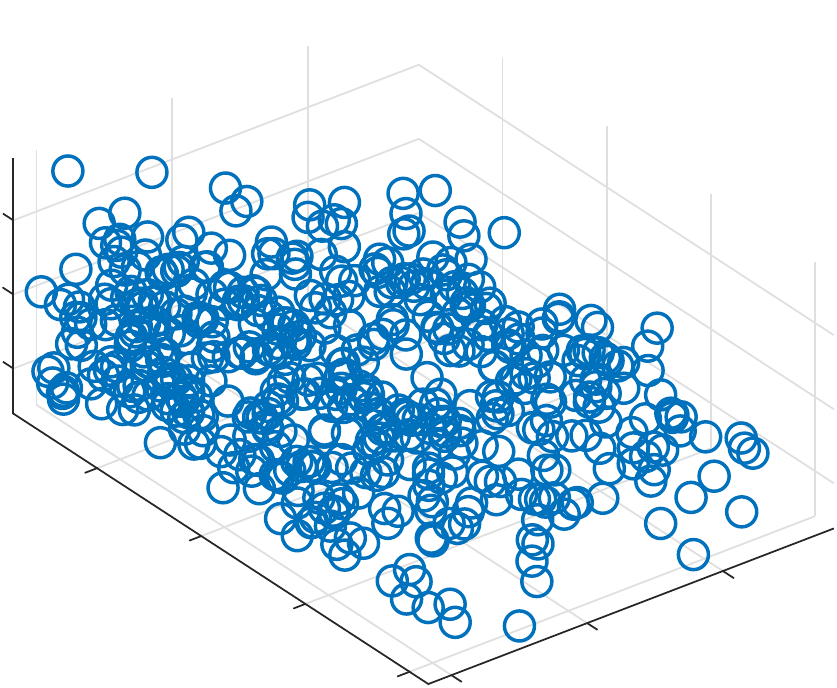}
            \includegraphics[height=18mm]{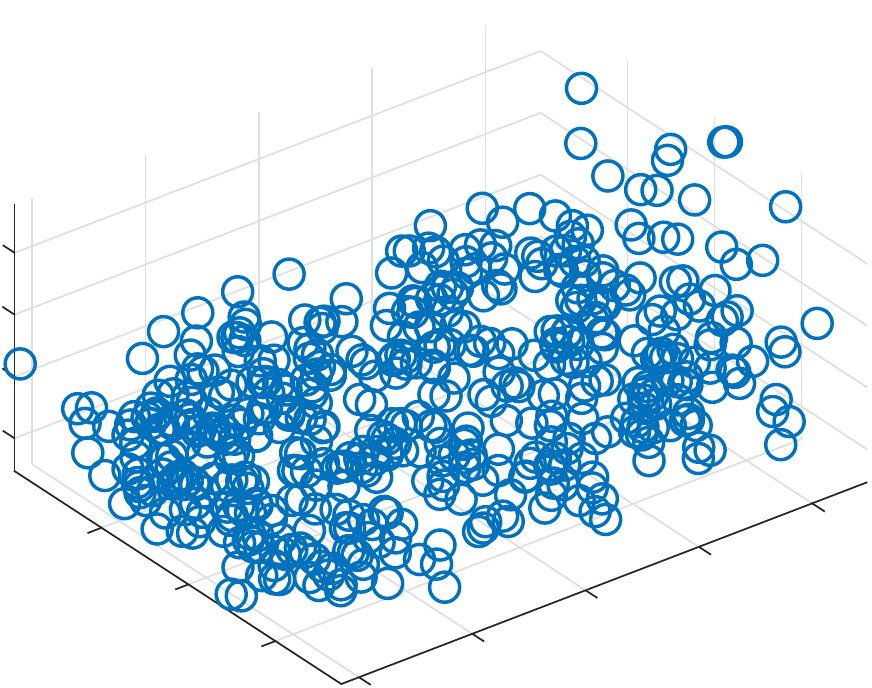}
        \end{minipage}
        \caption{\vae}
    \end{subfigure}
    \begin{subfigure}[t]{.32\linewidth}
        \begin{minipage}{\linewidth}
            \centering
            \includegraphics[height=20mm]{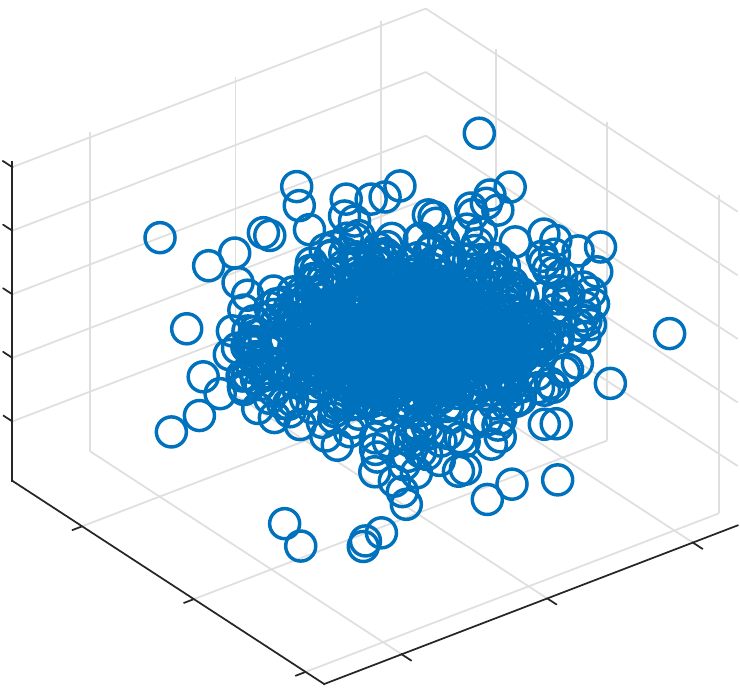}
            \includegraphics[height=20mm]{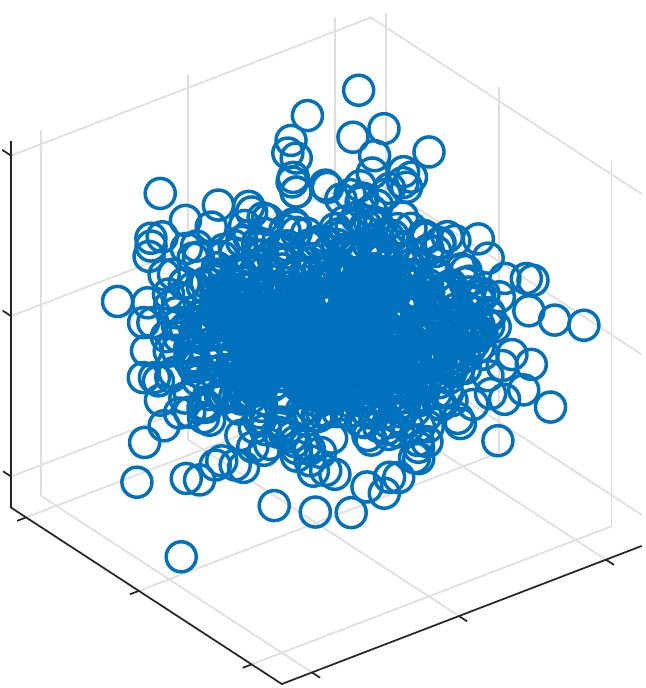} \\
            \includegraphics[height=20mm]{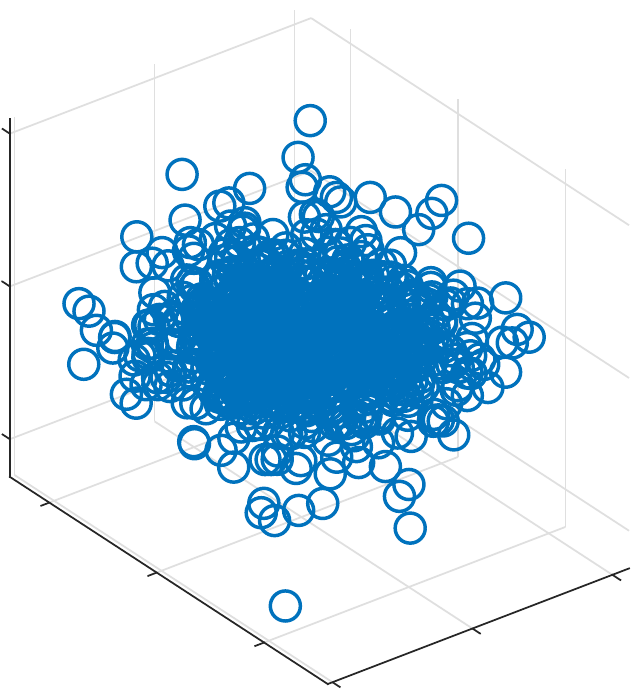}
            \includegraphics[height=20mm]{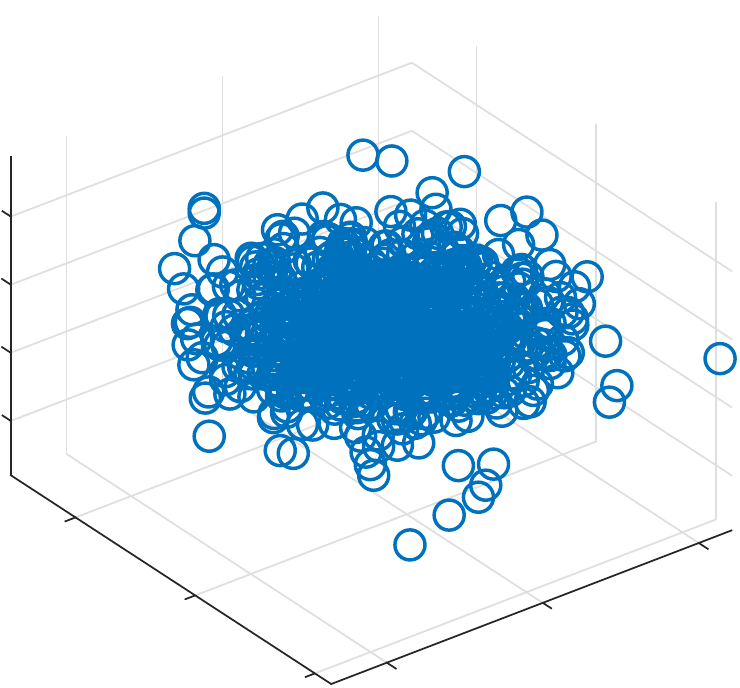}
        \end{minipage}
        \caption{BRUNO}
    \end{subfigure}
    \caption{ModelNet10a Samples }
    \label{fig:modelnet10a}
\end{figure}

\begin{figure}[b!]
    \centering
    \begin{subfigure}[t]{.32\linewidth}
        \begin{minipage}{\linewidth}
            \centering
            \includegraphics[height=25mm]{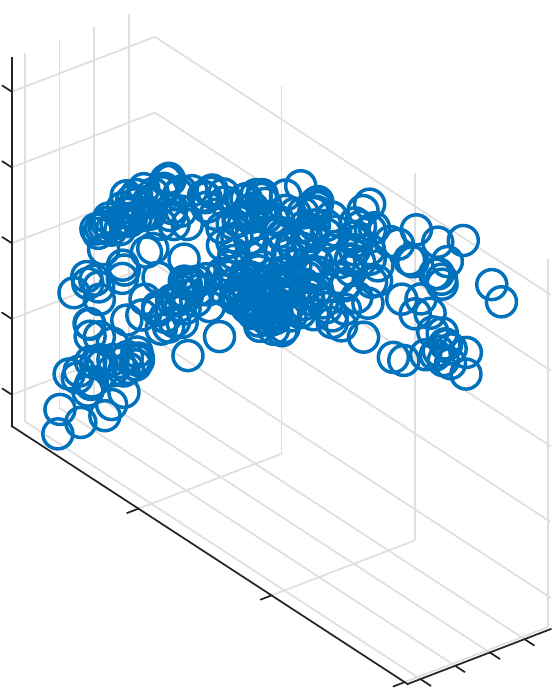}
            \includegraphics[height=25mm]{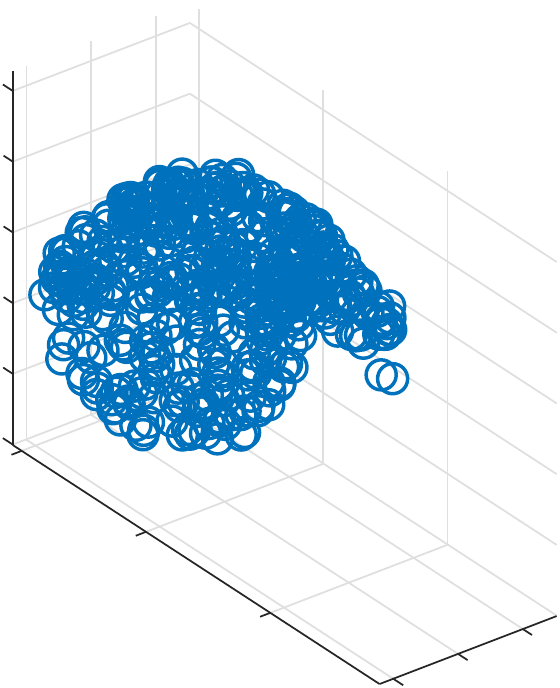} \\
            \includegraphics[height=25mm]{figures/samples/caudate/fs/sample_3.pdf}
            \includegraphics[height=25mm]{figures/samples/caudate/fs/sample_4.pdf}
        \end{minipage}
        \caption{\method}
    \end{subfigure}
    \begin{subfigure}[t]{.32\linewidth}
        \begin{minipage}{\linewidth}
            \centering
            \includegraphics[height=24mm]{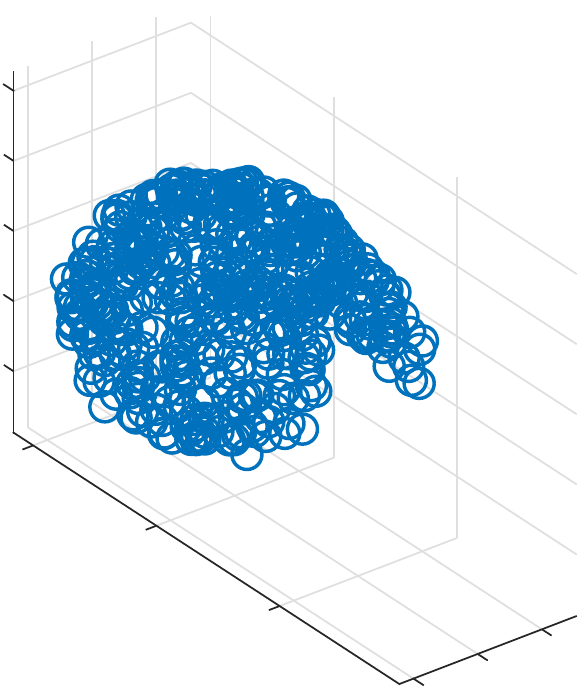}
            \includegraphics[height=24mm]{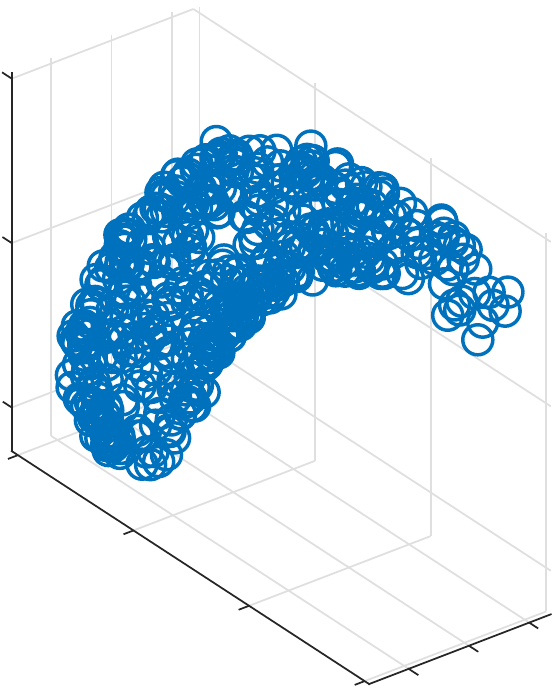} \\
            \includegraphics[height=24mm]{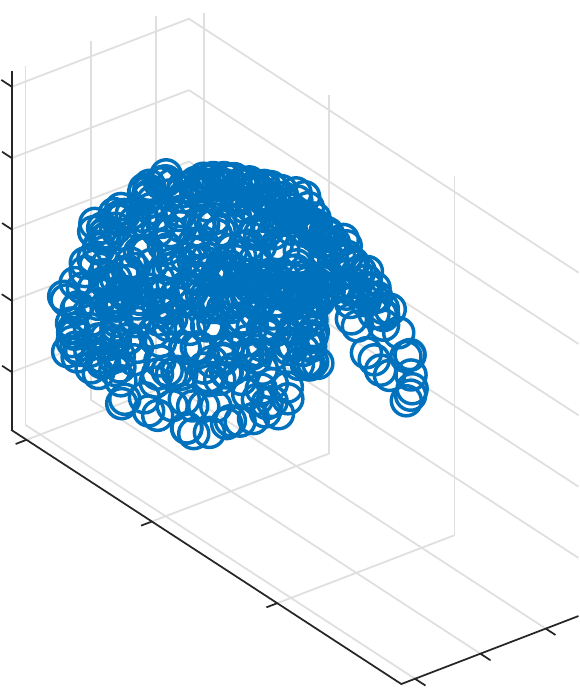}
            \includegraphics[height=24mm]{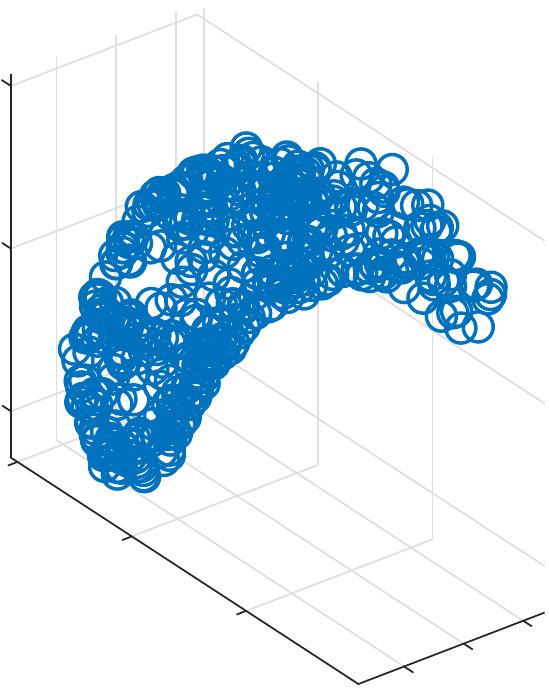}
        \end{minipage}
        \caption{\vae}
    \end{subfigure}
    \begin{subfigure}[t]{.32\linewidth}
        \begin{minipage}{\linewidth}
            \centering
            \includegraphics[height=25mm]{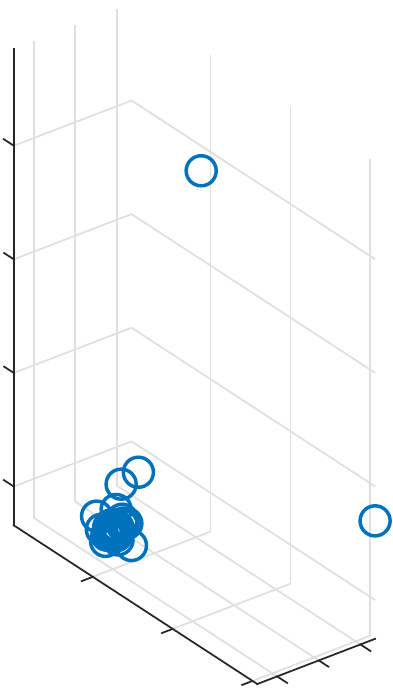}
            \includegraphics[height=22mm]{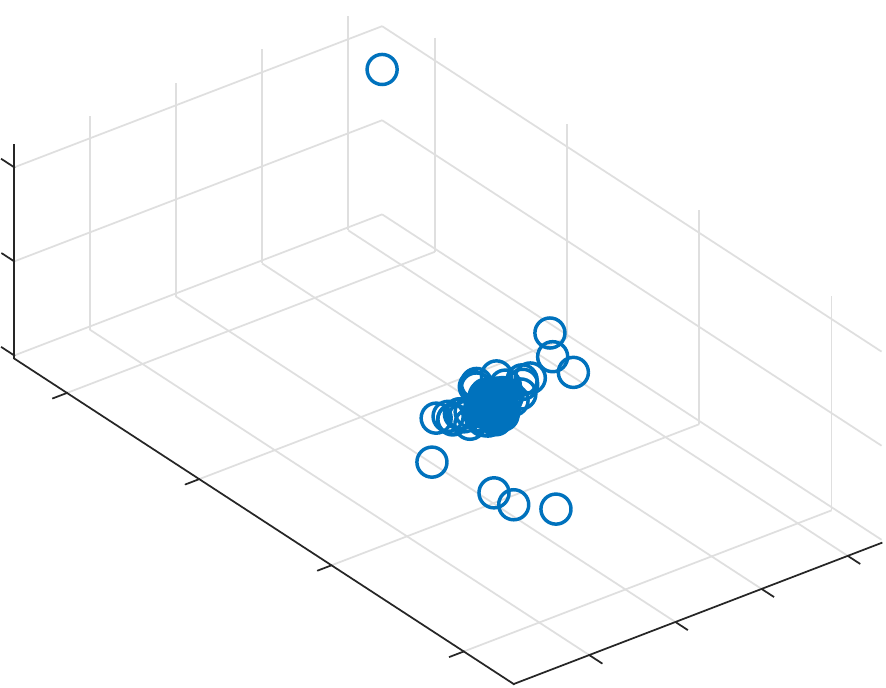} \\
            \includegraphics[height=25mm]{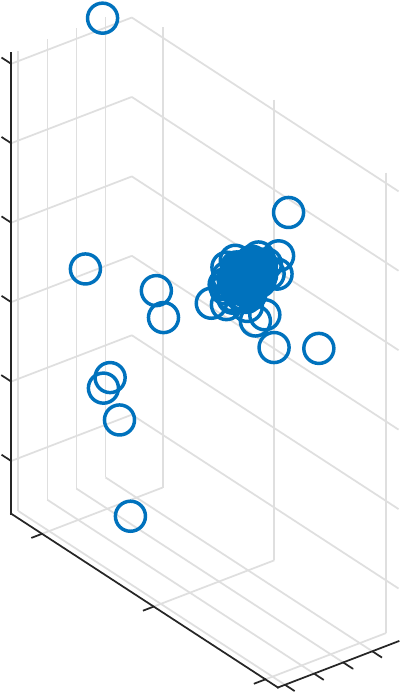}
            \includegraphics[height=20mm]{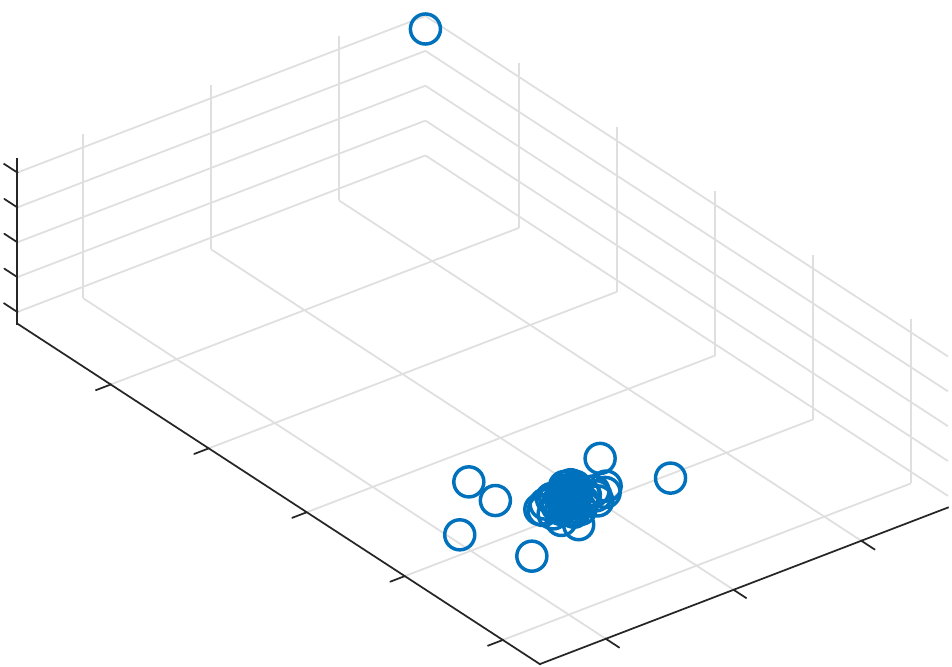}
        \end{minipage}
        \caption{BRUNO}
    \end{subfigure}
    \caption{Caudate Samples }
    \label{fig:caudate_samples}
\end{figure}

\begin{figure}[h!]
    \centering
    \begin{subfigure}[t]{.32\linewidth}
        \begin{minipage}{\linewidth}
            \centering
            \includegraphics[height=20mm]{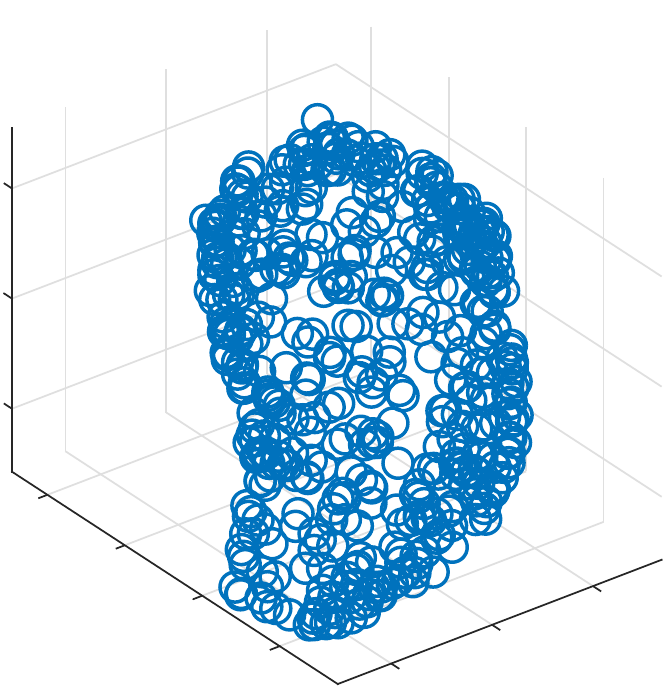}
            \includegraphics[height=20mm]{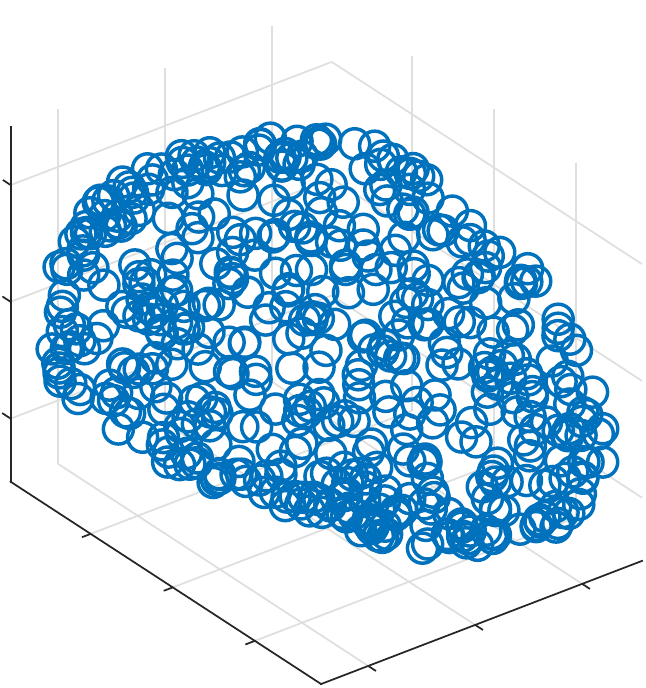} \\
            \includegraphics[height=20mm]{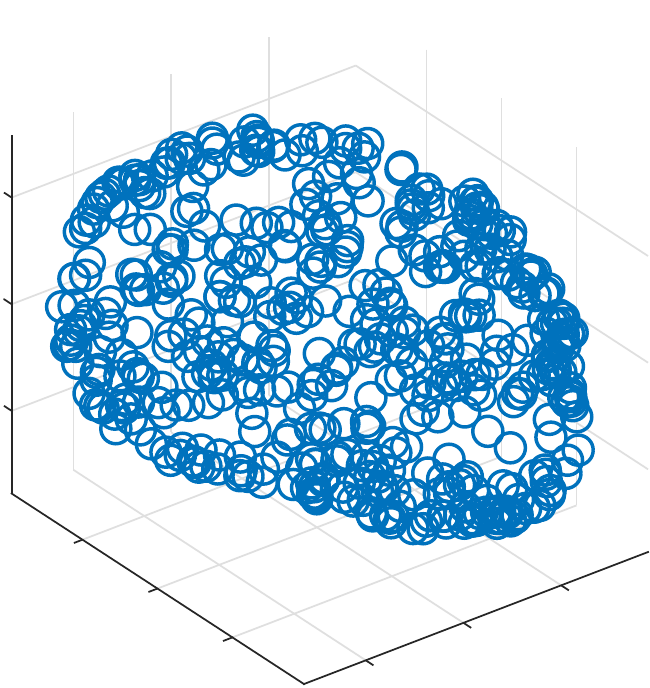}
            \includegraphics[height=20mm]{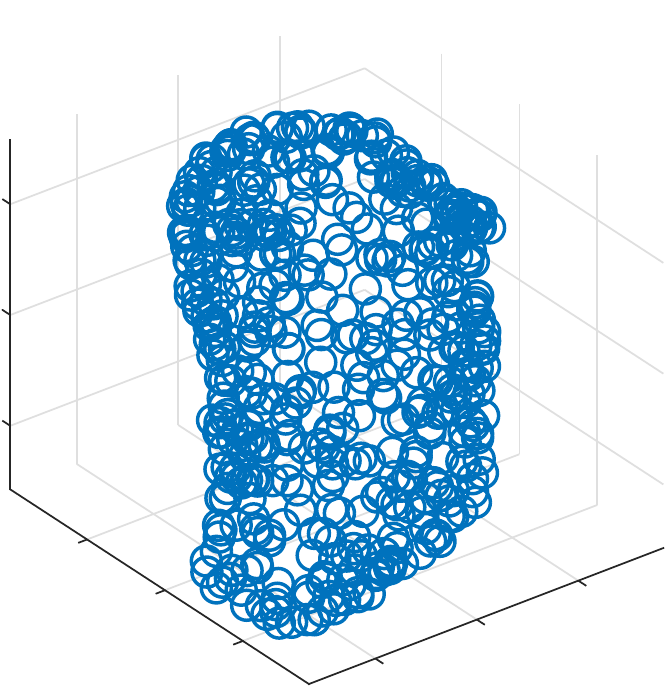}
        \end{minipage}
        \caption{\method}
    \end{subfigure}
    \begin{subfigure}[t]{.32\linewidth}
        \begin{minipage}{\linewidth}
            \centering
            \includegraphics[height=20mm]{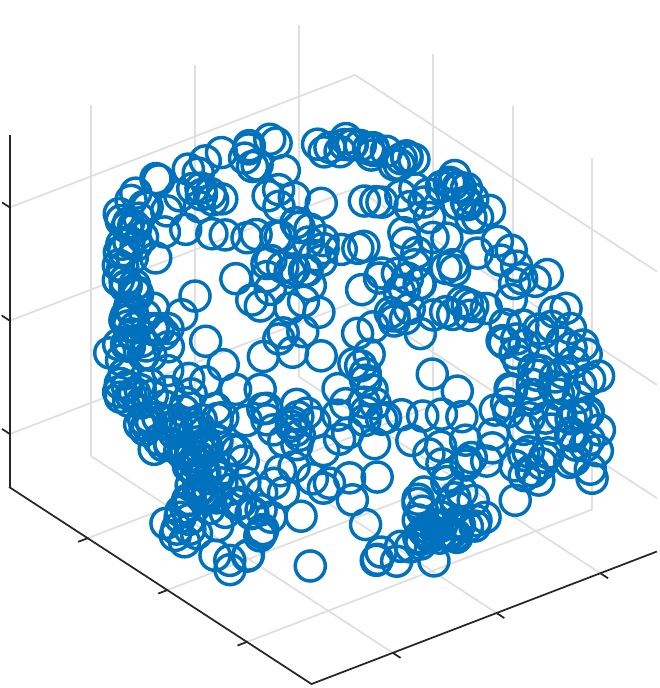}
            \includegraphics[height=20mm]{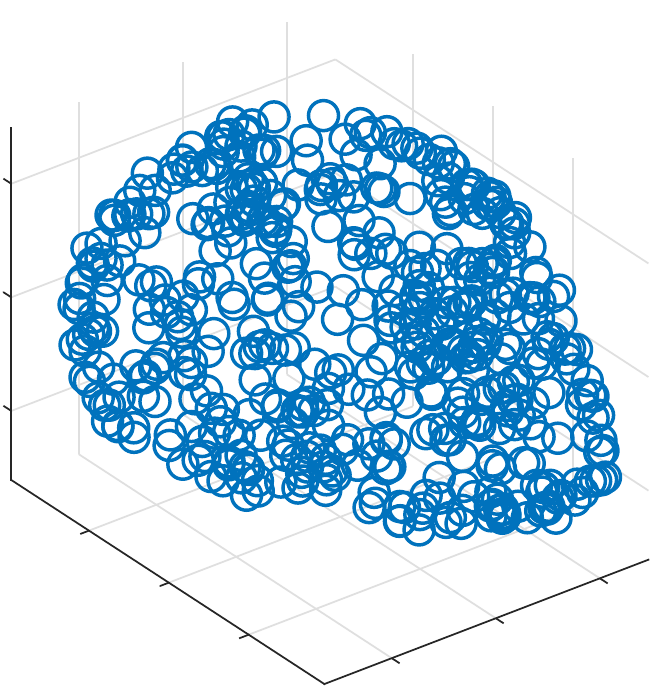} \\
            \includegraphics[height=20mm]{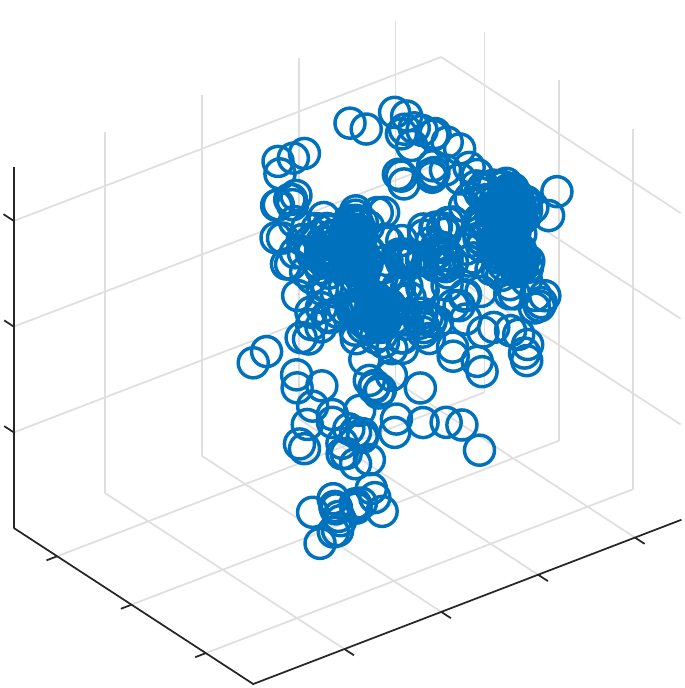}
            \includegraphics[height=20mm]{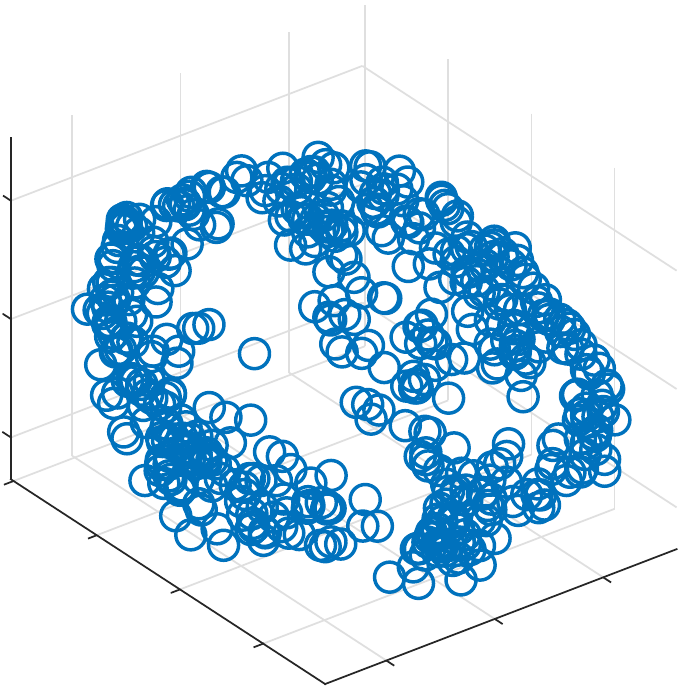}
        \end{minipage}
        \caption{\vae}
    \end{subfigure}
    \begin{subfigure}[t]{.32\linewidth}
        \begin{minipage}{\linewidth}
            \centering
            \includegraphics[height=18mm]{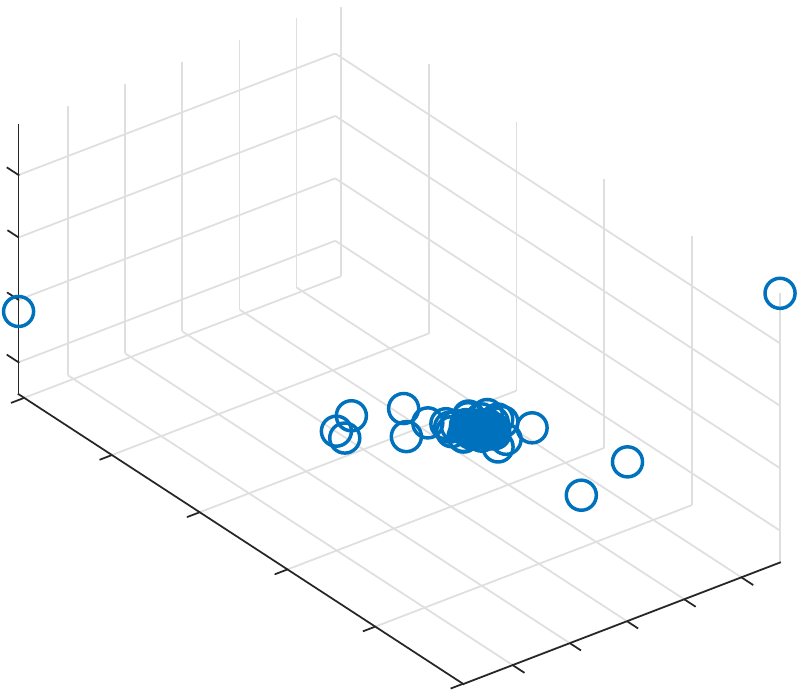}
            \includegraphics[height=20mm]{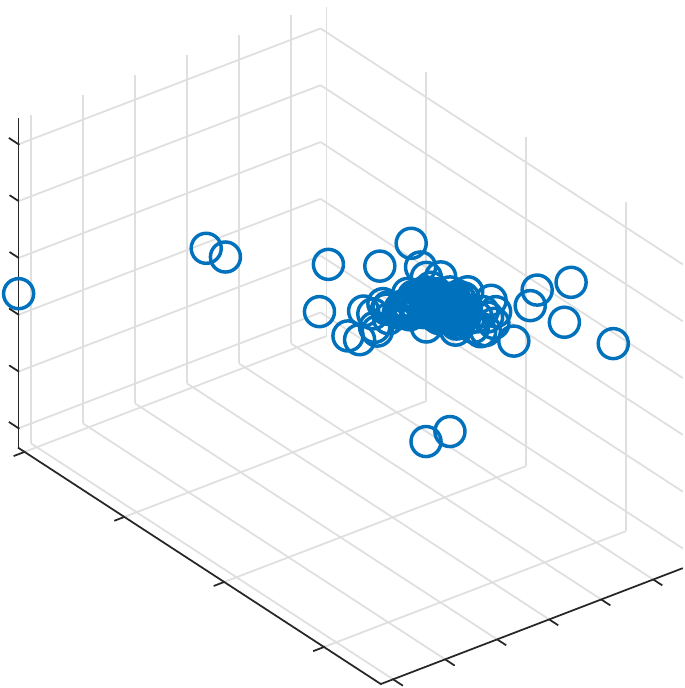} \\
            \includegraphics[height=14mm]{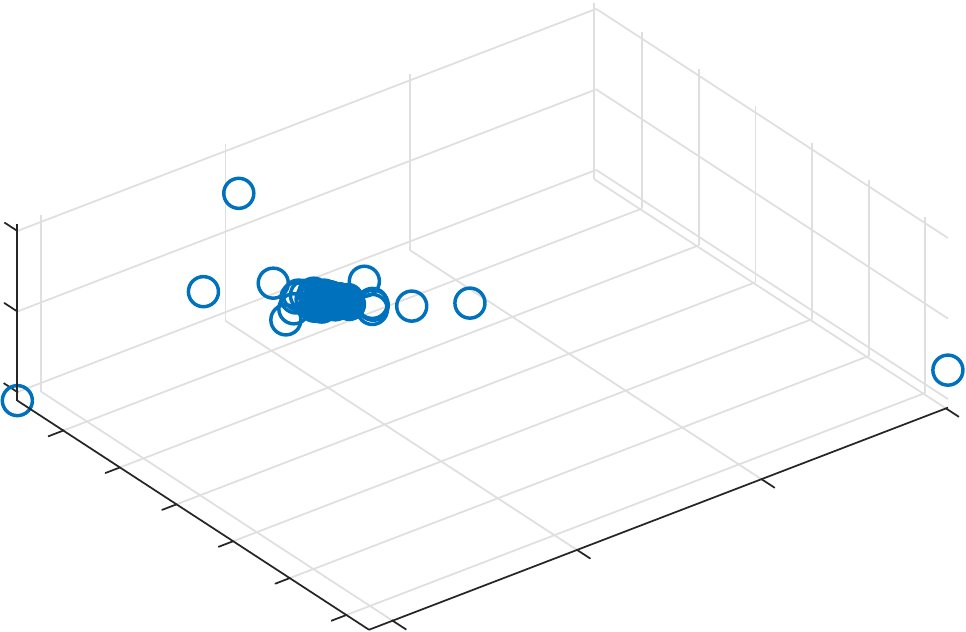}
            \includegraphics[height=15mm]{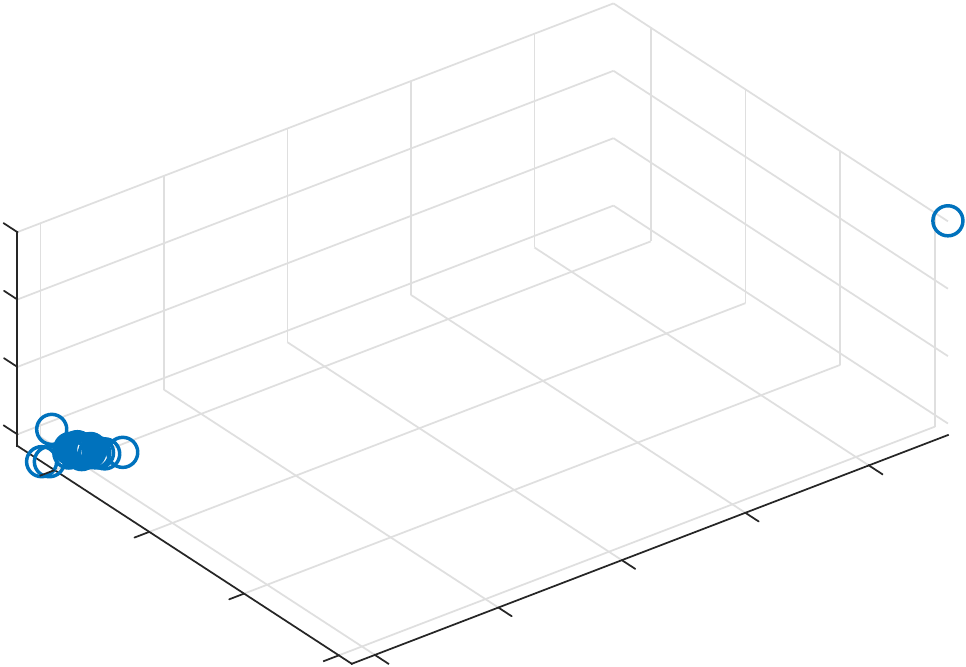}
        \end{minipage}
        \caption{BRUNO}
    \end{subfigure}
    \caption{Thalamus Samples }
    \label{fig:thalamus_samples}
\end{figure}

Below we plot additional sampled sets using the methods compared in our experiments in Figures \ref{fig:chair}-\ref{fig:thalamus_samples}.

%\ \\
%\ \\ \ \\

\section{Training Examples for Point Cloud Experiments}\label{app:training_examples}
\begin{figure}[h!]
    \centering
    \begin{subfigure}[t]{.25\linewidth}
        \begin{minipage}{\linewidth}
            \centering
            \includegraphics[scale=.175]{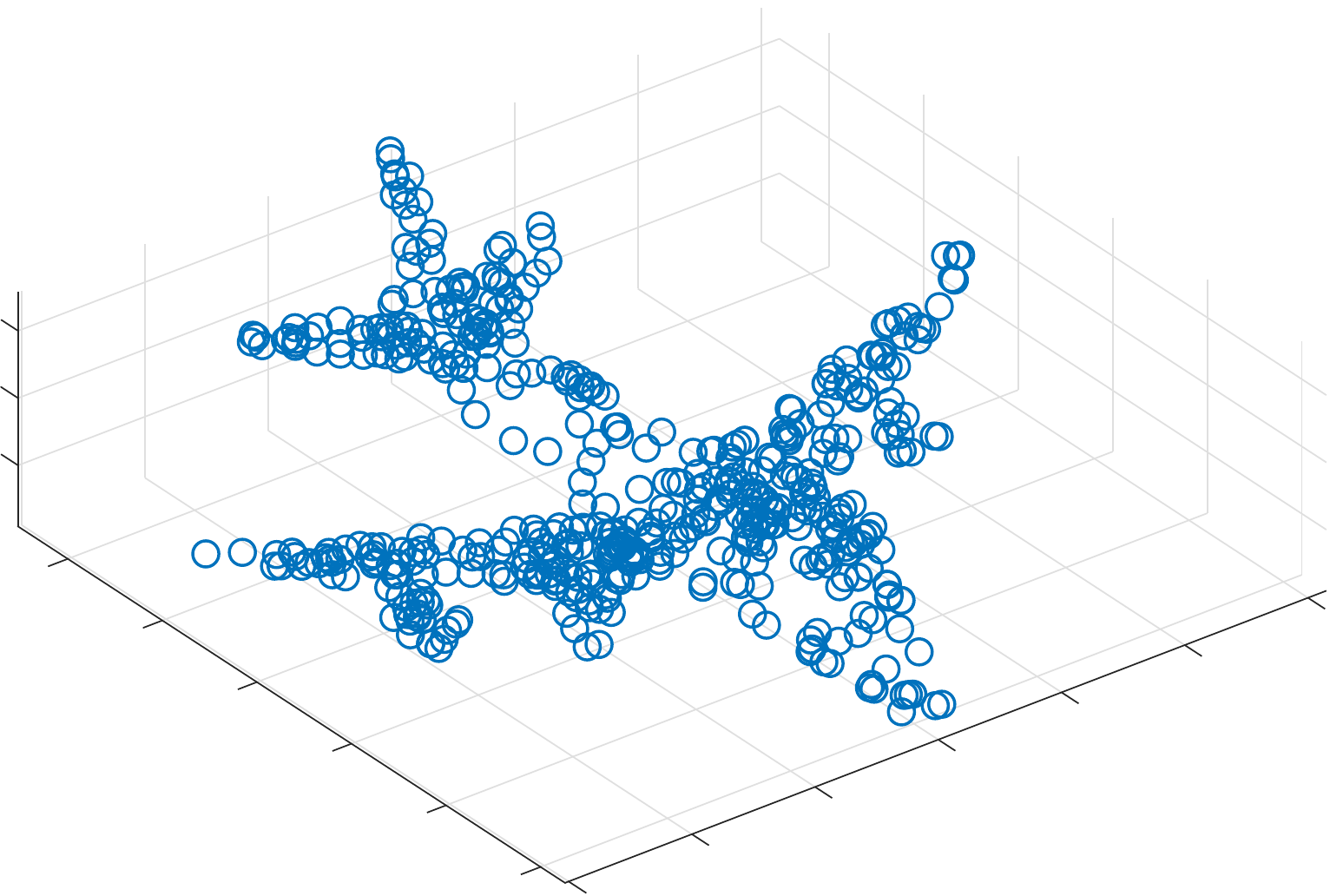}
            \includegraphics[scale=.175]{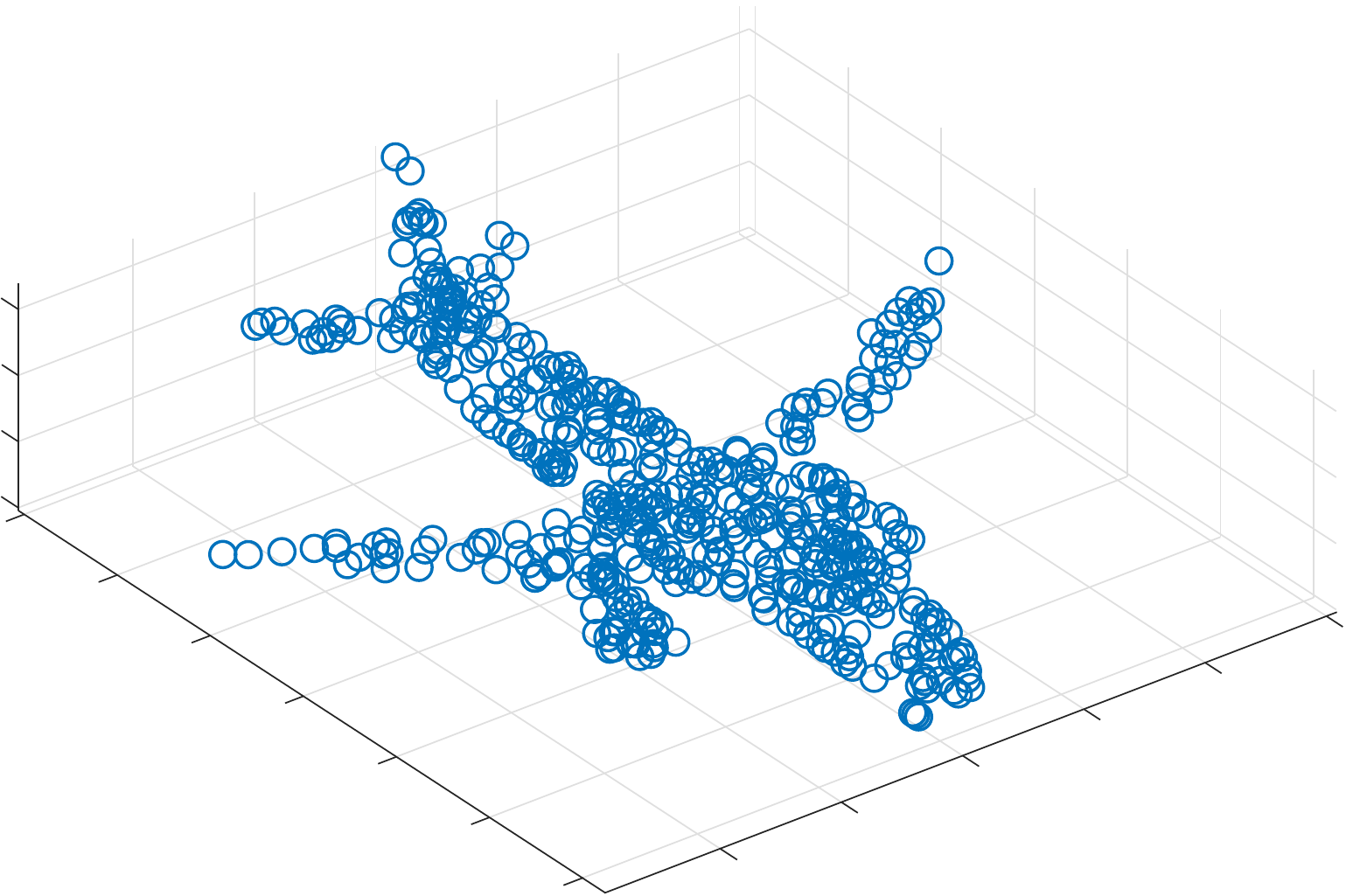}
            \includegraphics[scale=.175]{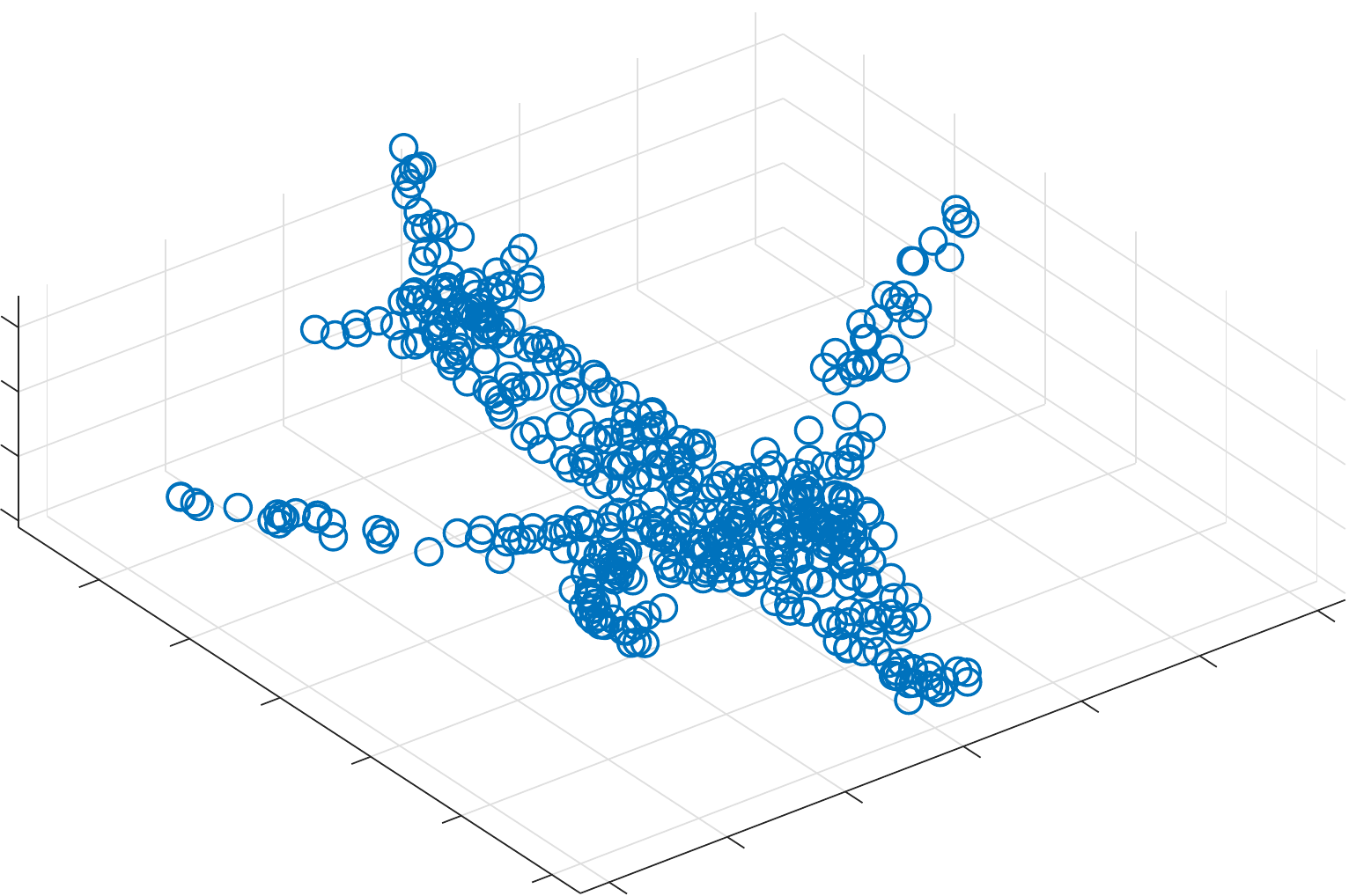}
            \includegraphics[scale=.175]{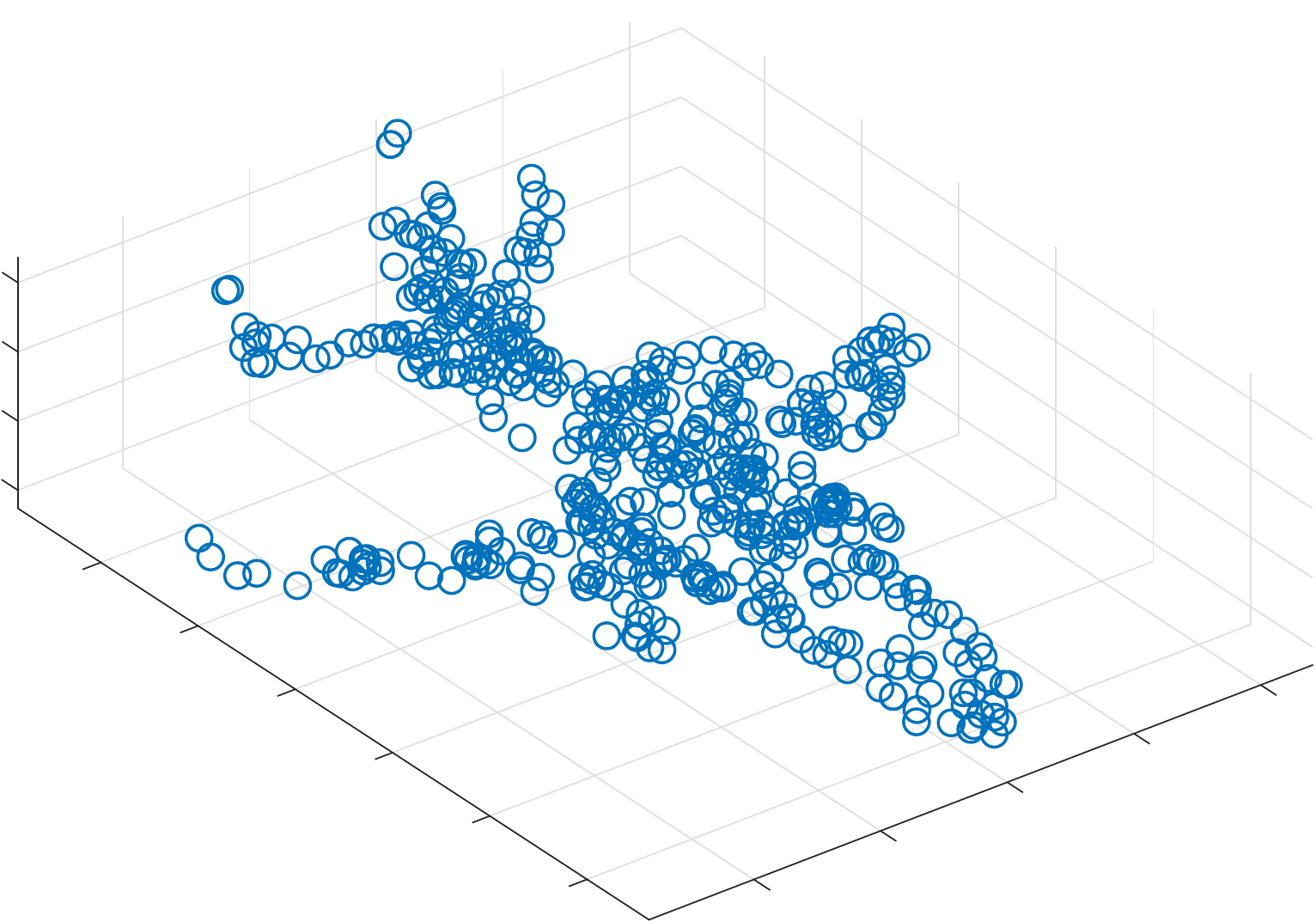}
        \end{minipage}
        \caption{\texttt{Airplanes}}
    \end{subfigure}
    \begin{subfigure}[t]{.25\linewidth}
        \begin{minipage}{\linewidth}
            \centering
            \includegraphics[scale=.175]{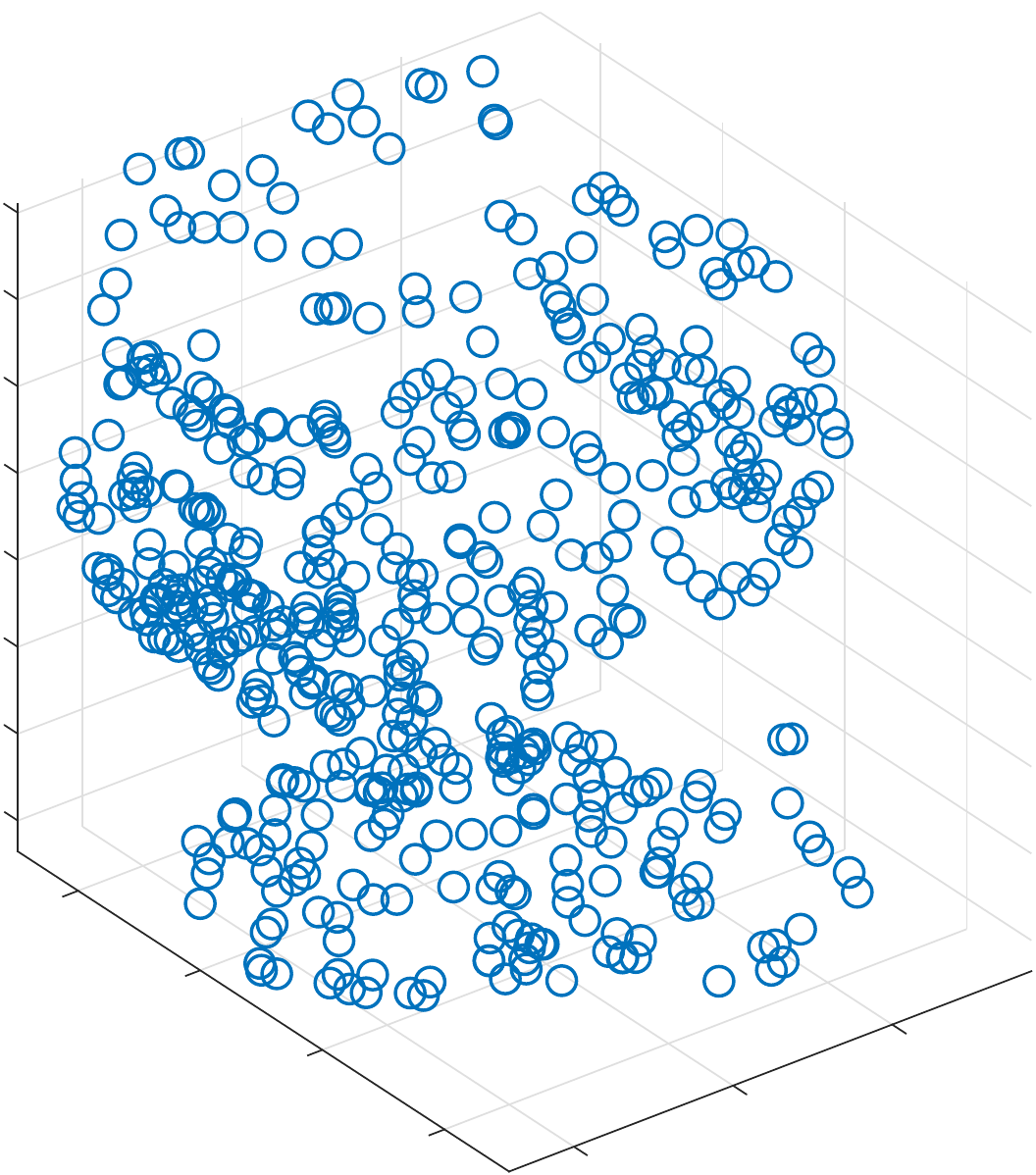} \\
            \includegraphics[scale=.175]{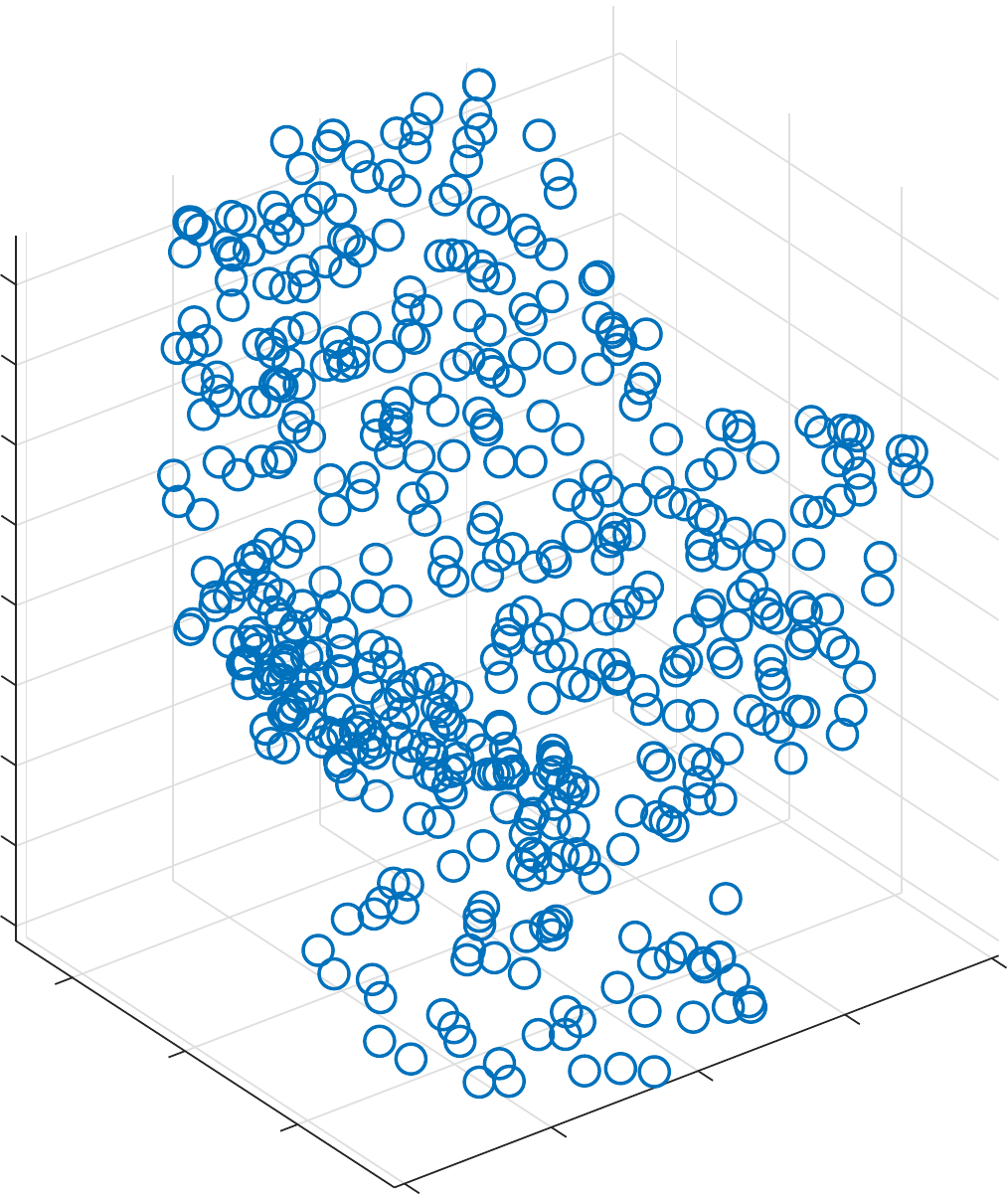} \\
            \includegraphics[scale=.175]{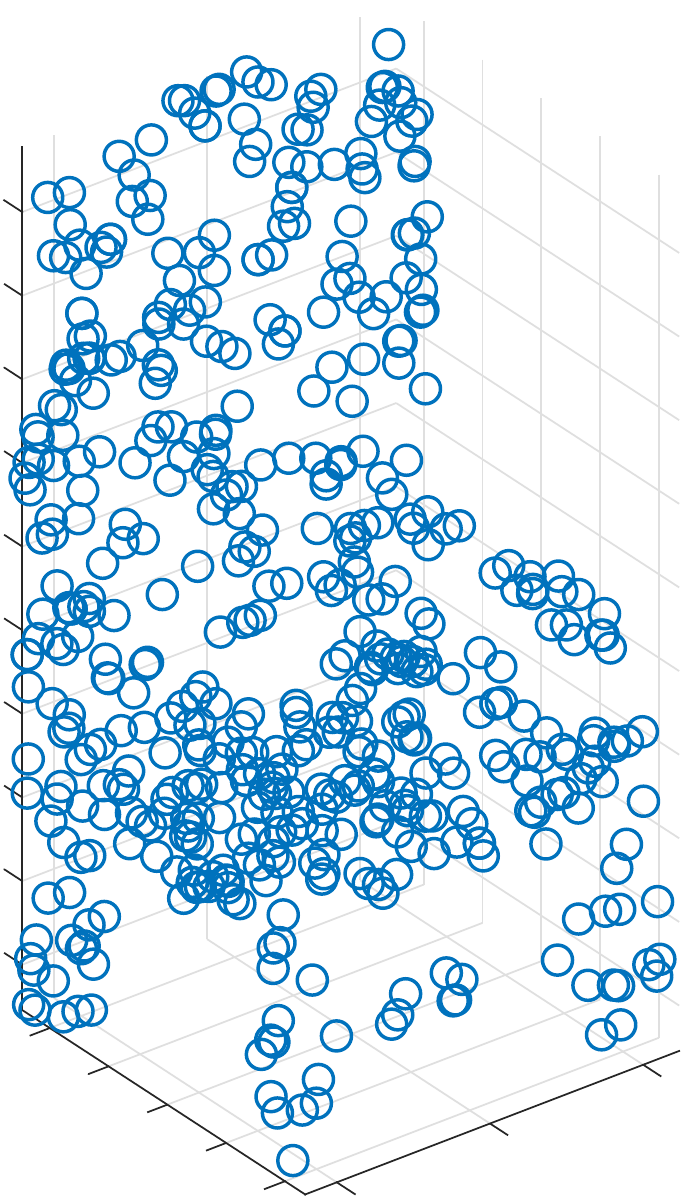} \\
            \includegraphics[scale=.175]{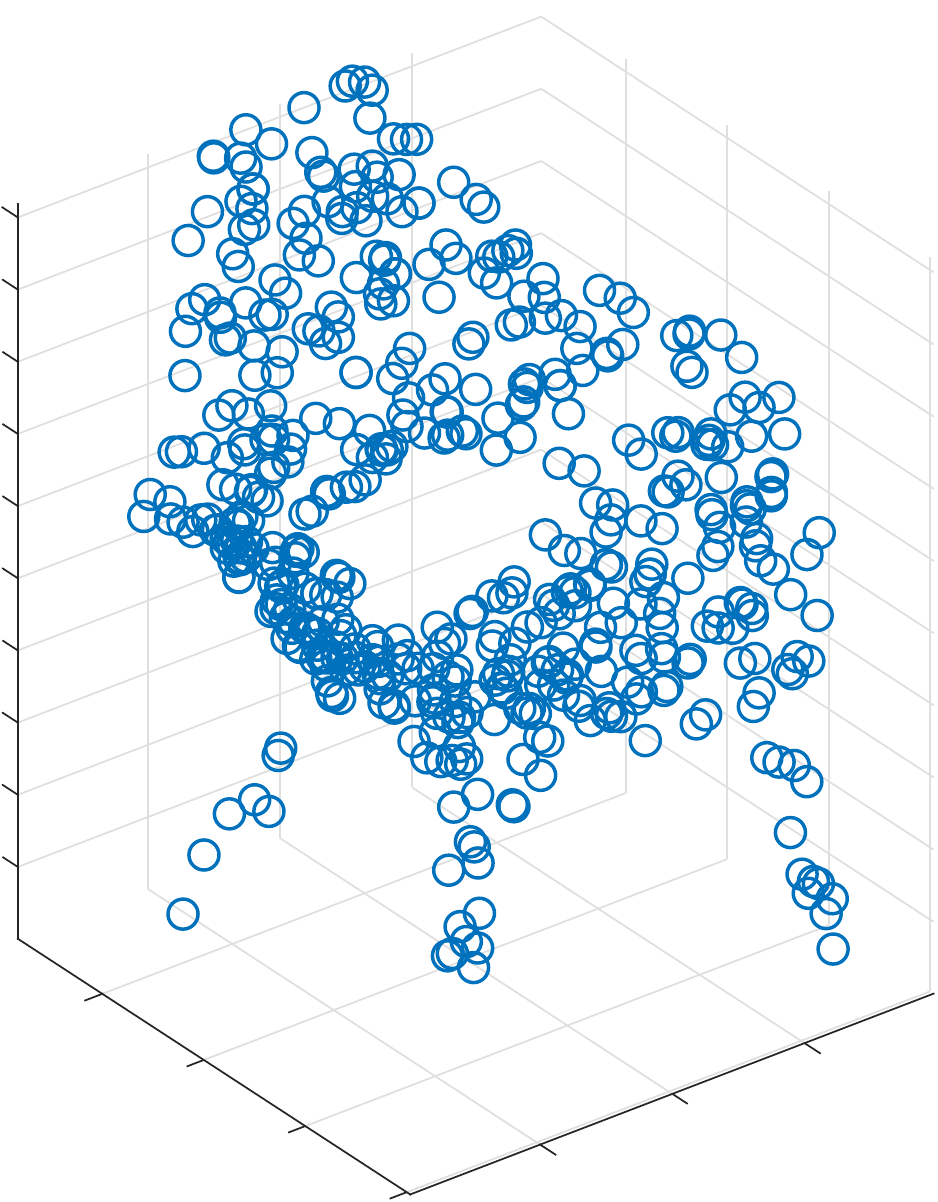}
        \end{minipage}
        \caption{\texttt{Chairs}}
    \end{subfigure}
    \begin{subfigure}[t]{.25\linewidth}
        \begin{minipage}{\linewidth}
            \centering
            \includegraphics[scale=.175]{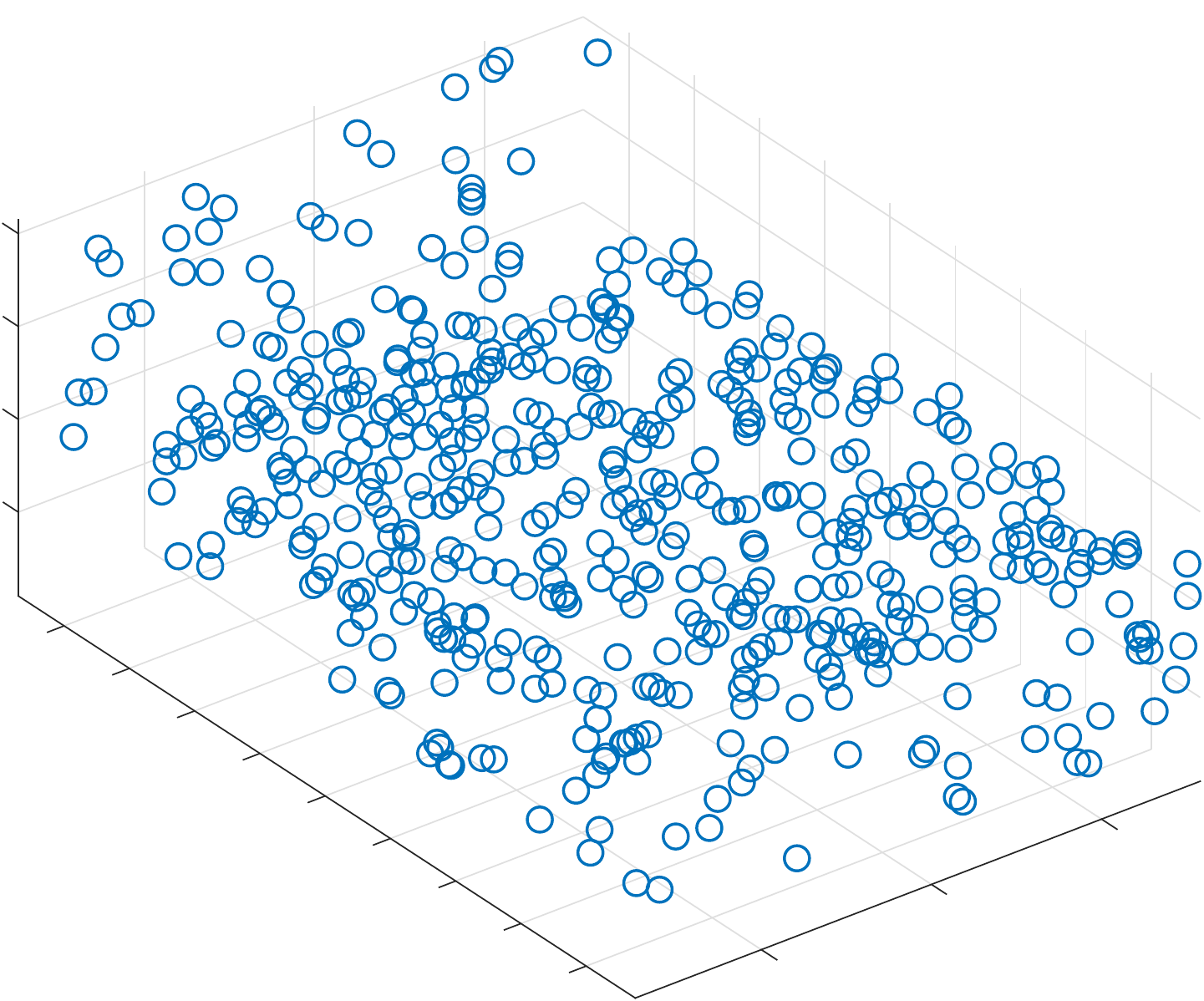}
            \includegraphics[scale=.175]{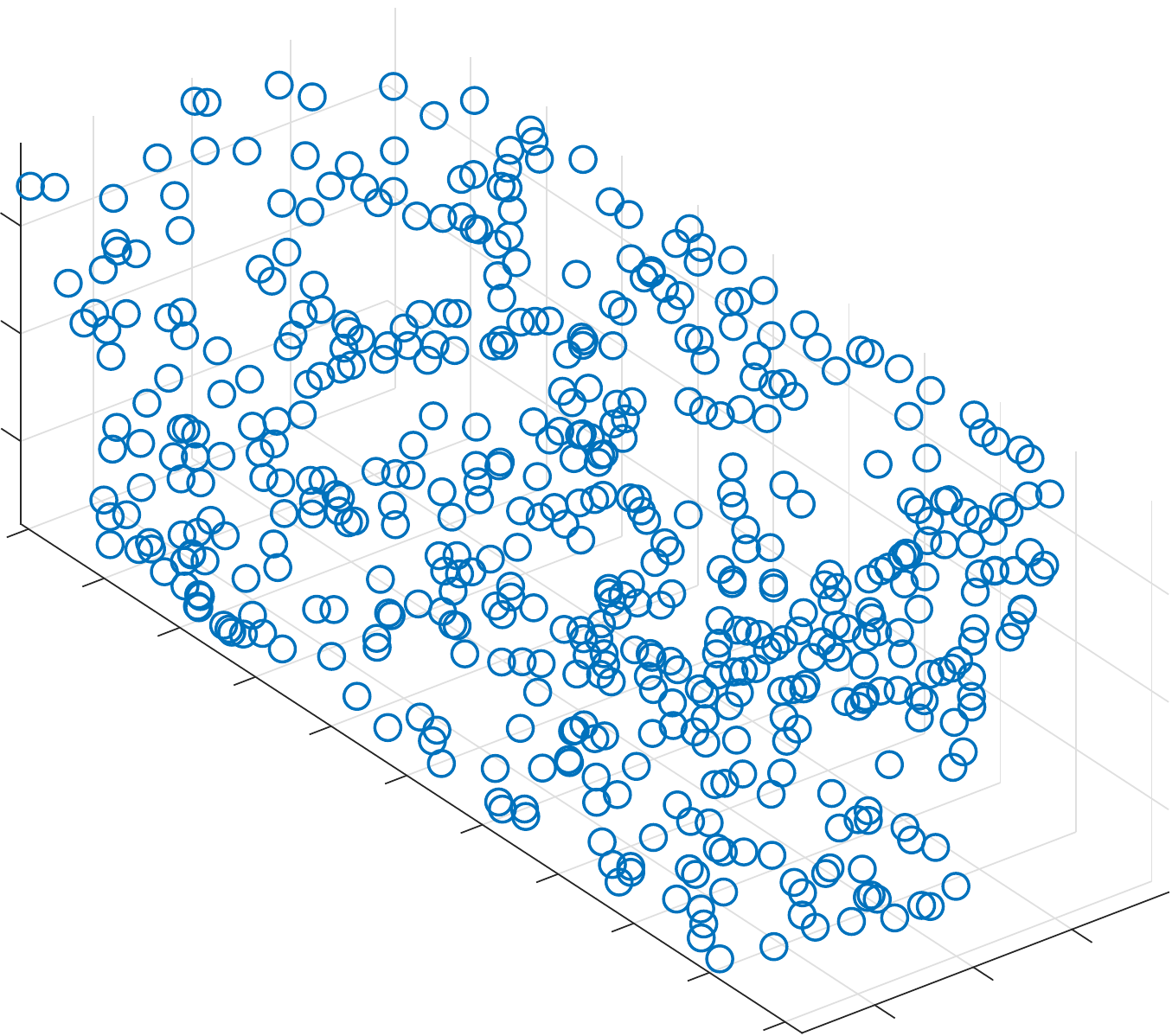} \\
            \includegraphics[scale=.175]{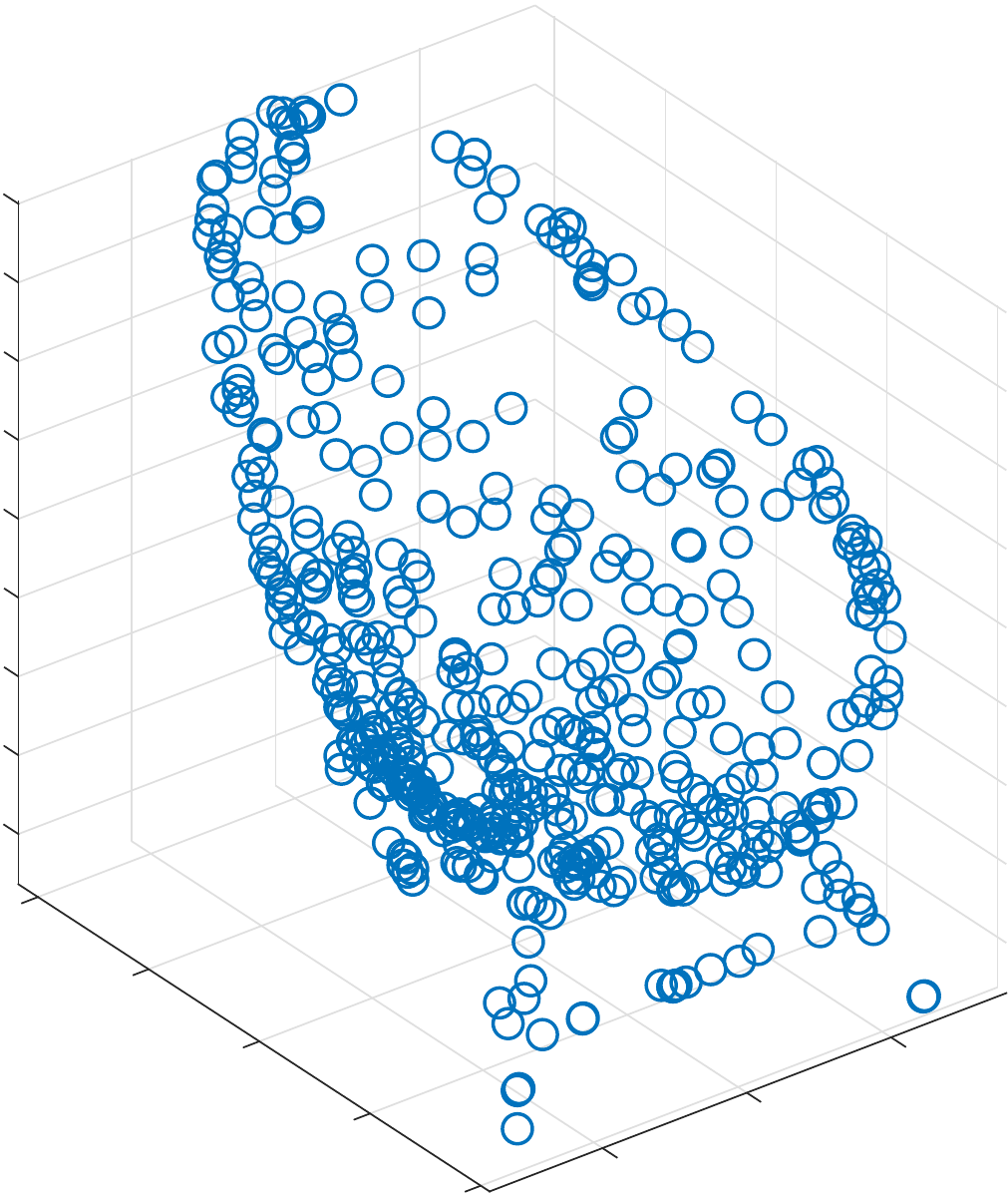}
            \includegraphics[scale=.175]{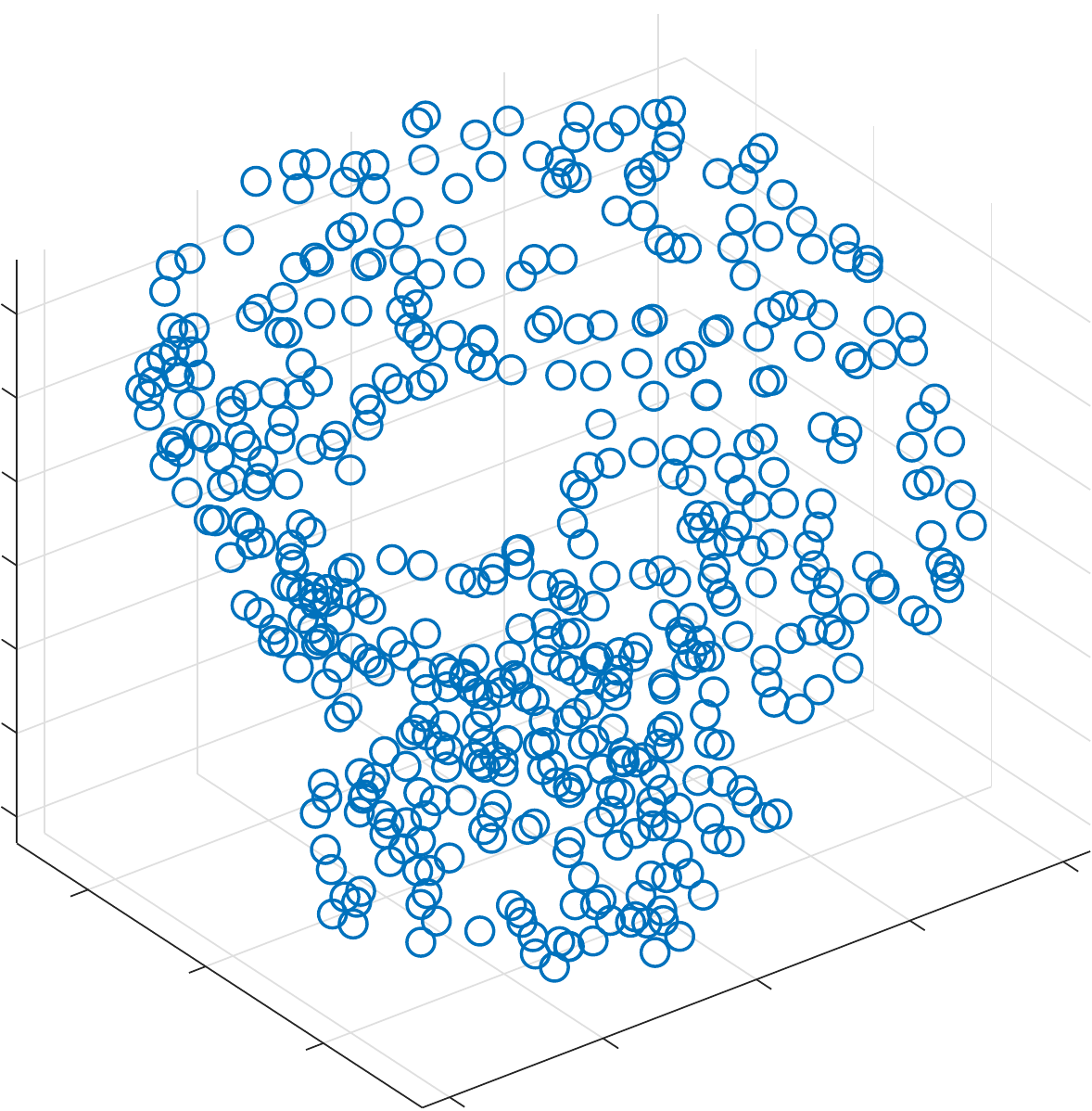}
        \end{minipage}\\
        \caption{\texttt{modelnet10}}
    \end{subfigure}
        \begin{subfigure}[t]{.25\linewidth}
            \begin{minipage}{\linewidth}
                \centering
                \includegraphics[scale=.175]{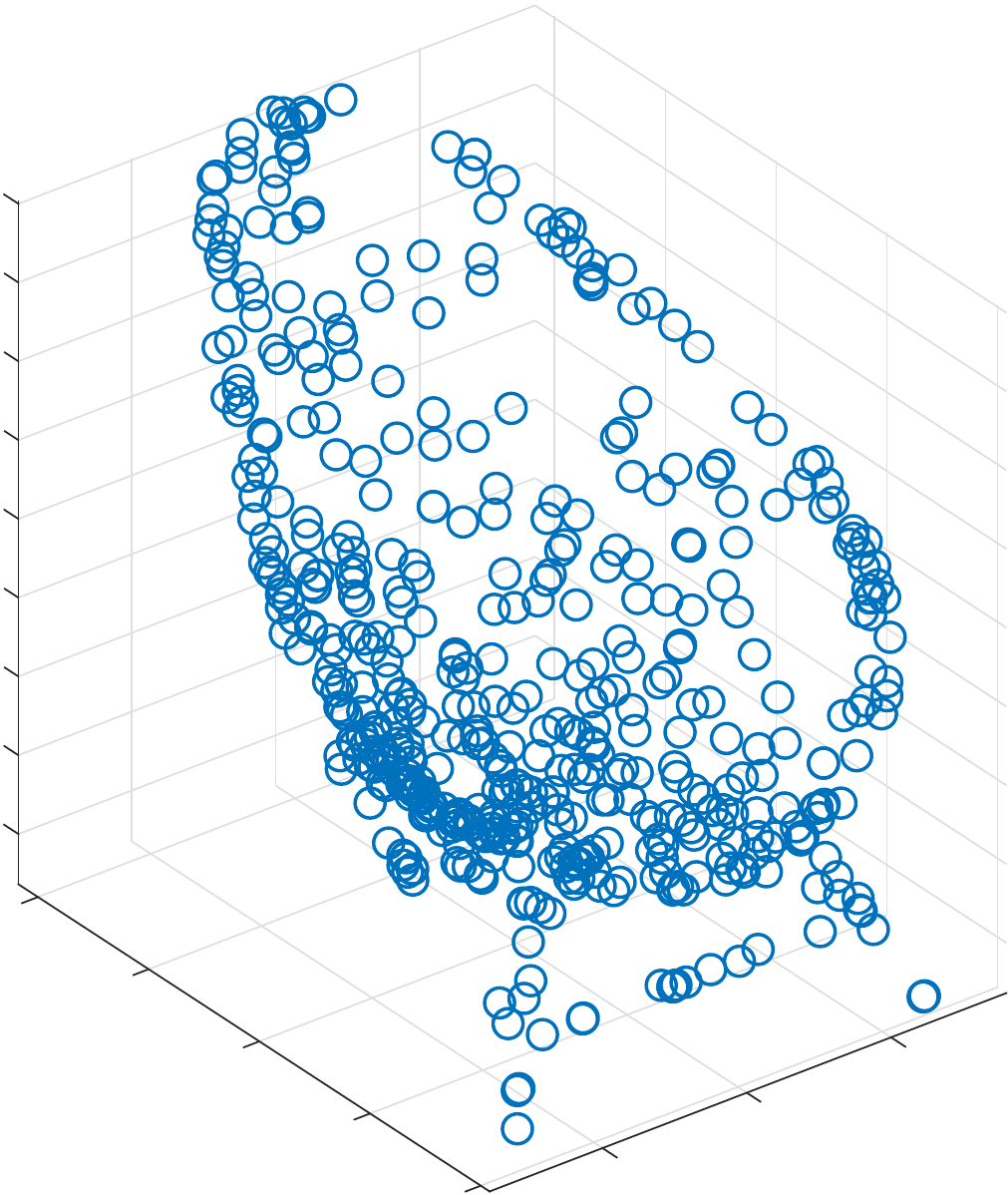}
                \includegraphics[scale=.175]{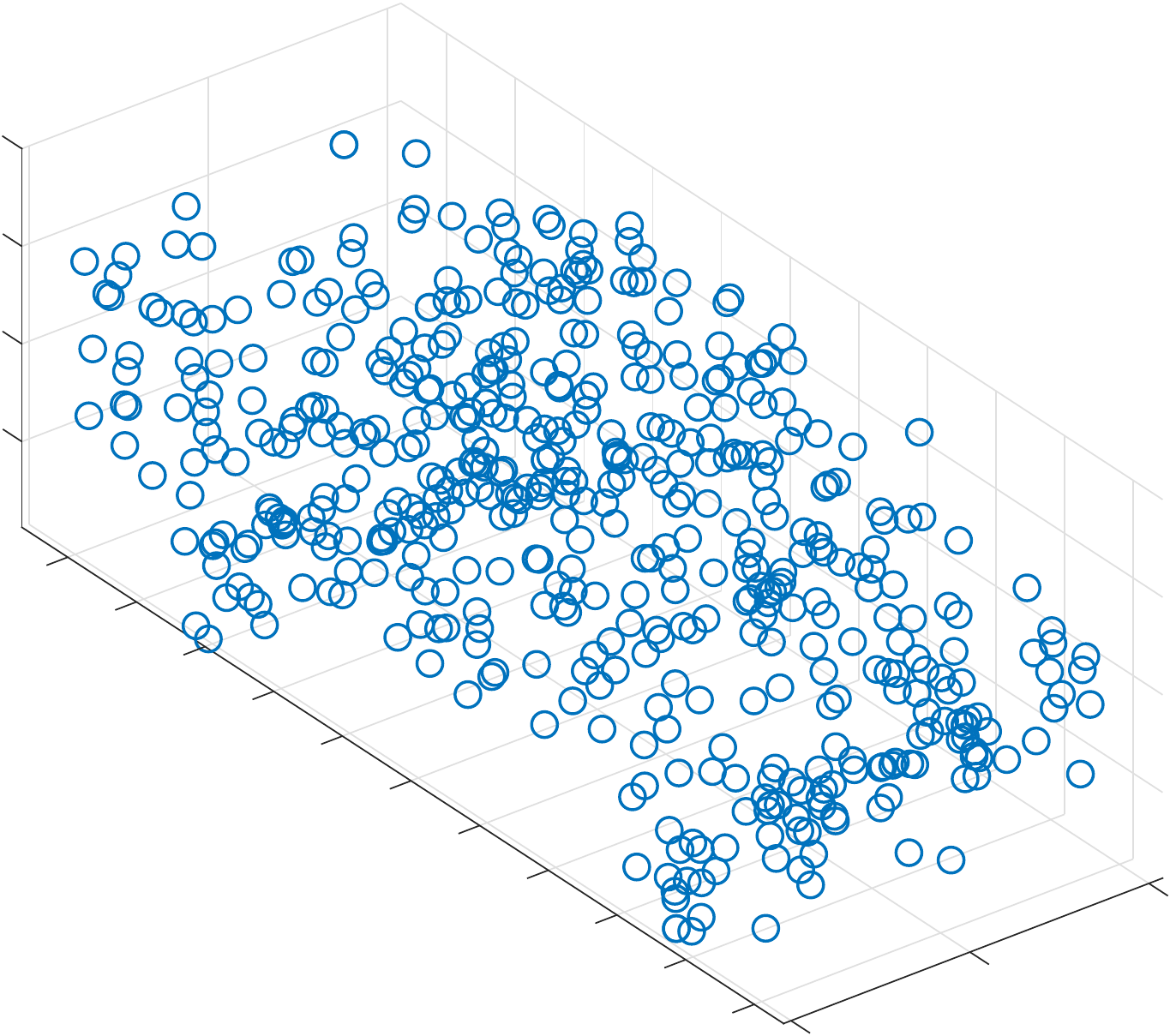} \\
                \includegraphics[scale=.175]{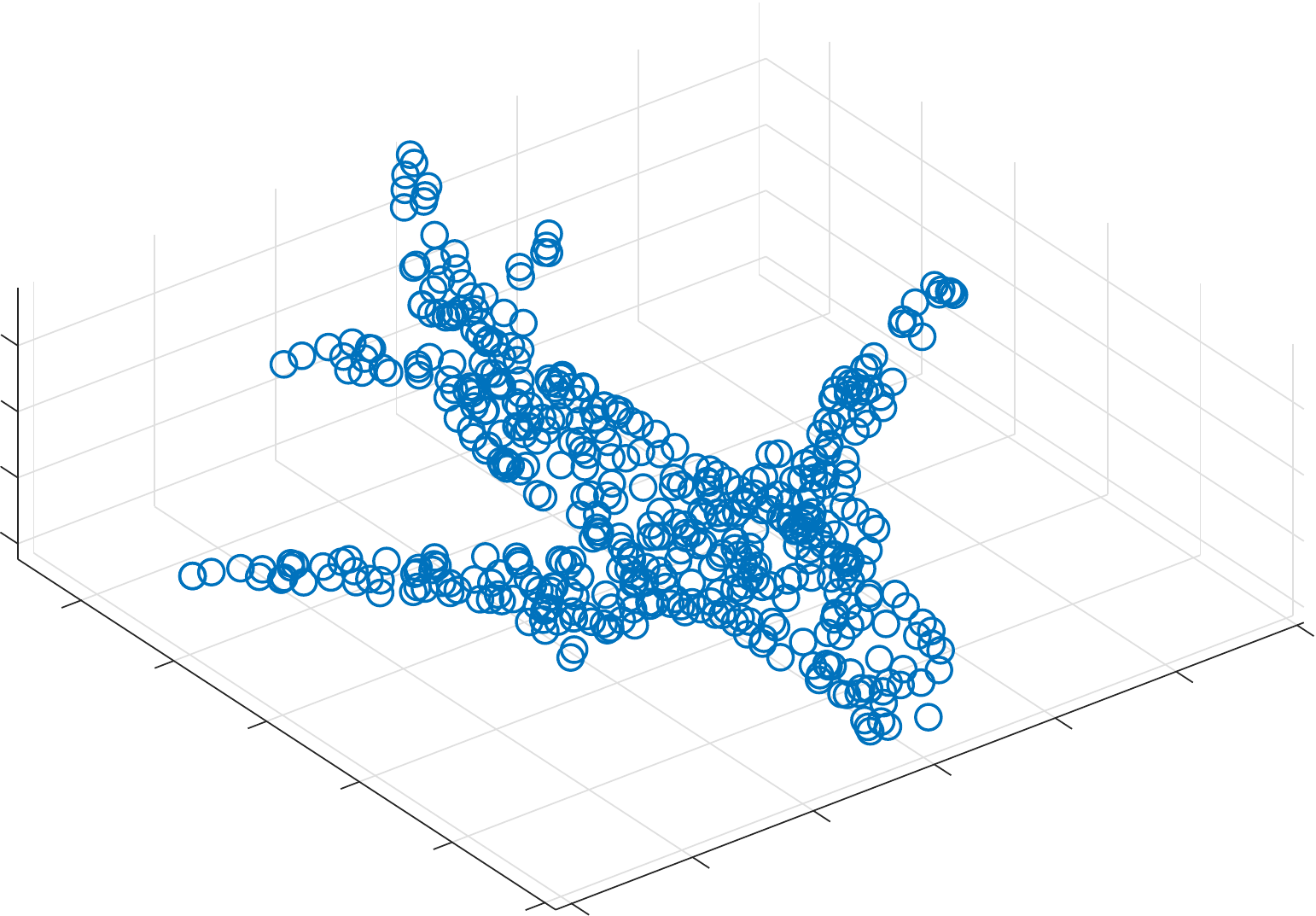}
                \includegraphics[scale=.175]{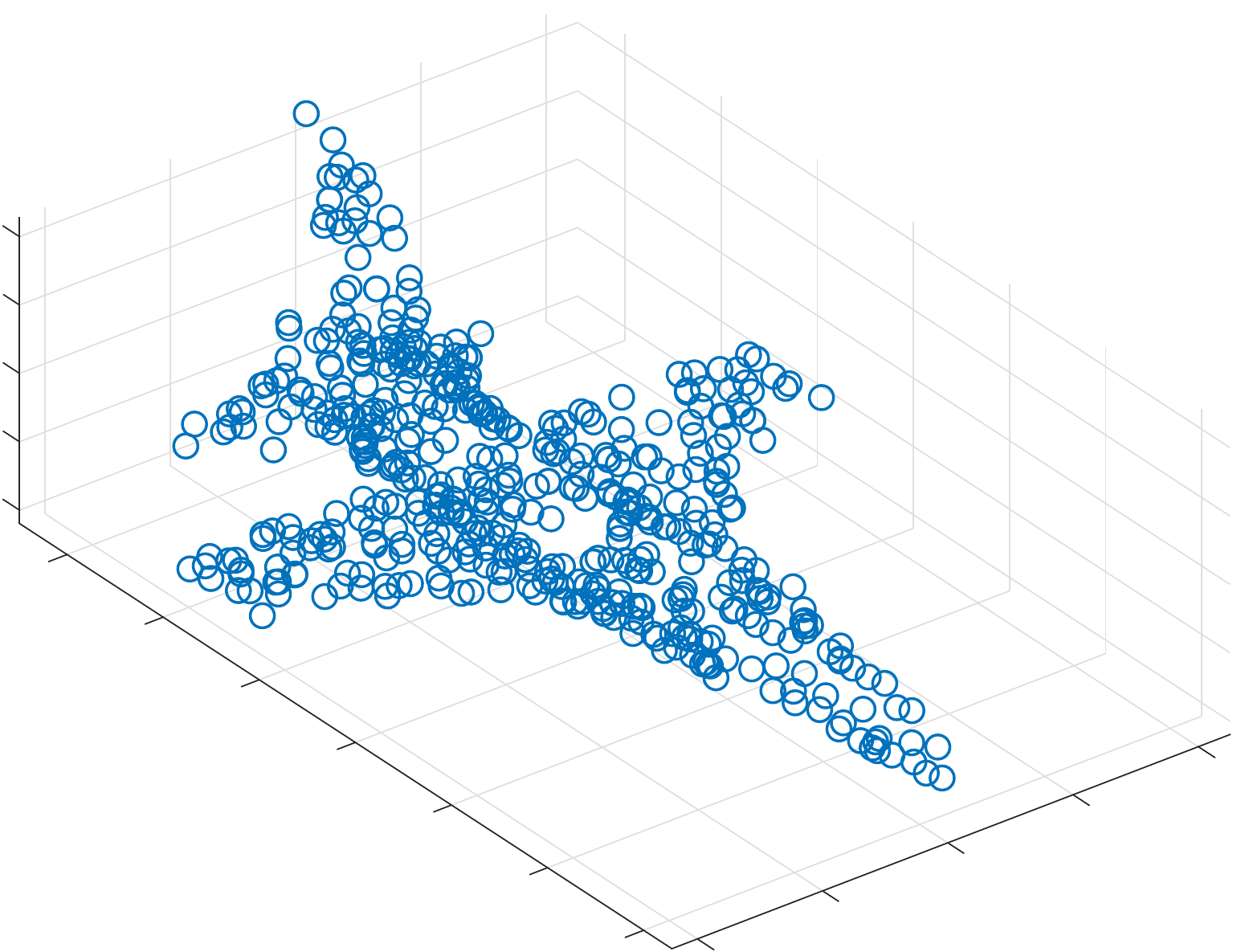}
            \end{minipage}
        \caption{\texttt{modelnet10a}}
    \end{subfigure}    
    \begin{subfigure}[t]{.25\linewidth}
            \begin{minipage}{\linewidth}
                \centering
                \includegraphics[scale=.175]{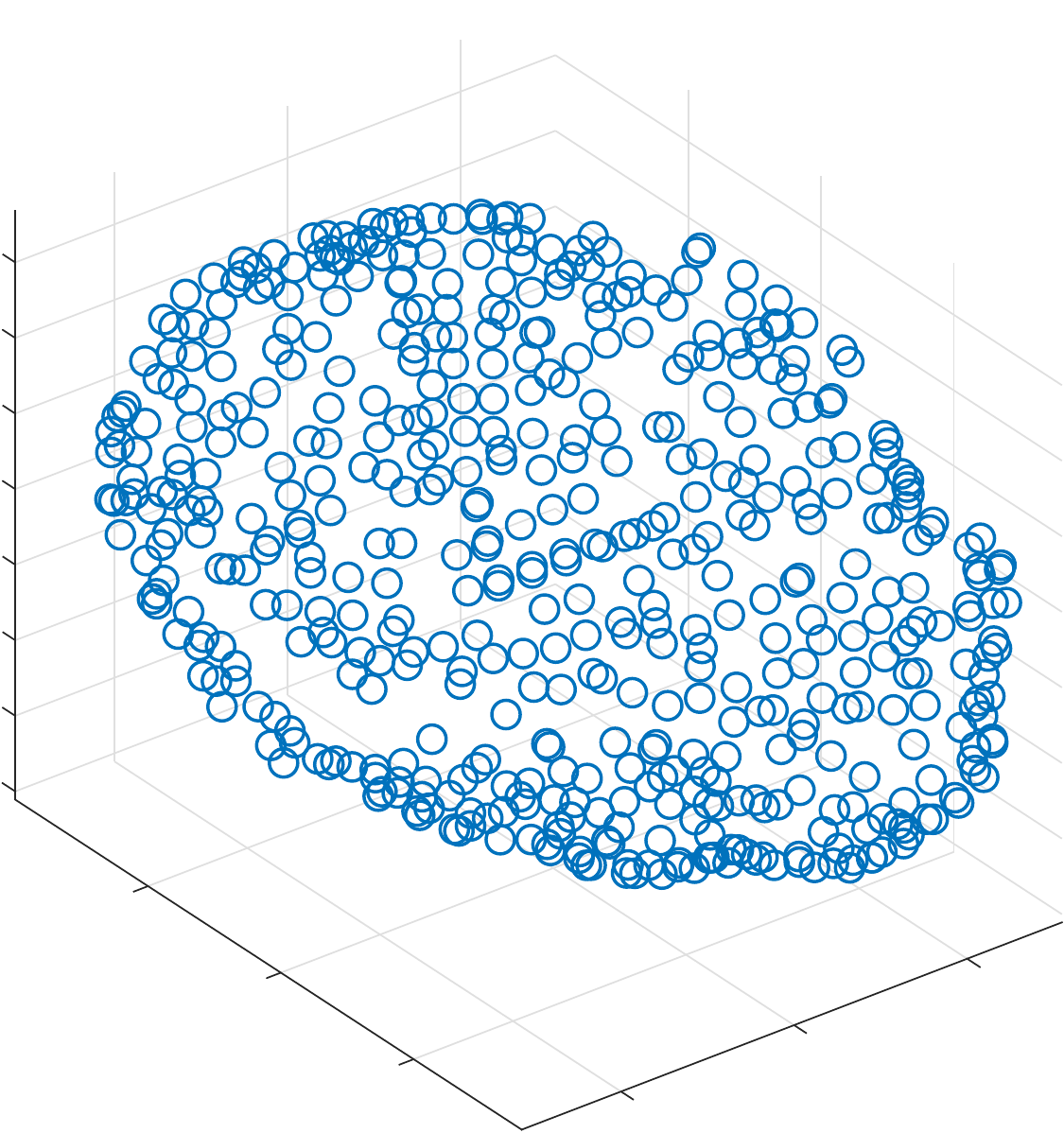}
                \includegraphics[scale=.175]{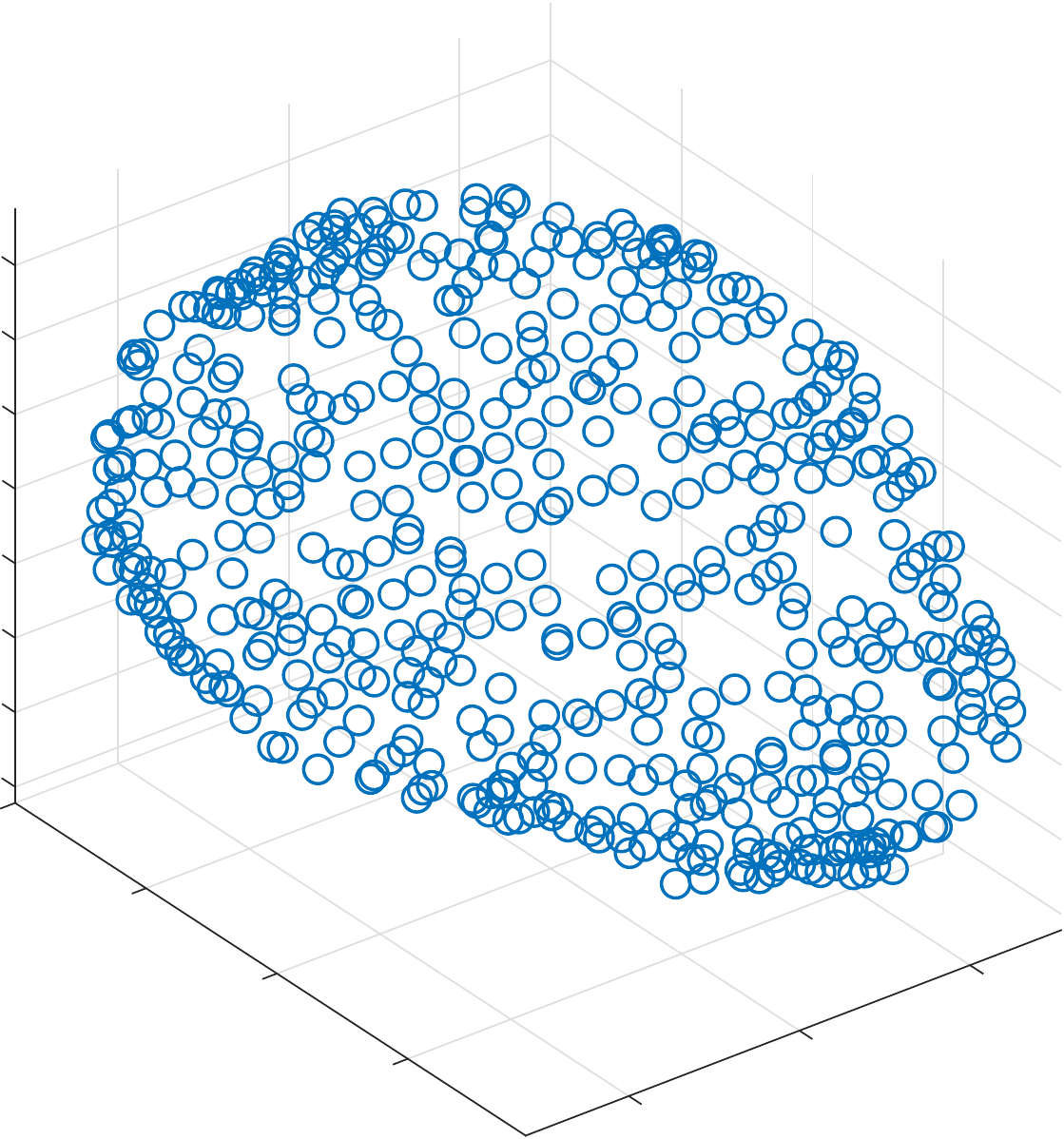} \\
                \includegraphics[scale=.175]{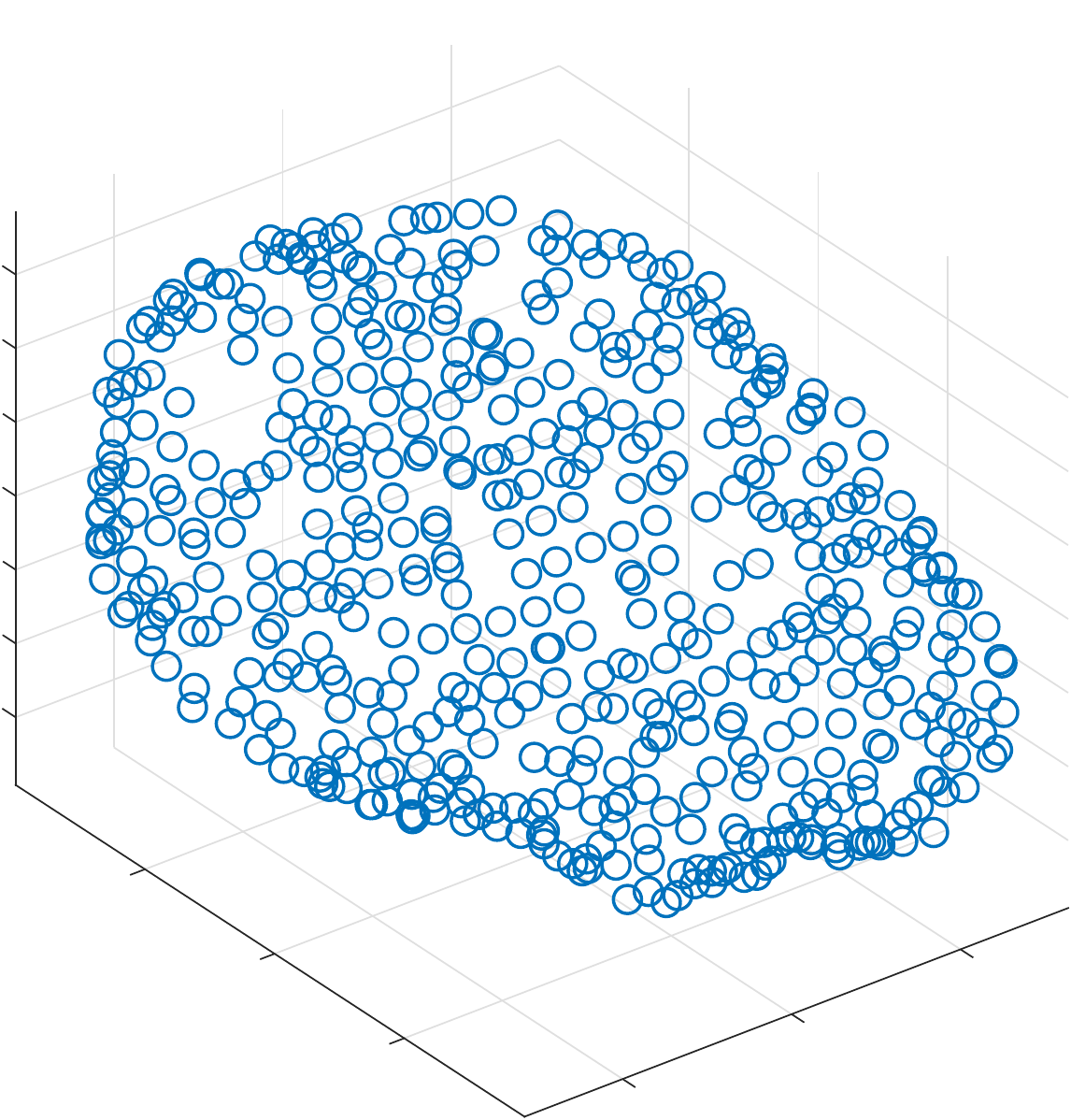}
                \includegraphics[scale=.175]{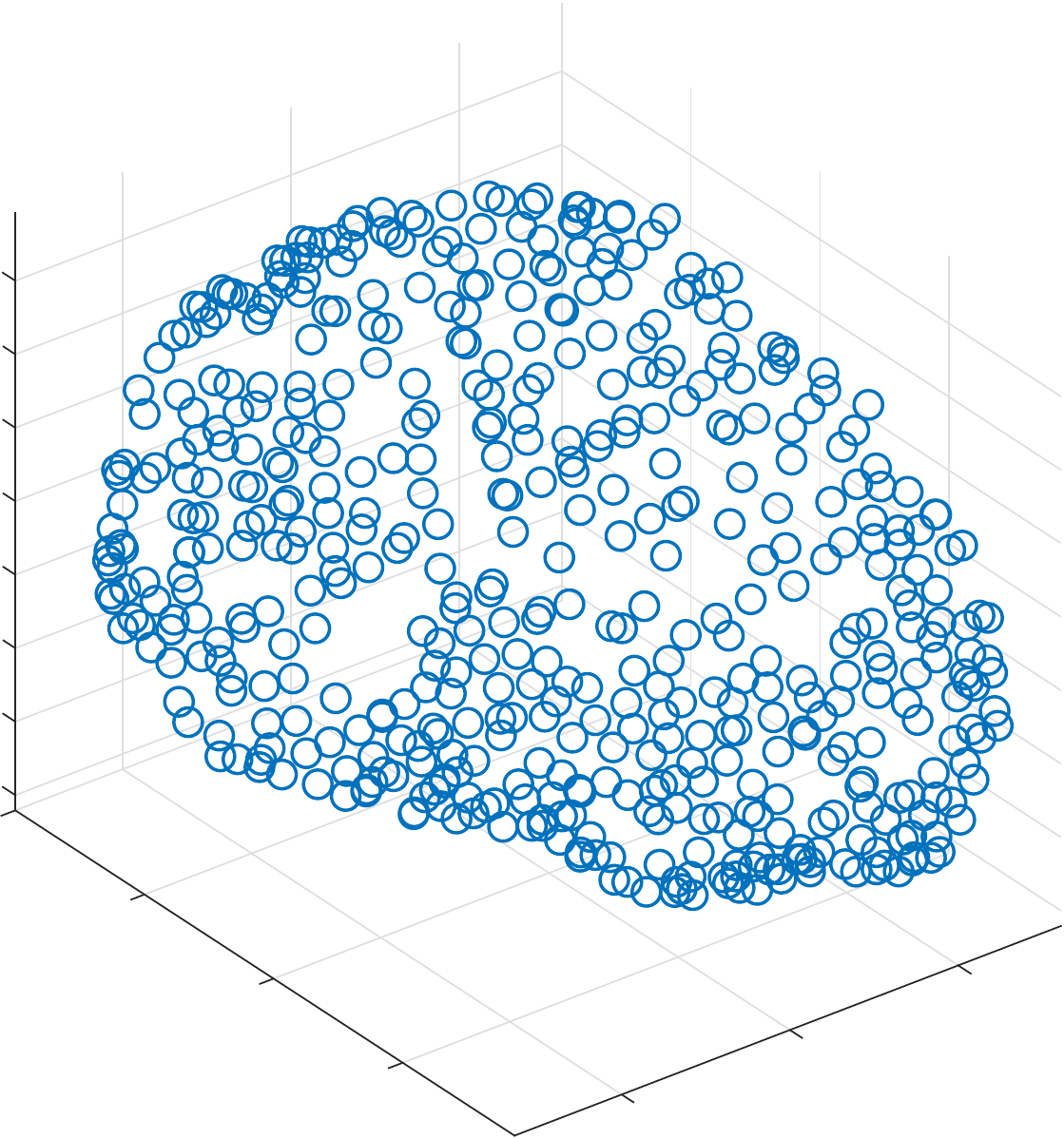}
            \end{minipage}
        \caption{\texttt{thalamus}}
    \end{subfigure}    
    \label{fig:Training samples.}
        \begin{subfigure}[t]{.25\linewidth}
            \begin{minipage}{\linewidth}
                \centering
                \includegraphics[scale=.175]{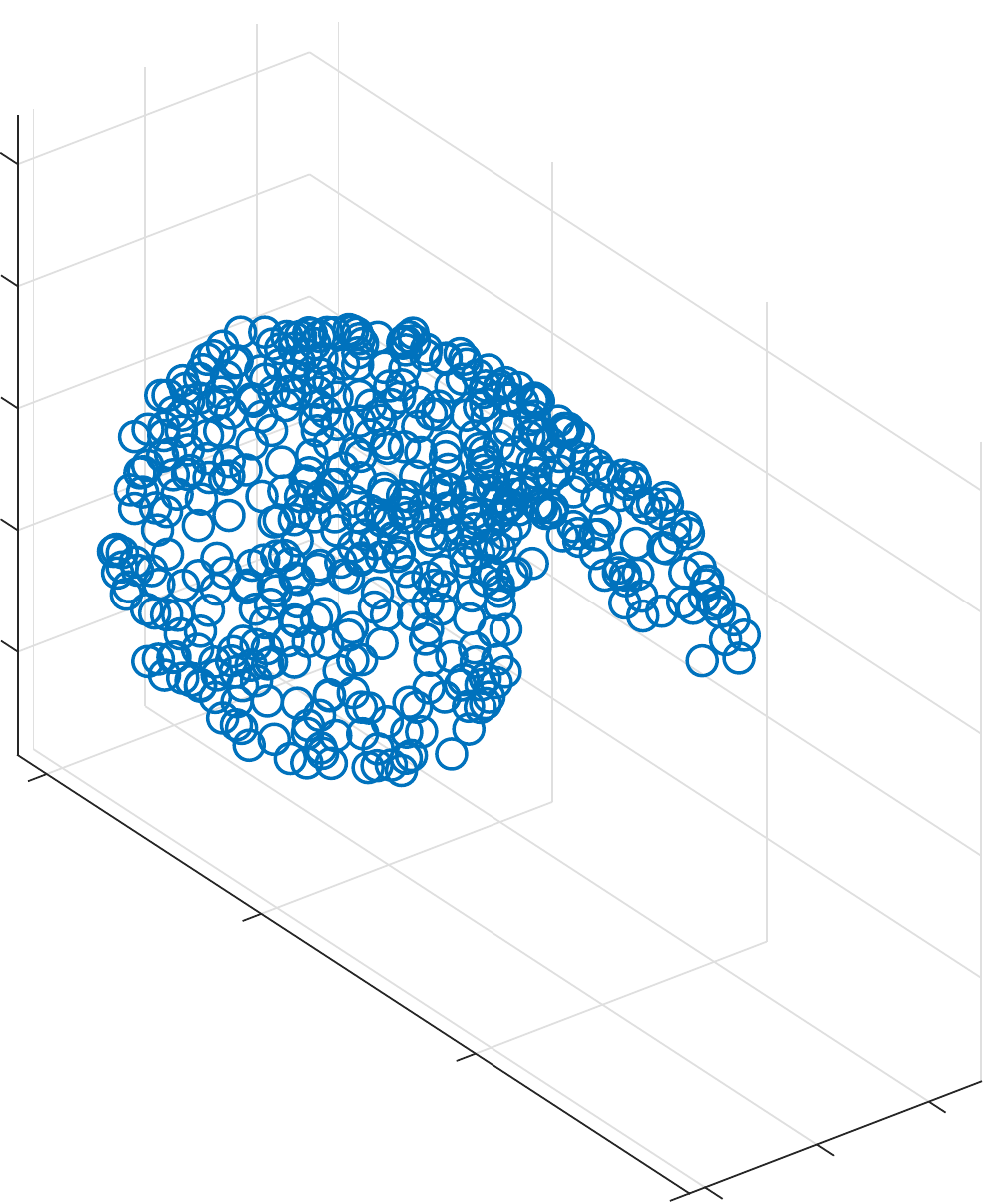}
                \includegraphics[scale=.175]{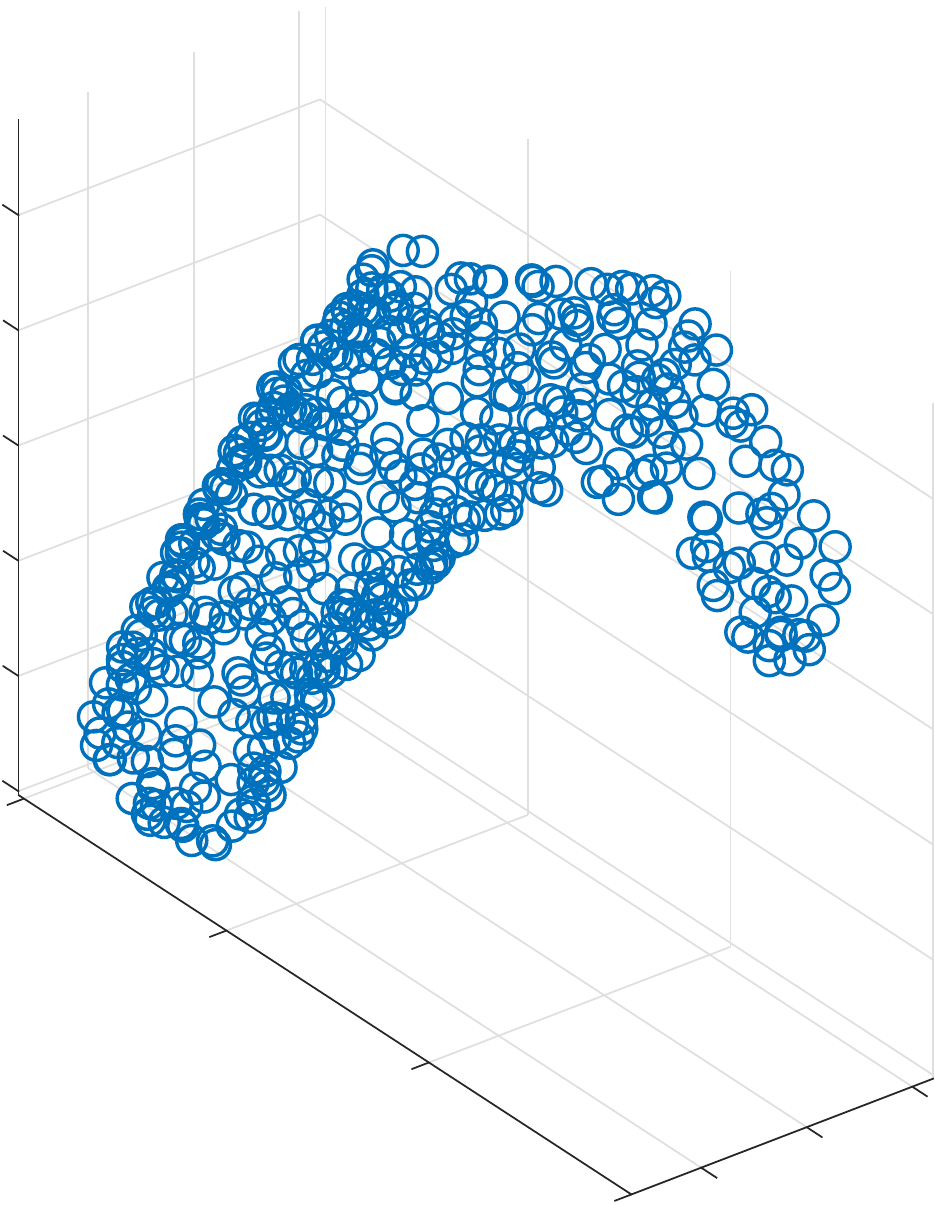} \\
                \includegraphics[scale=.175]{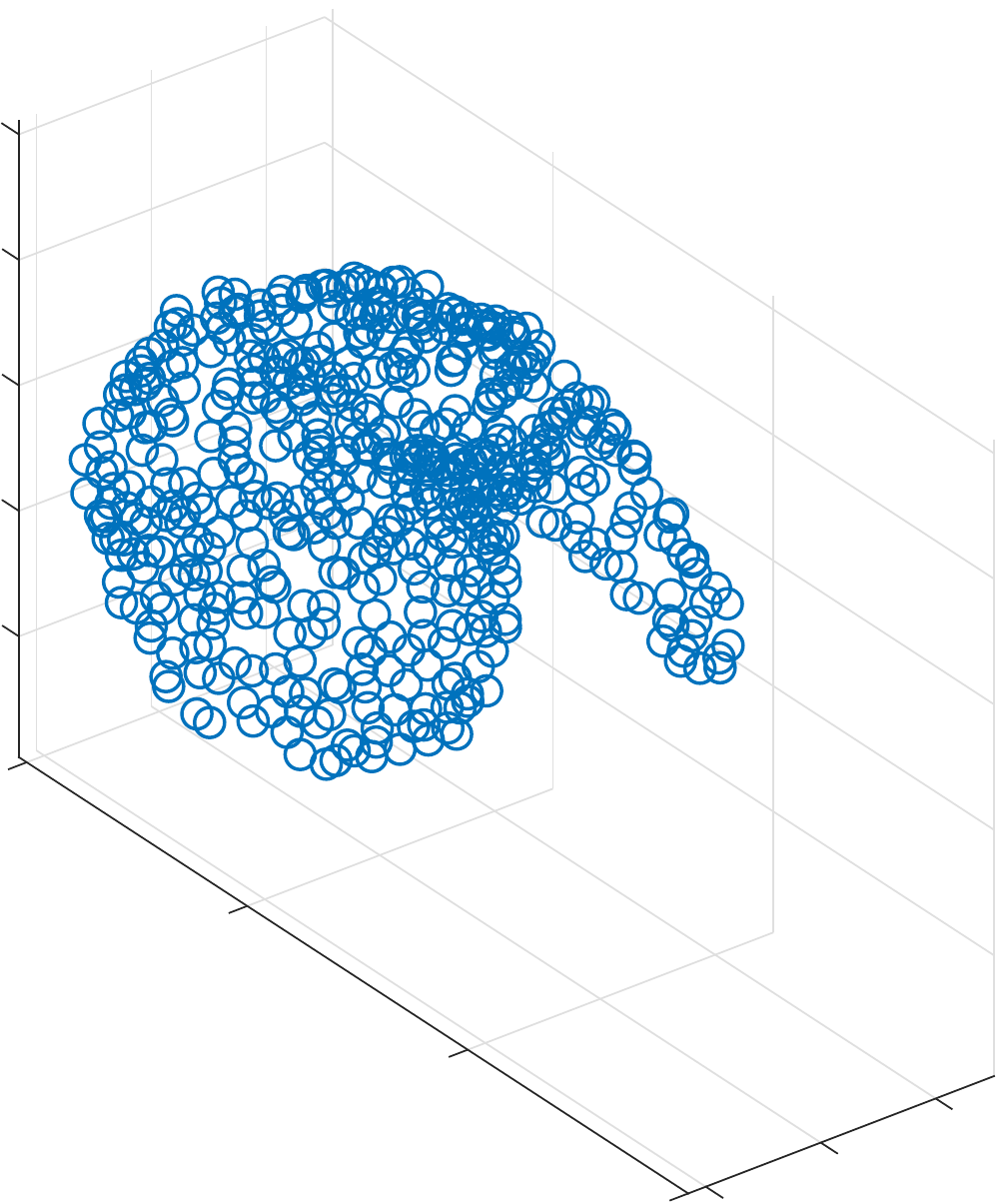}
                \includegraphics[scale=.175]{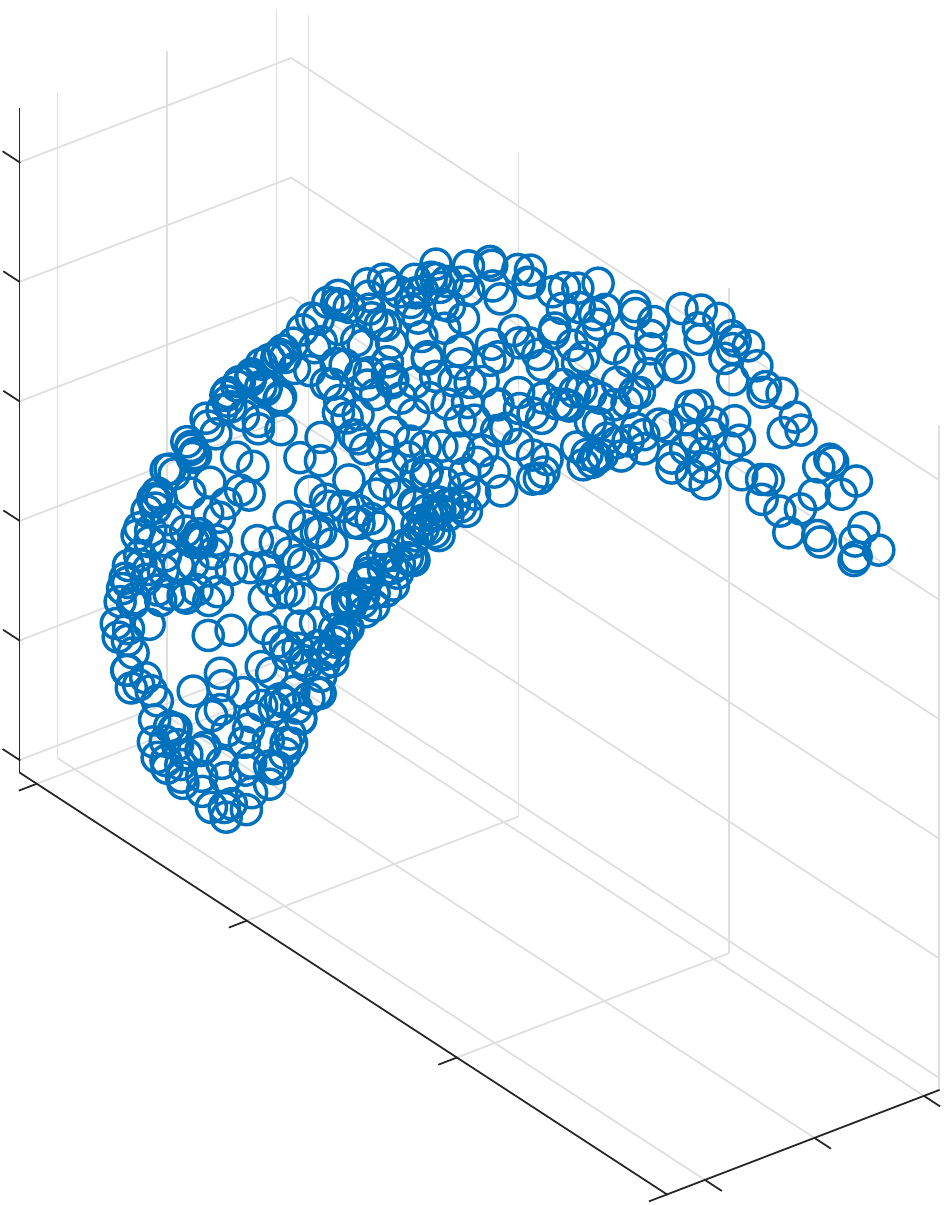}
            \end{minipage}
        \caption{\texttt{caudate}}
    \end{subfigure}    
    \caption{Training examples \label{fig:all_training}}
\end{figure}

The training data includes 10,000 points per set for all ModelNet data sets and approximately 1,000 points for both brain substructures. The various models used a randomly selected, 512 point subset during training, validation, and testing. We plot training instances in Figure \ref{fig:all_training}.

%% modelnet10 dataset
%% Training Chairs
%\begin{figure}[H]
%    \centering
%    \begin{subfigure}[t]{.24\linewidth}
%        \begin{minipage}{\linewidth}
%            \centering
%            \includegraphics[height=16mm]{figures/training_examples/chairs/sample_1.pdf}
%        \end{minipage}
%        \caption{}
%    \end{subfigure}
%    \begin{subfigure}[t]{.24\linewidth}
%        \begin{minipage}{\linewidth}
%            \centering
%            \includegraphics[height=10mm]{figures/training_examples/chairs/sample_2.pdf} \\
%        \end{minipage}
%        \caption{}
%    \end{subfigure}
%    \begin{subfigure}[t]{.24\linewidth}
%        \begin{minipage}{\linewidth}
%            \centering
%            \includegraphics[height=10mm]{figures/training_examples/chairs/sample_3.pdf}
%        \end{minipage}
%        \caption{}
%    \end{subfigure}
%        \begin{subfigure}[t]{.24\linewidth}
%        \begin{minipage}{\linewidth}
%            \centering
%            \includegraphics[height=14mm]{figures/training_examples/chairs/sample_4.pdf}
%        \end{minipage}
%        \caption{}
%    \end{subfigure}
%    \caption{Training examples}
%    \label{fig:training_examples}
%\end{figure}
%%% TRANSFORMATIONS %%%

\end{document}